\documentclass{article}

\PassOptionsToPackage{numbers}{natbib}
\usepackage[final]{nips_2017}

\usepackage{etoolbox}
\newtoggle{arxiv}

\toggletrue{arxiv}

\usepackage{amsmath}
\usepackage{amssymb}
\usepackage{amsthm}
\usepackage{algorithm}
\usepackage{algorithmic}
\usepackage{bbm}
\usepackage{bm}
\usepackage{booktabs}
\usepackage{cancel}
\usepackage{comment}
\usepackage{hyperref}
\usepackage{hyperref}
\usepackage{mathtools}
\usepackage{nicefrac}
\usepackage{subcaption}
\usepackage{capt-of}
\usepackage{thm-restate}
\usepackage{wrapfig}
\usepackage{cleveref}
\usepackage{graphicx}
\usepackage{pgf}

\newcommand{\eigvec}[1]{u^{(#1)}}
\newcommand{\peigvec}[1]{\tilde{u}^{(#1)}}
\newcommand{\eigval}{\lambda}
\newcommand{\peigval}{\tilde{\lambda}}
\newcommand{\nzeigval}{\lambda}

\newcommand{\nzeigvec}[1]{u^{(#1)}}
\newcommand{\nzpeigvec}[1]{\tilde{u}^{(#1)}}

\newcommand{\countingset}[1]{[#1]}
\newcommand{\transpose}{\intercal}
\newcommand{\Vecspan}{\operatorname{Span}}
\newcommand{\reals}{\mathbb{R}}
\newcommand{\posnaturals}{\mathbb{N}^+}
\newcommand{\term}{\emph}
\newcommand{\rv}{}
\newcommand{\expectation}{\mathbb{E}}
\newcommand{\prob}{\mathbb{P}}
\newcommand{\indicator}[1]{\mathbbm{1}_{#1}}

\newcommand{\header}[1]{\textbf{#1}}

\DeclarePairedDelimiter\abs{\lvert}{\rvert}
\DeclarePairedDelimiter\spectralnorm{\lVert}{\rVert}
\DeclarePairedDelimiter\frobnorm{\lVert}{\rVert_F}
\DeclarePairedDelimiter\twonorm{\lVert}{\rVert_2}
\DeclarePairedDelimiter\maxnorm{\lVert}{\rVert_\text{max}}
\DeclarePairedDelimiter\inftynorm{\lVert}{\rVert_\infty}

\newtheoremstyle{compacttheorem}
{2pt} 
{0pt} 
{\it} 
{} 
{\bfseries} 
{.} 
{.5em} 
{} 

\newtheoremstyle{compactdefinition}
{2pt} 
{0pt} 
{} 
{} 
{\bfseries} 
{.} 
{.5em} 
{} 

\newtheorem{lemma}{Lemma}
\newtheorem{corollary}{Corollary}
\newtheorem{theorem}{Theorem}
\newtheorem{toytheorem}{``Theorem''}
\newtheorem{claim}{Claim}
\theoremstyle{definition}
\newtheorem*{notation}{Notation}
\newtheorem{setting}{Setting}
\newtheorem{example}{Example}
\newtheorem{defn}{Definition}
\newtheorem{remark}{Remark}

\title{Unperturbed: spectral analysis beyond Davis-Kahan}

\author{
  Justin Eldridge \quad Mikhail Belkin \quad Yusu Wang\\
  The Ohio State University\\
  \texttt{\{eldridge, mbelkin, yusu\}@cse.ohio-state.edu}
}
\date{}

\begin{document}

\maketitle

\begin{abstract}
    Classical matrix perturbation results, such as Weyl's theorem for
    eigenvalues and the Davis-Kahan theorem for eigenvectors, are general
    purpose.  These classical bounds are tight in the worst case, but in many
    settings sub-optimal in the typical case.  In this paper, we present
    perturbation bounds which consider the nature of the perturbation and its
    interaction with the unperturbed structure in order to obtain significant
    improvements over the classical theory in many scenarios, such as when the
    perturbation is random. We demonstrate the utility of these new results by
    analyzing perturbations in the stochastic blockmodel where we derive much
    tighter bounds than provided by the classical theory.  We use our new
    perturbation theory to show that a very simple and natural clustering
    algorithm -- whose analysis was difficult using the classical tools --
    nevertheless recovers the communities of the blockmodel exactly even in very
    sparse graphs.
\end{abstract}

\section{Introduction}

In many applications the interesting structure of information is encoded by the
eigenvalues and eigenvectors of an appropriately-defined matrix. For instance,
the top eigenvectors of the covariance matrix reveal the principal directions of
the distribution, and the bottom eigenvalues and eigenvectors of a graph's
Laplacian capture important details about its cluster structure. When learning
from data, however, we typically do not have access to the matrix itself but
rather a version which has been contaminated by (oftentimes random)
noise. In such cases the following problem is of great interest: let
$M$ and $H$ be $n \times n$ symmetric matrices with real entries.  Suppose we
``perturb'' the matrix $M$ by adding $H$.  How do the eigenvalues and
eigenvectors of $M + H$ relate to those of $M$?

For eigenvalues, the classical answer to this question comes in the form of
Weyl's theorem \cite{Weyl1912-kl}. Let the eigenvalues of $M$ be $\eigval_1
\geq \cdots \geq \eigval_n$ and the eigenvalues of $M + H$ be $\peigval_1 \geq
\cdots \geq \peigval_n$. Denote by $\spectralnorm{H}$ the \term{spectral norm}
of $H$; that is, the largest eigenvalue of $H$ in absolute value. We have:

\begin{theorem}[Weyl's theorem]
    For any $i \in \countingset{n}$, $\abs{\eigval_i - \peigval_i} \leq
    \spectralnorm{H}.$ 
\end{theorem}

For the perturbation of eigenvectors, the classical result is
the Davis-Kahan  theorem \cite{Davis1969-pd}. For any fixed $t \in
\countingset{n}$, let $\eigvec{t}$ be an eigenvector of $M$ with eigenvalue
$\eigval_t$, and let $\peigvec{t}$ be an eigenvector of $M + H$ with eigenvalue
$\peigval_t$. Assume that $\eigval_t$ and $\peigval_t$ have unit multiplicity;
this assumption can be removed at the cost of complicating the statement of the
result.  The Davis-Kahan theorem bounds the angle $\theta_t$ between
$\eigvec{t}$ and $\peigvec{t}$:

\begin{theorem}[The Davis-Kahan theorem]
    Define 
    $
    \delta_t = \min \{
        \abs{\peigval_j - \eigval_t} : j \neq t
    \}.
    $
    Then
    $\sin \theta_t \leq \nicefrac{\spectralnorm{H}}{\delta_t}$.
\end{theorem}

These classical results bound matrix perturbations in general cases, and
do not use information about the structure of the matrices $M$ and $H$ or the
relation between them.  In applications, however, we often make assumptions
about the nature of $M$ and $H$; for example, we might assume that $H$ is
random noise added to a low rank $M$. In such instances the generality of Weyl's
theorem and Davis-Kahan may result in severely suboptimal bounds. 

In this work we present perturbation bounds which incorporate knowledge of the
interaction between $H$ and the eigenvectors of $M$.
We obtain significant improvements over the classical theory in settings
where this interaction is weak, such as when the perturbation $H$ is random. In
\Cref{sec:eigenvalues}, we present an eigenvalue perturbation bound
in the following spirit:
\begin{toytheorem}
    In many settings,
    $\abs{\peigval_t - \eigval_t}$ is on the order of $\abs{\langle \eigvec{t}, H
    \eigvec{t} \rangle} \ll \spectralnorm{H}$.
\end{toytheorem}
We will show that when $H$ is random the perturbation of the top
eigenvalues is on the order of $\sqrt{\log n}$, whereas Weyl's theorem gives
a bound on the order of $\sqrt{n}$.
Next, in \Cref{sec:eigenvectors}, we develop a theory of eigenvector perturbations in
$\infty$-norm which is informally stated as follows:
\begin{toytheorem}
    In many settings, $\inftynorm{\peigvec{t} - \eigvec{t}}$ is on the order of
    $
        {\textstyle
        \inftynorm*{
            \sum_{p \geq 1}^\infty
            \left(
                H/\eigval_t
            \right)^p
            \eigvec{t}
        }
        }.
    $
\end{toytheorem}
We will show that when $H$ is random and the top eigenvectors of $M$ have small
$\infty$-norm (which, for example, is the case when $M$ has block-constant
structure), the perturbation is also small.
In many natural settings, our perturbation result improves upon the classical theory by a
factor of $\nicefrac{1}{\sqrt{n}}$. 

Among the techniques used to derive the above results, we highlight the
importance of what we call the \term{Neumann trick} -- a particular expansion of
the perturbed eigenvector which diminishes the effect of components whose
interaction with $H$ is hard to bound.  To see the utility of the trick,
consider bounding some norm of the perturbation $\eigvec{1} - \peigvec{1}$:
Begin by writing $\peigvec{1}$ as $\alpha \eigvec{1} + \beta u^\perp$, where
$u^\perp$ is some unit vector orthogonal to $\eigvec{1}$. 
In the usual approach, controlling the norms of $H \eigvec{1}$ and $H u^\perp$
are crucial in bounding the size of $\eigvec{1} - \peigvec{1}$. 
In the worst case these norms are bounded by $\spectralnorm{H}$.
It turns out that $\twonorm{H \eigvec{1}}$ is often close to this worst-case
bound in practice, but that $\inftynorm{H \eigvec{1}}$ can be much smaller than
$\spectralnorm{H}$, particularly when $H$ is random. 
As a result, analyzing the interaction between $H$ and $\eigvec{1}$ often leads
to an improved perturbation bound in $\infty$-norm. 

However, while obtaining a tighter bound on $\inftynorm{H \eigvec{1}}$ is often
possible, it can be difficult to derive an improved bound on $\inftynorm{H
u^\perp}$.
Specifically, note that $\eigvec{1}$ is a fixed vector independent of the perturbation $H$, but $u^\perp$ depends on $H$. 
If $H$ is random, for instance, then $u^\perp$ is a random vector depending on
$H$ and the statistical interaction between $H$ and $u^\perp$ can be hard to
analyze.
As a result, we often cannot bound the norm of $H u^\perp$ any better than by
the spectral norm of $H$. 
%
%
The Neumann trick allows us to replace the hard-to-analyze norm of $H u^\perp$
with $\eigval_2$;
if $\eigval_2$ is smaller than
$\spectralnorm{H}$ the Neumann trick presents significant advantages over
the classical approach, as we will see.

We believe that the Neumann trick has the potential to substantially improve
eigenvector perturbation bounds in many settings. As an example, we use it to
analyze perturbations in the stochastic blockmodel and obtain much finer bounds
than provided by the classical theory. It was observed in \cite{Vu2014-lu}
that perhaps the most natural spectral algorithm for blockmodel clustering is
via low rank approximation of the matrix of edge probabilities, but that
analyzing this method is difficult.  However, with the perturbation tools
we introduce
the analysis becomes straightforward. We prove that this
natural algorithm indeed recovers the correct clustering of even very sparse graphs in
blockmodels with an arbitrary (constant) number of blocks. This result may be of independent interest. 

\header{Related work.}
Improving classical perturbation bounds has been the subject of recent interest.
In \cite{Fan2016-yy} the $\infty$-norm perturbation of singular vectors is
bounded under the assumption that $M$ is low rank and incoherent.  
Our theory does not place either of these assumptions on $M$.
Moreover, we will obtain improved bounds in some settings where
\cite{Fan2016-yy} does not apply, such as in the stochastic blockmodel.
Both \cite{Vu2010-ld} and \cite{ORourke2013-tp} consider the case where $H$ is
random and $M$ is low rank and present bounds in 2-norm which improve upon
Davis-Kahan in certain settings. In contrast, our results are for the
$\infty$-norm, we do not assume that $M$ is low rank, and $H$ needs not be
random. Furthermore, in certain settings where $M$ is low rank -- such as in the
case of the blockmodel -- the results of the aforementioned papers do
not necessarily improve upon the classical theory, while ours will.  We note
that the eigenvalue perturbation analysis in \cite{ORourke2013-tp} bears
resemblance to that presented herein, but ours will hold for
full-rank $M$ and non-random $H$. 

Also related to the present work are the theories of random graphs and matrices.
Perhaps most relevant is \cite{Erdos2011-ol}, which analyzes the spectral
statistics of Erd\H{os}-R\'enyi graphs using the Neumann trick. In contrast, we
will develop the Neumann trick into a tool for analyzing general perturbations.
Another related work is \cite{Mitra2009-pl}, which bounds the $\infty$-norm perturbation
of the top eigenvector of an Erd\H{o}s-R\'enyi graph and provides a simple
algorithm for clustering a sparse stochastic blockmodel with two communities.
However, it is not clear how to generalize this method beyond the first
eigenvector and therefore to blockmodels with $K \geq 2$ communities.  In
contrast, our method will give useful bounds on the top $K$ eigenvectors, and
our algorithm will work on models with an arbitrary (but constant) number of
communities.

The stochastic blockmodel has been well-studied; see
\cite{Abbe2017-ed} for a survey. A problem of particular
interest is that of exact recovery of the latent communities in a sparse
blockmodel. It is well-known that exact recovery is possible in the balanced
2-block model if the expected node degrees are super-logarithmic; when they are
logarithmic, recovery is possible for some choices of constant factors but not
for others. Recently, \cite{Vu2014-lu} analyzed an algorithm based on the SVD
which recovers clusters exactly all the way down to the $\log n$ degree barrier.
We will use our perturbation results to analyze a related algorithm
which exactly recovers the communities of graphs with polylogarithmic
degree. While our algorithm does not improve on that of \cite{Vu2014-lu} in
terms of performance, it is very natural and simple,  and the guarantee of its correctness is the
byproduct of our general perturbation results.
It is also easy to generalize our method to blockmodels with a super-constant
number of communities, and to models in which the block sizes scale at different
rates.

\header{Conventions and notations.}
We write $\countingset{n}$ to denote the set $\{1, \ldots, n\}$.
If $X^{(n)}$ is a sequence of random variables indexed by $n$, we say $X
= O(f(n))$ with high probability (w.h.p.) if there exists a constant $C$ such
that $\prob(\abs{X^{(n)}} \leq C f(n)) \to 1$ as $n \to \infty$. We adopt the
analogous definitions for the other asymptotic notations, such as
$\Theta(f(n))$. We assume that eigenvectors have unit 2-norm.

\section{Application: the stochastic blockmodel}
\label{sec:application}

Our perturbation results are sometimes rather technical when stated in their
full generality. Hence in this section we begin by assuming the setting of the
\term{stochastic blockmodel} -- a popular random graph model with community
structure. In this setting, our results have simpler statements and we are
able to bound the perturbation of eigenvalues and eigenvectors much more finely
than the classical theory. These finer bounds will allow us to analyze a simple
clustering algorithm whose analysis was difficult with the classical theory. Our main general perturbation results will be given in Sections \ref{sec:eigenvalues} and \ref{sec:eigenvectors}. 
First, we formally define the $K$-block model:

\begin{defn}
    An \term{$(n,K)$-stochastic blockmodel} is a pair $(z,P)$, where $z :
    \countingset{n} \to \countingset{K}$ is a surjective map and $P$ is a $K
    \times K$ symmetric matrix of rank $K$, with $P_{ij} \in [0,1]$. We call $z$
    the \term{assignment} and $P$ the \term{inter-community edge probability
    matrix}.
    The \term{edge
    probability matrix} $M$ is the $n \times n$ symmetric matrix with entries
    $M_{ij} = P_{z(i), z(j)}$.
\end{defn}

To generate a graph $G$ from a blockmodel we sample to obtain its symmetric adjacency matrix
$A = A_G$, where the upper triangular entries $(j \geq i)$ are such that
$A_{ij} \sim \operatorname{Bernoulli}(M_{ij})$ and the lower triangular entries
$(j < i)$ are constrained to $A_{ij} = A_{ji}$. 
We view the random matrix $A$ as a
perturbation of $M$ by the symmetric random matrix $H = A - M$, so that $A = M +
H$. In what follows, let the eigenvectors and eigenvalues of $M$ be $\eigvec{1},
\ldots, \eigvec{n}$ and $\eigval_1 \geq \cdots \geq \eigval_n$; similarly, let
the eigenvectors and eigenvalues of $A$ be $\peigvec{1}, \ldots, \peigvec{n}$
and $\peigval_1 \geq \cdots \geq \peigval_n$. 

We will study sequences of blockmodels
in which the expected node degree is permitted to grow sublinearly in the
size the network; this is the \term{sparse} r\'egime.  For simplicity, we
assume that every community has the same number of nodes, and that 
$P$ is shared by all blockmodels in the
sequence up to a density scaling factor of $\rho$.
More precisely, we will adopt the following setting:

\begin{minipage}{\textwidth}
\begin{minipage}{.72\textwidth}
\begin{setting}[$\rho$-sparse balanced blockmodel]
    \label{setting:sparse}
    Let $K \in \posnaturals$ and 
    fix a $K \times K$ inter-community edge probability matrix $P^{(0)}$.
    Assume for simplicity that each of the eigenvalues of $P^{(0)}$ is positive
    and unique.
    Let $\rho : \posnaturals \to (0,1]$ be such that $\rho =
    \Omega(\nicefrac{1}{n})$. For any $m \in \posnaturals$, let $n = mK$ and
    define $P^{(m)} = \rho(n) \cdot P^{(0)}$.
    Consider a sequence of blockmodels $((z^{(m)}, P^{(m)}))_{m=1}^\infty$ in
    which $z^{(m)} : \countingset{n} \to \countingset{K}$ is an assignment of
    $n$ nodes into $K$ communities such that each is of size $m$.
\end{setting}
\end{minipage}%
\hfill
\begin{minipage}{.27\textwidth}
    \centering
    \vspace{.6em}
    \begin{tabular}{ll}
        \toprule
        $\spectralnorm{H}$ & $O(\sqrt{\rho n})$\\
        $\expectation (H_{ij})^2$ & $\Theta(\rho)$\\
        $\rho$ & $\Omega(\nicefrac{1}{n})$\\
        $\eigval_t$ & $\Theta(\rho n)$\\
        $\inftynorm{\eigvec{t}}$ & $\Theta(\nicefrac{1}{\sqrt{n}})$\\
        \bottomrule
    \end{tabular}
    \captionof{table}{\label{table:qts} $t \in \countingset{K}$}
\end{minipage}
\end{minipage}

The sequence of blockmodels has associated sequences of edge probability
matrices $M^{(m)}$, random adjacency $A^{(m)}$ matrices, and so forth. 
For conciseness, we often omit the sequence index. 
We also remark that the
assumptions on the eigenvalues of $P^{(0)}$ are made to simplify the exposition; 
the following results will hold in general with minor modification.

The asymptotic behaviors of the important quantities of \Cref{setting:sparse}
are collected in \Cref{table:qts} for $t \in \countingset{K}$.  The bound on
$\spectralnorm{H}$ follows from a result of random matrix theory (see
\Cref{result:spectralnorm} in \Cref{proof:spectralnorm}).  The nonzero
eigenvalues of $M$ are the eigenvalues of $P$ scaled by $\rho n$, and hence
$\eigval_t = \Theta(\rho n)$ for any $t \in \countingset{K}$.  It can also be
shown that an eigenvector $u$ of $M$ which corresponds to a nonzero eigenvalue
is constant on each block; i.e., $z(i) = z(j) \Rightarrow u_i = u_j$.
Since each community has $m$ members, it follows from the normalization
constraint that $\inftynorm{\eigvec{t}} = \Theta(\nicefrac{1}{\sqrt{m}}) =
\Theta(\nicefrac{1}{\sqrt{n}})$.

\begin{figure}[t]
        \centering
        \begin{minipage}{.5\textwidth}
            \centering
            \input{./fig/gen/eigvalerr.pgf}
            \vspace{-1em}
            \captionof{figure}{
                \label{fig:experiment:eigenvalues}
                Empirical eigenvalue perturbations.}
        \end{minipage}%
        \begin{minipage}{.5\textwidth}
            \centering
            \input{./fig/gen/eigvecerr.pgf}
            \vspace{-1em}
            \captionof{figure}{
                \label{fig:experiment:eigenvectors}
                Empirical eigenvector perturbations.}
        \end{minipage}%
        \vspace{-.75em}
\end{figure}

\begin{wraptable}{r}{.48\textwidth}
    \vspace{-1em}
    \begin{tabular}{lll}
        \toprule
        & Classical & Ours\\
        \midrule
        $\abs{\peigval_t - \eigval_t}$ 
            & $O(\sqrt{\rho n})^\dagger$ 
            & ${O(\sqrt{\log n})}$
            \\
        $\peigval_t$ 
            & $\Theta(\rho n)$ 
            & 
            \\
        $\sin \angle(\peigvec{t}, \eigvec{t})$ 
            & $O(\nicefrac{1}{\sqrt{\rho n}})$ 
            & 
            \\
        $\twonorm{\peigvec{t} - \eigvec{t}}$ 
            & $O(\nicefrac{1}{\sqrt{\rho n}})$ 
            &
            \\
        $\inftynorm{\peigvec{t} - \eigvec{t}}$ 
            & $O(\nicefrac{1}{\sqrt{\rho n}})$ 
            & $O\left(\frac{\log^\xi n}{n \sqrt{\rho}}\right)$ 
            \\
        \bottomrule
    \end{tabular}
    \caption{\label{table:pqts} $t \in \countingset{K}$, $\quad\dagger: t \in
        \countingset{n}$}
    \vspace{-1em}
\end{wraptable}
The predictions of the classical matrix perturbation theory as applied in this
setting are collected in \Cref{table:pqts}: Weyl's theorem bounds the
eigenvalues and Davis-Kahan bounds the eigenvectors.
To assess the quality of
these bounds, the perturbation in the top eigenvalue and eigenvector of a
sequence of growing blockmodels $(K = 1, \rho = 1, P = \nicefrac{1}{2})$ was
measured; the results are shown in
\Cref{fig:experiment:eigenvalues,fig:experiment:eigenvectors}.  In the case of
eigenvalues, we see that the actual perturbation is much smaller than Weyl's
bound of $\spectralnorm{H}$. For eigenvectors, the perturbation in 2-norm is
close to the bound provided by the Davis-Kahan theorem, but the perturbation in
$\infty$-norm is much smaller than predicted.  Our general perturbation
theory will explain both of
these phenomena.  In particular, our results will imply the following:
\begin{theorem}
    \label{result:application}
    Assume \Cref{setting:sparse}; i.e., the $\rho$-sparse balanced stochastic
    blockmodel with $K \geq 1$. Suppose that $\rho = \Omega(n^{-1} \log^\epsilon
    n)$ for some $\epsilon > 2$. Let $1 < \xi <
    \nicefrac{\epsilon}{2}$.  Then there exist constants $C_1,C_2$ such that for
    any blockmodel in the sequence and all $t \in \countingset{K}$,
    with high probability as $n \to \infty$:
    \[
        \abs{\eigval_t - \peigval_t} \leq C_1 \sqrt{\log n}
        \qquad
        \text{and}
        \qquad
        \inftynorm{\eigvec{t} - \peigvec{t}} \leq 
        \frac{C_2 (\log n)^\xi}{n \sqrt{\rho}}.
    \]
\end{theorem}
These bounds are compared to their classical counterparts in \Cref{table:pqts}.
The proof of \Cref{result:application} will be given as two examples in later sections 
which serve to demonstrate how the more general perturbation results can be
applied to specific settings.
The eager reader can find the proof of the eigenvalue perturbation in
\Cref{sec:eigenvalues},
\Cref{example:proofpt1} and the proof for eigenvectors in
\Cref{sec:eigenvectors},
\Cref{example:proofpt2}.

The fact that the eigenvectors of the blockmodel can be recovered to such
precision suggests the very 
simple clustering algorithm in
\Cref{alg:cluster}.
The method first computes a rank-$K$ approximation
$\hat{M}$ 
\begin{wrapfigure}{r}{0.54\textwidth}
\vspace{-2em}
\begin{minipage}{.54\textwidth}
\begin{algorithm}[H]
    \caption{%
        \label{alg:cluster}
        Blockmodel clustering}
    \begin{algorithmic}
        \REQUIRE Adjacency matrix $A$, $\tau \in \reals^+$, $K \in \posnaturals$
        \STATE $\peigval_{s_1}, \ldots, \peigval_{s_K} \gets $ top $K$
            eigvals of $A$ by magnitude
        \STATE $\peigvec{s_1}, \ldots, \peigvec{s_K} \gets$ corresponding
            eigvecs of $A$
            \STATE $\hat{M} \gets \sum_{i=1}^K 
            \peigval_{s_i} \peigvec{s_i} \otimes \peigvec{s_i}$
        \STATE $E \gets \{(i,j) : \inftynorm{\hat{M}_i - \hat{M}_j} < \tau\}$
        \STATE $G \gets$ graph with node set $\countingset{n}$, edge set $E$
        \RETURN connected components of $G$
    \end{algorithmic}
\end{algorithm}
\end{minipage}
\vspace{-2em}
\end{wrapfigure}
of $M$ using the top $K$ 
eigenvectors of $A$ 
ordered by the magnitude of their eigenvalues. It
then clusters together all columns which are within a threshold $\tau$ in
$\infty$-norm.
Intuitively, the correctness of the algorithm relies on the assumption that
$\hat{M}$ is close to $M$ entrywise.  The following \namecref{result:entrywise}
proves that this is indeed the case; the algorithm's consistency is a corollary.
The proofs of both are located in \Cref{sec:blockmodel_proofs}.
\begin{restatable}{lemma}{restateentrywise}
    \label{result:entrywise}
    Suppose that the assumptions of \Cref{result:application} hold. Define
    $\hat{M}$ as in \Cref{alg:cluster}. Then $\maxnorm{\hat{M} - M} =
    O(\sqrt{\nicefrac{\rho}{n}} \cdot \log^\xi n)$ with high probability.
\end{restatable}
\begin{restatable}[Consistency of \Cref{alg:cluster}]{theorem}{restateconsistency}
    \label{result:consistency}
    Suppose that the assumptions of \Cref{result:application} hold.
    Let $\tau = \omega(\sqrt{\nicefrac{\rho}{n}} \cdot \log^\xi n)$
    and $\tau = o(\rho)$.
    Define $\Gamma = \{ z^{-1}(k) \}_{k = 1}^K$ to be the partition of
    $\countingset{n}$ into the ground-truth communities, and let $\hat{\Gamma}$
    be the clustering returned by \Cref{alg:cluster} with inputs $A$, $\tau =
    \tau(n)$, and $K$.  Then $\prob(\text{communities recovered exactly}) =
    \prob(\Gamma = \hat{\Gamma}) \to 1$ as $n \to \infty$.
\end{restatable}

\begin{remark}
    \label{remark:svd}
    It was noted in \cite{Vu2014-lu} that \Cref{alg:cluster} is very natural,
    but difficult to analyze. With the perturbation tools presented in this
    paper, however, the analysis becomes straightforward. One reason for this is
    that the classical perturbation theory 
    only provides a useful bound on
    the Frobenius norm of $M - \hat{M}$. It turns out that this is not
    sufficient for exact recovery. 
    Our theory instead provides a
    tight bound on $\maxnorm{M - \hat{M}}$, which \emph{is}
    sufficient. See \Cref{sec:consremark} for details.
\end{remark}

\section{Eigenvalue perturbation}
\label{sec:eigenvalues}

In this section we derive an eigenvalue perturbation bound that is stated in
terms of the interaction between the perturbation matrix $H$ and the
eigenvectors of the base matrix $M$.  We will see that in many cases,
particularly when $H$ is random, this bound is much tighter than Weyl's.
The perturbation for eigenvectors is much more sophisticated to analyze, and will be given in Section \ref{sec:eigenvectors}.  

\newcommand{\allsubspaces}{\mathcal{V}}
\newcommand{\subspace}{V}

To see how incorporating the interaction between $H$ and the eigenvectors of $M$
may lead to improved bounds, consider the following informal analysis of the
perturbation in the first eigenvalue. As usual, let $M$ and $H$ be $n \times n$
and symmetric. The eigenvalues and eigenvectors of $M$ are $\eigval_1 \geq
\cdots \geq \eigval_n$ and $\eigvec{1}, \ldots, \eigvec{n}$, and the
eigenvalues/vectors of $M + H$ are $\peigval_1 \geq \cdots \geq \peigval_n$ and
$\peigvec{1}, \ldots, \peigvec{n}$.  We have $\eigval_1 = \langle \eigvec{1}, M
\eigvec{1} \rangle$ and $\peigval_1 = \langle \peigvec{1}, (M+H)\, \peigvec{1}
\rangle$.  Intuitively, if $\peigvec{1}$ is close to $\eigvec{1}$ then
$\peigval_1 \approx \langle \eigvec{1}, (M+H)\, \eigvec{1} \rangle$; hence
$\peigval_1 - \eigval_1 \approx \langle \eigvec{1}, H \eigvec{1} \rangle$. In
the worst case $\abs{\langle \eigvec{1}, H \eigvec{1}\rangle}$ can be as large
as $\spectralnorm{H}$ and we recover Weyl's bound.  However, 
$\abs{\langle \eigvec{1}, H \eigvec{1} \rangle}$ could be much smaller than
$\spectralnorm{H}$. For example, suppose that the entries of $H$ are independent
random variables with standard Gaussian distribution.  Then $\langle \eigvec{1},
H \eigvec{1} \rangle$ is the sum of centered and independent random variables
and therefore concentrates around zero. In this case the spectral norm of $H$ is
$O(\sqrt{n})$ while $\abs{\langle \eigvec{1}, H \eigvec{1} \rangle}$ is much
smaller at $O(\sqrt{\log n})$; this leads to an $O(\sqrt{\log n})$
bound on the eigenvalue perturbation instead of Weyl's bound of $O(\sqrt{n})$.

\label{sec:eigvalresult}

We now formalize this argument. 
We use the following well-known characterization of eigenvalues.

\begin{theorem}[Courant-Fischer-Weyl min-max/max-min principles \cite{Horn2012-cm}]
    \label{fact:min_max}
    Let $B$ be an $n \times n$ symmetric matrix with eigenvalues $\mu_1 \geq
    \ldots \geq \mu_t \geq \ldots \mu_n$. For any $d \in \{1, \ldots,
    n\}$, write $\allsubspaces_d$ for the set of $d$-dimensional subspaces of
    $\mathbb R^n$. Then
    \begin{align*}
        \mu_t 
        \quad
        = 
        \quad
            \min_{\subspace \in \allsubspaces_{n -t + 1}}
            \;
            \max_{\substack{
                x \in \subspace\\
                \|x\| = 1
                }}
            \;
            \left\langle
                x, B x
            \right\rangle
        \quad
        = 
        \quad
            \max_{\subspace \in \allsubspaces_{t}}
            \;
            \min_{\substack{
                x \in \subspace\\
                \|x\| = 1
                }}
            \;
            \left\langle
                x, B x
            \right\rangle.
    \end{align*}
\end{theorem}

We will use the max-min principle to get a lower bound on the perturbed
eigenvalue and the min-max principle to obtain an upper bound. We prove
the lower bound here to provide intuition:

\newcommand{\ixupper}{T}
\newcommand{\ixlower}{s^{\downarrow}}
\newcommand{\dupper}{\ixupper}
\newcommand{\smallh}{h}
\newcommand{\topeigspan}{
    \Vecspan{
    (
        \{\eigvec{1}, \ldots, \eigvec{\ixupper}\}
    )}
}
\newcommand{\boteigspan}{
    \Vecspan{
    (
        \{\eigvec{\ixlower}, \ldots, \eigvec{n}\}
    )}
}

\begin{theorem}[Eigenvalue lower bound]
    \label{result:eigvallower}
    Let $\ixupper \in \countingset{n}$ and $\smallh$ be such that
    $\abs{\left\langle x, H x \right\rangle} \leq \smallh$ for all $x \in
    \topeigspan$.
    Then $\peigval_t \geq \eigval_t - h$ for all $t \leq \ixupper$.
\end{theorem}

\begin{proof}
    The max-min principle tells us that
    \[
        \peigval_t 
        = 
                \max_{\subspace \in \allsubspaces_t} 
                \min_{\substack{x \in \subspace\\\|x\| = 1}}
                \left\langle
                    x, (M + H) x
                \right\rangle
                .
    \]
    Let $\subspace^* = \topeigspan$. Then the above is
    lower-bounded by:
    \[
                \min_{\substack{x \in \subspace^*\\ \|x\| = 1}}
                \left\langle
                    x, (M + H) x
                \right\rangle
            \geq
                \min_{\substack{x \in \subspace^*\\ \|x\| = 1}}
                \left\langle
                    x, M x
                \right\rangle
                - 
                \max_{\substack{x \in \subspace^*\\ \|x\| = 1}}
                \left\langle
                    x, H x
                \right\rangle
                .
    \]
    The first term is minimized by taking $x = \eigvec{t}$, such that
    $
        \left\langle x, M x \right\rangle 
        = 
        \left\langle \eigvec{t}, M \eigvec{t} \right\rangle 
        = \eigval_t
    $.
    The magnitude of the second term is bounded by $h$.
\end{proof}

The proof of the following upper bound is 
more involved and is therefore located in
\Cref{proof:eigvalupper}.

\begin{restatable}[Eigenvalue upper bound]{theorem}{restateeigvalupper}
    \label{result:eigvalupper}
    Let $\ixupper \in \countingset{n}$ and $\smallh$ be such that
    $\abs{\left\langle x, H x \right\rangle} \leq \smallh$ for all $x \in
    \topeigspan$.
    Let $t \leq \ixupper$ and
    suppose that $\eigval_t - \eigval_{\ixupper + 1} > 2 \spectralnorm{H} -
    \smallh$. Then:
    \[
        \peigval_t \leq
        \eigval_t
        +
        \smallh
        +
        \frac{
            \spectralnorm{H}^2
        }{
            \eigval_t
            -
            \eigval_{\ixupper + 1}
            +
            \smallh
            -
            \spectralnorm{H}
        }
        .
    \]
\end{restatable}

\newcommand{\topsmallh}{\smallh^\uparrow}
\newcommand{\botsmallh}{\smallh^\downarrow}

Similar lower and upper bounds can be obtained for eigenvalues at the bottom of
the spectrum by negating $M$ and $H$. For ease of reference, the statement
of that result is located in \Cref{proof:eigval}.

\header{Interactions with random perturbations.}
\Cref{result:eigvallower,result:eigvalupper} show that a tighter
bound on eigenvalue perturbations can be obtained when $\abs{\langle x, H x
\rangle} \ll \spectralnorm{H}$ for any $x$ in a subspace spanned by the top (or
bottom) eigenvectors of $M$. We now show that this is often the case when 
$H$ is random. The following is an application of the usual
Hoeffding inequality; the proof is located in \Cref{proof:hoeffding}.

\begin{restatable}{lemma}{restatehoeffding}
    \label{result:hoeffding}
    Let $u, v$ be any two fixed unit vectors in $\reals^n$. Let $H$ be an $n
    \times n$ symmetric random matrix with independent entries along the
    upper-triangle such that for all $j \geq i$, $\expectation H_{ij} = 0$ and
    $H_{ij}$ is sub-Gaussian with parameter $\sigma_{ij} \leq \sigma$.
    Then
    $
        \prob(
            \abs{\langle u, H v \rangle}
            \geq \gamma
        )
        \leq
        2 
        \exp \{-\gamma^2 / (8 \sigma^2)\}.
    $
\end{restatable}

\Cref{result:hoeffding} applies generally to many types of random
perturbation, including Gaussian noise and Bernoulli noise, as well as the
random graph noise encountered in the stochastic blockmodel example in
\Cref{sec:application}.  
We typically integrate the \namecref{result:hoeffding} with \Cref{result:eigvallower,result:eigvalupper} 
in the following way:
We first partition the spectrum into a top (large positive) and the
remainder (small positive and negative) by choosing $\ixupper \in
\countingset{n}$ such that $\eigval_\ixupper \gg \eigval_{\ixupper + 1}$.
We then apply \Cref{result:hoeffding} to argue that $\abs{\langle \eigvec{i}, H
\eigvec{j} \rangle}$ is small $(\leq h)$ with high probability for any indices
$i, j \leq \ixupper$.
It follows that $\abs{\langle x, H x\rangle} \leq \ixupper h$ for any unit
vector $x$ lying within the span of the top $\ixupper$ eigenvectors of $M$; see
\Cref{result:spanh} in \Cref{proof:spanh} for a proof.
To bound the negative eigenvalues we negate $M$ and $H$ and repeat the above
process. 

\begin{example}
\begin{proof}[Proof of eigenvalue perturbation bound stated in
        \Cref{result:application}]
    \label{example:proofpt1}
    To demonstrate the application of our eigenvalue perturbation results,
    we will prove that in the blockmodel setting assumed in
    \Cref{result:application}, $\abs{\peigval_t - \eigval_t} \leq C \sqrt{\log
    n}$ for $t \in \countingset{K}$.
    We begin by applying \Cref{result:hoeffding}.  Since
    $\eigval_{K+1},\ldots,\eigval_n$ are zero, we naturally choose $T = K$ such
    that $\eigval_\ixupper - \eigval_{\ixupper+1} = \eigval_\ixupper =
    \Theta(\rho n)$.  Each entry along the diagonal and in the upper triangle of
    $H$ is bounded and hence sub-Gaussian with a variance
    parameter upper-bounded by some constant
    $\sigma$. Choosing $\gamma = \sqrt{C \log n}$ in
    \Cref{result:hoeffding}, we find that $\abs{\langle \nzeigvec{i}, H
    \nzeigvec{j}\rangle} \leq \sqrt{C \log n}$ for all $i, j \leq \ixupper$
    w.h.p. Thus $\abs{\langle x, H x \rangle} \leq T \sqrt{C \log n}
    = O(\sqrt{\log n})$ for all $x \in \Vecspan(\{\nzeigvec{s} : s \leq
    \ixupper\})$. We therefore bound $h$ by $O(\sqrt{\log n})$ w.h.p.\ in
    \Cref{result:eigvallower,result:eigvalupper}.
    It follows from the assumption that $\rho =
    \omega(n^{-1} \log n)$ and the results in \Cref{table:qts} that $\nzeigval_t
    + h - \spectralnorm{H}$ is dominated by $\nzeigval_t$, and therefore
    $\Theta(\rho n)$.  Hence the second term in \Cref{result:eigvalupper} is
    $O(\nicefrac{\spectralnorm{H}^2}{\nzeigval_t}) = O(1)$, and both the upper
    and lower bounds are dominated by $h = O(\sqrt{\log n})$.
\end{proof}
\end{example}

\section{Eigenvector perturbation}
\label{sec:eigenvectors}

We now study how a tigher bound on eigenvector perturbations might be achieved by analyzing the interaction between $H$ and eigenvectors of $M$. 
Proofs of results in this section are rather technical and mostly in appendices. 
To build intuition, we make a series of simplifying assumptions; 
our formal theory will be much more general.
First suppose that all eigenvalues of $M$ are non-negative and that 
$\eigval_1 \gg \eigval_2$.
By writing
$\peigvec{1}$ as $\alpha \eigvec{1} + \beta u^\perp$ for some unit vector
$u^\perp$ orthogonal to $\eigvec{1}$ and using the definition of an eigenvector,
we obtain:
$
    \peigvec{1}
    =
        \peigval_1^{-1}
        \left(
            M + \rv{H}
        \right)
        \,
        \peigvec{1}
        \nonumber
    =
        \peigval_1^{-1}
        \left(
            \alpha \eigval_1 \eigvec{1}
            +
            \beta M u^\perp
            +
            \alpha H \eigvec{1}
            +
            \beta H u^\perp
        \right).
$
Note that $\twonorm{M u^\perp} \leq \twonorm{M \eigvec{2}} = \eigval_2 \ll
\eigval_1$. If $\eigval_2$ is sufficiently
small, the contribution of $\beta M u^\perp$ to $\peigvec{1}$ is negligible.
Assume that this is so, that $\peigval_1 \approx \eigval_1$, and that $\alpha
\approx 1$ such that $\beta \ll 1$.  Then
$
    \label{eqn:decomp}
    \eigvec{1} - \peigvec{1}
    \approx
    \eigval_1^{-1} (
        H \eigvec{1} 
        + 
        \beta H u^\perp
    ).
$
Therefore we see that to bound the norm of the perturbation it suffices
to control the norms of $H \eigvec{1}$ and $H u^\perp$.

The classical approach is to bound these quantities by the spectral norm of $H$.
For instance, to derive a bound in 2-norm we observe that $\twonorm{H
\eigvec{1}} \leq \spectralnorm{H}$ and that $\twonorm{\beta H u^\perp} \leq
\spectralnorm{\beta H}$, and therefore $\twonorm{\peigvec{1} - \eigvec{1}}
\lesssim \eigval_1^{-1} \spectralnorm{H}$.  Furthermore, since the 2-norm
upper-bounds the $\infty$-norm, we get a bound of $\inftynorm{\peigvec{1} -
\eigvec{1}} \lesssim \eigval_1^{-1} \spectralnorm{H}$ ``for free''. However, the
spectral norm does not utilize information about the interaction between $H$ and
$M$. Our hope is that by analyzing this interaction, tighter bounds on the norms
of $H \eigvec{1}$ and $H u^\perp$ might be obtained.

In particular, consider 
a random, centered $H$ and
$\eigvec{1}$ (which is independent of $H$). Unfortunately, $\twonorm{H \eigvec{1}}$ is
typically on the same order as $\spectralnorm{H}$ and analyzing the
interaction does not improve the bound.  On the other hand,
$\inftynorm{H \eigvec{1}}$ is often much smaller than $\spectralnorm{H}$ and
analyzing the interaction leads to much tighter bounds. To see
why, note that
$
    \|\rv{H} \eigvec{1}\|_2^2
    =
            \textstyle
            \sum_{i = 1}^n
            (
            \sum_{j = 1}^n
                \rv{H}_{ij}
                \eigvec{1}_j
            )^2
            .
$
As the summand of the outer sum is squared and thus non-negative, it does
not concentrate around zero. In contrast, 
the sum in
$
    \abs{\textstyle[\rv{H} \eigvec{1}]_i}
    =
        \abs{
            \textstyle
            \sum_{j = 1}^n
                \rv{H}_{ij}
                \eigvec{1}_j
        }
$
\emph{does} concentrate around zero, and is often much less than the
worst-case bound of $\spectralnorm{H}$.
For example, if $H$ is the random Gaussian matrix described above then $[H
\eigvec{1}]_i$ is on the order of one, and a union bound over the $n$ entries
results in a high-probability bound of $\inftynorm{H \eigvec{1}} \leq
\sqrt{\log n}$. On the other hand, $\spectralnorm{H} = O(\sqrt{n})$.

In this case and in others, $\inftynorm{H \eigvec{1}}$ can be bounded to be much
smaller than $\spectralnorm{H}$. Can a similar analysis be used to show that
$\inftynorm{H u^\perp}$ is much smaller than $\spectralnorm{H}$? It turns out
that this is difficult for a subtle reason: while $\eigvec{1}$ is fixed,
$u^\perp$ depends on the perturbation. When $H$ is random,
$u^\perp$ is also random and statistically dependent on $H$.
As such, the interaction between $H$ and $u^\perp$ is often difficult to
analyze, and we must resort to using the worst-case bound of $\inftynorm{H
u^\perp} \leq \spectralnorm{H}$, giving:
\begin{equation}
    \label{eqn:decompinfty}
    \inftynorm{\eigvec{1} - \peigvec{1}}
    \lesssim 
    \peigval^{-1}
    \left(
        \inftynorm{H \eigvec{1}}
        +
        \spectralnorm{\beta H}
    \right).
\end{equation}
In many cases $\spectralnorm{\beta H}$ is small enough that it is dominated
by our bound on $\inftynorm{H \eigvec{1}}$ and we have
$
    \inftynorm{\eigvec{1} - \peigvec{1}}
    \lesssim \peigval^{-1}\inftynorm{H \eigvec{1}}
$.
For example, it can be shown that in the sparse stochastic blockmodel described
in \Cref{setting:sparse}, $\inftynorm{H \eigvec{1}} = O(\sqrt{\rho \log n})$
w.h.p., while $\spectralnorm{\beta H} = O(1)$. Therefore, if $\rho =
\Omega(\nicefrac{1}{\log n})$ (recall we allow $\rho$ to be much smaller to be $\omega(\log n / n)$), the bound on $\inftynorm{H \eigvec{1}}$ dominates
and we have
$\inftynorm{\peigvec{1} - \eigvec{1}} = O(\rho^{\nicefrac{-1}{2}} \, n^{-1}
\sqrt{\log n})$. Comparing this to the trivial bound of
$O(\nicefrac{1}{\sqrt{\rho n}})$ implied by Davis-Kahan, we see that
analyzing the interaction leads to a $\tilde{O}(\nicefrac{1}{\sqrt{n}})$
improvement over the classical theory.

\header{The Neumann trick.}
There are important settings, however, in which using the spectral norm
to bound $H u^\perp$ is sub-optimal; for instance, in the
blockmodel described above when $\rho = o(\nicefrac{1}{\log n})$ (recall we allow $\rho$ to be much smaller to $\omega(\log n/n)$). 
In this sparser r\'egime, $\spectralnorm{\beta H} = O(1)$ dominates our bound on
$\inftynorm{H \eigvec{1}}$ and we find that $\inftynorm{\peigvec{1} -
\eigvec{1}} = O(\nicefrac{1}{\peigval_1}) = O(\nicefrac{1}{\rho n})$, which is
not tight.
In general, if
$\inftynorm{H \eigvec{1}}$ can be bounded to be much smaller than
$\spectralnorm{\beta H}$, the latter term dominates \Cref{eqn:decompinfty}. 
Therefore, while the simple approach described in the previous section
improves upon the classical bound, the presence of the
hard-to-control $H u^\perp$ limits its effectiveness. 

It turns out that we can often obtain a better bound by applying
what we call the \term{Neumann trick},
which we now describe for $\peigvec{1}$.
From the definition of an eigenvector, we have $(M + H)\peigvec{1} = \peigval_1
\peigvec{1}$, which implies $(\peigval_1 - H)\peigvec{1} = M \peigvec{1}$.
If $\peigval_1$ is not an eigenvalue of $H$ we may invert
$(\peigval_1 - H)$ to obtain $\peigvec{1} = {\peigval_1}^{-1} (I -
H/\peigval_1)^{-1} M \peigvec{1}$. 
Expanding the inverse in a Neumann series and decomposing $\peigvec{1}$ as
above, we find:
$
    {
    \textstyle
        \peigvec{1} = 
        \peigval_1^{-1}
        \sum_{p \geq 0}
        \left(
            \nicefrac{H}{\peigval_1}
        \right)^p
        \left[
            \alpha \eigval_1 \eigvec{1} + \beta M u^\perp
        \right].
    }
    $
Assuming that $\alpha \approx 1$ and $\eigval_1 \approx \peigval_1$, we have:
\begin{equation}
    {\textstyle
    \eigvec{1} - \peigvec{1} 
    \approx
        \left[
            \sum_{p \geq 1}
            \left(
                \nicefrac{H}{\peigval_1}
            \right)^p
            \eigvec{1}
        \right]
        +
        \left[
            \frac{\beta}{\peigval_1}
            \sum_{p \geq 0}
            \left(
                \nicefrac{H}{\peigval_1}
            \right)^p
            M u^\perp
        \right].
    }
\end{equation}
If the series involving $u^\perp$ converges,
it is dominated by its first term: $M
u^\perp$. Since $u^\perp$ lies in the subspace orthogonal to
$\eigvec{1}$, $\twonorm{M u^\perp}$ is
upper-bounded by $\eigval_2$, and hence so is $\inftynorm{M u^\perp}$.
Hence:
\begin{equation}
    {\textstyle
    \label{eqn:neumanntrick}
    \inftynorm{
        \eigvec{1} - \peigvec{1} 
    }
    \lesssim
        \inftynorm*{
            \sum_{p \geq 1}
            \left(
                \nicefrac{H}{\peigval_1}
            \right)^p
            \eigvec{1}
        }
        +
        \nicefrac{\abs{\beta} \eigval_2}{\peigval_1}
        .
    }
\end{equation}
Thus the contribution of $u^\perp$ is bounded here by $\peigval_1^{-1}
\abs{\beta} \eigval_2$. Comparing this to the previous result of
\Cref{eqn:decompinfty} in which the contribution of $u^\perp$ was bounded by
$\peigval_1^{-1} \abs{\beta} \cdot \spectralnorm{H}$, we see that the Neumann
trick permits us to replace $\spectralnorm{H}$ with the
top eigenvalue corresponding to the subspace orthogonal to $\eigvec{1}$. The
tradeoff is that we must now analyze the interaction between all powers of $H$
and $\eigvec{1}$ in order to bound the first term in \Cref{eqn:neumanntrick}. 

The Neumann trick allows us to tighten the eigenvector perturbation bound in the
sparse stochastic blockmodel discussed above. We have seen that the first
approach of \Cref{eqn:decompinfty} leads to a bound of $\inftynorm{\peigvec{1} -
\eigvec{1}} = O(\nicefrac{1}{\rho n})$ when $\rho = o(\nicefrac{1}{\log n})$. 
Now if we use Neumann trick, we can 
show that the norm of the series in \Cref{eqn:neumanntrick} is
$O(\log^\xi n / (\sqrt{\rho}n))$, 
where $\xi > 1$.
Assume the blockmodel has only one block (for multiple blocks we will use the more general results in Theorem \ref{result:neumanntrick}).
Then $\eigval_2 = 0$ and the
second term in \Cref{eqn:neumanntrick} disappears.  We thus have 
$\inftynorm{\peigvec{1} - \eigvec{1}} = O(\log^\xi n / (\sqrt{\rho}n))$, 
which significantly outperforms $O(\nicefrac{1}{\rho n})$ in this
sparse r\'egime (where $\rho = o(1/\log n)$). 

We now formally state the general \term{Neumann trick}. 
See \Cref{proof:neumanntrick} for the proof.

\newcommand{\tailboundraw}{\zeta}
\newcommand{\tailbound}[1]{\tailboundraw^{(#1)}}
\newcommand{\normalize}{\gamma}
\newcommand{\eigerr}{\epsilon}
\newcommand{\ix}{\alpha}
\newcommand{\elemboundraw}{\chi}
\newcommand{\elembound}[1]{\elemboundraw^{(#1)}}
\newcommand{\gap}{\delta}

\begin{restatable}[Neumann trick]{theorem}{restateneumanntrick}
    \label{result:neumanntrick}
    Fix a $t \in \countingset{n}$. Suppose that $\spectralnorm{H} <
    \abs{\peigval_t}$. 
    Then:
    \[
        {\textstyle
        \peigvec{t} =
        \sum_{s=1}^n
                \nicefrac{\eigval_s}{\peigval_t}
                \cdot
                \left\langle
                    \peigvec{t},
                    \eigvec{s}
                \right\rangle
            \sum_{p \geq 0} 
            \left(
                \nicefrac{H}{\peigval_t}
            \right)^p
                \eigvec{s}.
        }
    \]
\end{restatable}
Observe that the contribution of $\eigvec{s}$ is filtered by its eigenvalue,
$\eigval_s$. In the special case when $M$ is rank-$K$, $\peigvec{t}$ is
expressed totally in terms of $\eigvec{1}, \ldots, \eigvec{K}$.
The Neumann trick can be used in combination with Weyl's theorem and the
Davis-Kahan theorem to obtain a tighter bound on the elementwise perturbation of
eigenvectors.

The following theorem states the result in its full generality, where $M$ may be
full-rank with non-distinct eigenvalues. Its proof in \Cref{proof:dkperturb} is
a corollary of \Cref{result:neumannperturb} in \Cref{proof:neumannperturb}.
Let $u_\alpha$ denote the $\alpha$-th entry of vector $u$.

\newcommand{\loweigval}{\eigval^*_t}%
\begin{restatable}{theorem}{restatedkperturb}
    \label{result:dkperturb}
    For any $s \in \countingset{n}$, let $\Lambda_s = \{ i : \eigval_i =
    \eigval_s \}$. 
    Define $d_s = \abs*{\Lambda_s}$, and let the gap be defined
    as $\gap_s = \min_{i \not \in \Lambda_s} \abs*{\eigval_s - \eigval_i}$.
    Let $\Delta^{-1}_{s,t} = \min \{ d_i / \gap_i \}_{i \in \{s,t\}}$.
    Define $\loweigval = \abs{\eigval_t} - \spectralnorm{H}$.
    There exists an orthonormal set of eigenvectors $\eigvec{1}, \ldots,
    \eigvec{n}$ satisfying $M \eigvec{s} = \eigval_s \eigvec{s}$ such that for
    all $t \in \countingset{n}$:
    \begin{align}
        \abs*{\peigvec{t}_\ix - \eigvec{t}_\ix}
        &\leq\nonumber
        {\textstyle
            \abs*{\eigvec{t}_\ix}
            \cdot
            \left(
                8 d_t 
                \left[
                    \frac{\spectralnorm{H}}{\gap_t}
                \right]^2
            +
            \frac{\spectralnorm{H}}{\loweigval}
            \right)
            +
            \left(
                \frac{\abs*{\eigval_t}
                    }{
                        \loweigval
                    }
            \right)^2
            \cdot
            \tailboundraw_\ix(\eigvec{t}; H, \eigval_t)
        }
            \\
        &\qquad\label{eqn:dkperturb}
            {\textstyle
            +
            \frac{
                2\sqrt{2}\cdot \spectralnorm{H}
            }{
                \loweigval
            }
            \sum_{s \neq t}
                \frac{\abs{\eigval_s}}{\Delta_{s,t}}
                \left[
                    \abs{\eigvec{s}_\ix}
                    +
                        \frac{\abs{\eigval_t}}{\loweigval}
                    \cdot
                    \tailboundraw_\ix(\eigvec{s}; H, \eigval_t)
                \right],
            }
    \end{align}
    where $\tailboundraw(u; H,\eigval)$ is the $n$-vector whose $\ix$th entry is
    defined to be
    $
        {\textstyle
            \tailboundraw_\ix(u; H, \eigval)  = 
        \abs*{
        \left[
        \sum_{p \geq 1} \left(
            \frac{H}{\eigval} 
        \right)^p
        u
        \right]_\ix
        }.
        }
        $
\end{restatable}

\header{Interactions with random perturbations.}
The interaction between the eigenvectors of $M$ and the perturbation $H$ appears
in \Cref{result:dkperturb} through $\tailboundraw$; in many applications
$\tailboundraw$ will dominate the bound. It turns out that when $H$ is
random and the eigenvectors of $M$ have small $\infty$-norm, $\tailboundraw$ is
also small. The following result makes this precise. 
See \Cref{proof:boundzeta} for the proof.

\begin{restatable}{theorem}{restateboundzeta}
    \label{result:boundzeta}
    Let $H$ be an $n \times n$ symmetric random matrix with independent entries
    along the diagonal and upper triangle satisfying $\expectation
    H_{ij} = 0$. Suppose $\gamma$ is such that
    $
        \expectation \abs{H_{ij}/\gamma}^p \leq \nicefrac1n
    $ for all $p \geq 2$.
    Choose $\xi > 1$ and $\kappa \in (0,1)$. 
    Let $\eigval \in \reals$ and suppose that $\gamma < \eigval (\log n)^\xi$
    and $\eigval > \spectralnorm{H}$.
    Fix $u \in \reals^n$. Then:
    with probability
    $
        1 
        - 
            n^{-\frac{1}{4} (\log_b n)^{\xi-1} (\log_b
            e)^{-\xi} + 1},
    $
    where 
    $b = \left(\frac{\kappa + 1}{2}\right)^{-1}$.
    \begin{equation}
        \label{eqn:boundzeta}
            \inftynorm*{
                \textstyle
                \sum_{p \geq 1} 
                \left(\frac{H}{\eigval}\right)^p
                u
            }
        \leq
            \frac{\gamma (\log n)^\xi}{\eigval - \gamma (\log n)^\xi}
            \cdot \inftynorm{u}
            +
            \frac{
                \spectralnorm{H/\eigval}^{\lfloor 
                    \frac{\kappa}{8}(\log n)^\xi + 1
                \rfloor}
            }{
                1 - \spectralnorm{H/\eigval}
            }
            \cdot
            \twonorm{u}
            .
    \end{equation}
\end{restatable}

In some cases it is possible to achieve a finer bound on individual entries of
$\tailboundraw$ as opposed to  $\inftynorm{\tailboundraw}$.
The analogous \Cref{result:boundzetamag,result:boundzetablock} are given in
\Cref{proof:boundzetamag}.

\begin{example}
    \label{example:proofpt2}
    \begin{proof}[Proof of eigenvector perturbation bound stated in
        \Cref{result:application}]
    Consider again the setting of \Cref{result:application}. We will use
    \Cref{result:dkperturb,result:boundzeta} to derive the bound of
    $\inftynorm{\eigvec{t} - \peigvec{t}} = O(\rho^{-\nicefrac12} \, n^{-1}
    \log^\xi n )$ w.h.p.\ for all $t \in \countingset{K}$.
    \newcommand{\lownzeigval}{\nzeigval^*}

    First note that all but $K - 1$
    terms of the sum in \Cref{eqn:dkperturb} vanish due to $\eigval_s$ being
    zero; only the terms corresponding to $s \in \countingset{K}$ remain.
    Referring to \Cref{table:qts}, we find that for any $s \in
    \countingset{K}$:
    $\spectralnorm{H} = O(\sqrt{\rho n})$,
    $
        \lownzeigval_t = \Theta(\rho n),
    $
    $
        \gap_t = \Theta(\rho n),
    $
    and
    $
        \inftynorm{\nzeigvec{s}} = \Theta(\nicefrac{1}{\sqrt{n}}).
    $
    Substituting these bounds into \Cref{eqn:dkperturb} and assuming that
    $Z$ is an upper bound for $\inftynorm{\tailboundraw(\nzeigvec{s}; H,
    \nzeigval_t)}$ for all $s \in \countingset{K}$, we see that the first term
    in \Cref{result:dkperturb} is $O(n^{-1} \rho^{-\nicefrac12})$, the second
    term is $O(Z)$ and the third term is $O(n^{-1} \rho^{\nicefrac{-1}2} +
    (\rho n)^{\nicefrac{-1}{2}} Z)$.  Therefore
    $
        {\textstyle
            \inftynorm{\eigvec{t} - \peigvec{t}}
        =
        O(n^{-1} \rho^{-\nicefrac12} +  Z)
        }
    $
    with high probability.

    We now bound $Z$.
    It can be shown that there exists a constant $C$ such that setting $\gamma =
    C \sqrt{\rho n}$ results in $\expectation \abs{H_{ij}/\gamma}^k \leq
    \nicefrac1n$ for all $k \geq 2$ w.h.p.
    Since $\rho = \omega(n^{-1} \log^\epsilon n)$ and
    $\epsilon > 2 \xi$ by assumption, $\nzeigval_t - \gamma (\log n)^\xi$ is
    dominated by $\nzeigval_t$ and so the first term in \Cref{eqn:boundzeta} is
    $O(\nzeigval_t^{-1} \,\gamma \cdot\inftynorm{\nzeigvec{s}} \cdot \log^\xi n)
    = O(\log^\xi n / (\sqrt{\rho}n))$. 
    Next, we have $\spectralnorm{H/\nzeigval_t} = O(\nicefrac{1}{\sqrt{\rho n}})$
    w.h.p.
    Since $\kappa$ and $\xi$ are fixed constants, the exponent $\frac{\kappa}{8}
    \log^\xi n$ is unbounded as $n \to \infty$ and hence the second term is
    dominated by the first. 
    Using this result
    as $Z$, we find that 
    $\inftynorm{\nzeigvec{t} - \nzpeigvec{t}} =
    O(\log^\xi n /(\sqrt{\rho} n))$
    w.h.p.
    \end{proof}
\end{example}

\textbf{Acknowledgements.} This work was supported by NSF grants 
CCF-1422830,
DMS-1547357, \&
RI-1550757.
It was done in part while the authors were visiting the Simons Institute
for the Theory of Computing.

\newpage
\bibliography{citations}

\normalsize
\newpage
\appendix

\section{Regarding the consistency of \Cref{alg:cluster}}
\label{sec:blockmodel_proofs}

\subsection{Proof of \Cref{result:entrywise}}

We now prove the following result which was originally stated in
\Cref{sec:application}:

\restateentrywise*{}

\begin{proof}
    Recall that we define $\hat{M}$ to be the rank-K approximation of $M$ using
    the top $K$ eigenvectors of $A$ in magnitude. Let $s_1, \ldots, s_K$ be such
    that $\abs{\eigval_{s_1}} \geq \abs{\eigval_{s_2}} \geq \cdots \geq
    \abs{\eigval_{s_K}}$ are the top $K$ eigenvalues of $M$ in absolute value.
    We first argue that $\abs{\peigval_{s_1}} \geq \abs{\peigval_{s_2}} \geq
    \cdots \geq \abs{\peigval_{s_K}}$ are the top eigenvalues of $A$ in absolute
    value with high probability as $n \to \infty$. This follows from a simple
    eigenvalue perturbation argument: By Weyl's theorem, for any $t \in
    \countingset{n}$, $\abs{\peigval_t - \eigval_t} \leq \spectralnorm{H} =
    O(\sqrt{\rho n})$. As a result, if $\eigval_t = 0$ then $\peigval_t =
    O(\sqrt{\rho n})$.  Since $\peigval_{s_1}, \ldots, \peigval_{s_K}$ are
    $\Theta(\rho n)$, there is a gap of size $\Theta(\rho n)$ w.h.p., between
    them and the remaining eigenvalues of $A$, and therefore the top $K$
    eigenvalues of $A$ are as claimed.

    Therefore, we assume that the top $K$ eigenvalues of $A$ in absolute value
    are $\peigval_{s_1}, \ldots, \peigval_{s_K}$. Then:
    \[
        \hat{M} = 
        \sum_{k=1}^K
            \peigval_{s_k}
            \peigvec{s_k}
            \otimes
            \peigvec{s_k},
    \]
    where $\peigvec{s_k}\otimes \peigvec{s_k}$ is the outer product of these two
    vectors. 
    Since $M$ is rank $K$, we have
    \[
        M =
        \sum_{k=1}^K
            \eigval_{s_k}
            \eigvec{s_k}
            \otimes
            \eigvec{s_k}.
    \]
    As a result, we have
    \[
        M_{ij} = \sum_{k=1}^K \eigval_{s_k} \eigvec{s_k}_i  \eigvec{s_k}_j,
        \qquad
        \hat{M}_{ij} = \sum_{k=1}^K \peigval_{s_k} 
            \peigvec{s_k}_i  \peigvec{s_k}_j.
    \]
    \newcommand{\eigvecdiff}[1]{\Delta^{(#1)}}
    For any $t \in \{s_1, \ldots, s_K\}$, define $\eigvecdiff{t} = \peigvec{t} -
    \eigvec{t}$ and let $\epsilon_t = \peigval_t - \eigval_t$. Then
    $\peigvec{t} = \eigvec{t} + \eigvecdiff{t}$ and $\peigval_t = \eigval_t +
    \epsilon_t$. Hence:
    \[
        \hat{M}_{ij}
        =
        \sum_{k=1}^K
        (\eigval_{s_k} + \epsilon_{s_k})
        (\eigvec{s_k}_i + \eigvecdiff{s_k}_i)
        (\eigvec{s_k}_j + \eigvecdiff{s_k}_j).
    \]
    From \Cref{sec:application}, we have that $\abs{\eigvecdiff{t}_i} \leq
    C \rho^{\nicefrac{-1}{2}} n^{-1} 
    \log^\xi n$ simultaneously for all $t \in \{s_1, \ldots, s_K\}$ and $i \in
    \countingset{n}$ with high probability. Furthermore, consulting
    \Cref{table:qts} shows that
    $\abs{\eigvec{t}_i} = \Theta(\nicefrac{1}{\sqrt{n}})$. Combining this with
    Weyl's bound of $\epsilon_t \leq \spectralnorm{H} = O(\sqrt{\rho n})$, it is
    easy to see that:
    \begin{align*}
        \hat{M}_{ij}
        &=
            M_{ij}
            +
            O\left(
                \,
                \sum_{k=1}^K
                \eigval_{s_k}
                \eigvec{s_k}_i
                \eigvecdiff{s_k}_j
            \right)
        ,\\
        &=
            M_{ij}
            +
            O\left(
                \,
                K
                \cdot
                \rho n
                \cdot
                \frac{1}{\sqrt{n}}
                \cdot
                \frac{\log^\xi n}{n \sqrt{\rho}}
            \right)
        ,\\
        &=
            M_{ij}
            +
            O\left(
                \sqrt{
                \frac{ \rho }{ n }
                }
                \log^\xi n
            \right)
        .\\
    \end{align*}
\end{proof}

\subsection{Proof of \Cref{result:consistency}}

We now prove \Cref{result:consistency}, restated below for convenience:

\restateconsistency*{}

\begin{proof}
    We will use \Cref{result:entrywise} to show that, with high probability as
    $n \to \infty$, for all pairs of graph nodes $i$ and $j$ simultaneously, $i$
    and $j$ belong to the same latent community if and only if 
    $\inftynorm{M_i - M_j} < \tau$.

    Recall that we write $z(i)$ to denote the latent community label of node
    $i$.  Define:
    \[
        \Delta = 
        \min_{\substack{i, j \\ z(i) \neq z(j)}}
        \inftynorm{M_i - M_j}.
    \]
    Since $M_{ij} = \rho \cdot P^{(0)}_{z(i), z(j)}$ we have:
    \[
        \Delta = 
        \rho \cdot
        \min_{k \neq k'}
        \inftynorm{P^{(0)}_k - P^{(0)}_{k'}}
        =
        \Theta(\rho).
    \]
    Thus there exists a constant $C$ (depending on $P^{(0)}$) such that for all
    blockmodels in the sequence, if $i$ and $j$ belong to different communities,
    then $\inftynorm{M_i - M_j} \geq C \rho$. Therefore we are able to recover
    the communities exactly if $M$ is known.

    Observe that:
    \begin{align*}
        \inftynorm{ \hat{M}_i - \hat{M}_j }
        &=
            \inftynorm{ M_i + (\hat{M}_i - M_i) - M_j - (\hat{M}_j - M_j) }
            ,\\
        &=
            \inftynorm{ (M_i - M_j) + (\hat{M}_i - M_i) - (\hat{M}_j - M_j) }
            .
    \end{align*}
    As a result,
    \begin{align*}
        \abs*{\inftynorm{M_i - M_j} - \inftynorm{\hat{M}_i - \hat{M}_j}}
        &\leq
            \inftynorm{\hat{M}_i - M_i} + \inftynorm{\hat{M}_j - M_j}
            ,\\
        &=
            O\left(
                \sqrt{\frac{\rho}{n}} \cdot \log^\xi n
            \right),
    \end{align*}
    where we have substituted the result of \Cref{result:entrywise}. 
    Since $\xi < \epsilon/2$ by assumption, we have that
    \[
        \frac{\log^\xi n}{\sqrt{n}}
        =
        o\left(
            \sqrt{
                \frac{\log^\epsilon n}{n}
            }
        \right)
        =
        o(\sqrt{\rho}),
    \]
    where in the last step we used the assumption that $\rho =
    \omega(n^{-1} \log^\epsilon n)$. Therefore $\sqrt{\nicefrac{\rho}{n}} \cdot
    \log^\xi n = o(\rho)$.
    In particular, if $i$ and $j$ belong to different communities then
    \[
        \inftynorm{\hat{M}_i - \hat{M}_j}
        \geq
        C \rho - 
        O\left(
            \sqrt{\frac{\rho}{n}} \cdot \log^\xi n
        \right)
        =
        \Omega(\rho).
    \]
    Hence if $\tau = o(\rho)$, 
    $\inftynorm{\hat{M}_i - \hat{M}_j} > \tau$ w.h.p.\ and thus
    $i$ and $j$ will be clustered into different
    communities by \Cref{alg:cluster} with high probability as $n \to \infty$.

    On the other hand, suppose that $i$ and $j$ belong to the same community.
    Then, as shown above, $\inftynorm{\hat{M}_i - \hat{M}_j} =
    O(\sqrt{\nicefrac{\rho}{n}} \cdot \log^\xi n)$. Therefore, if $\tau =
    \omega(\sqrt{\nicefrac{\rho}{n}} \cdot \log^\xi n)$, $\inftynorm{\hat{M}_i -
    \hat{M}_j} \leq \tau$ with high probability as $n \to \infty$, and therefore
    $i$ and $j$ are clustered together.
\end{proof}

\subsection{A remark on the classical theory}
\label{sec:consremark}

In \Cref{remark:svd} it was claimed that proving the consistency of
\Cref{alg:cluster} is difficult with the classical theory. We now expand on
this.

We have seen that in the context of the sparse stochastic blockmodel (i.e.,
\Cref{setting:sparse}) the classical bound on the perturbation of the top $K$
eigenvectors in 2-norm is $\Theta(\nicefrac{1}{\sqrt{\rho n}})$; see
\Cref{table:pqts} and the discussion in \Cref{result:application} for reference.
We now argue that this implies a bound of 
\[
    \frobnorm{\hat{M} - M} = \sqrt{\sum_{i,j} (\hat{M}_{ij} - M_{ij})^2}
    = O(\sqrt{\rho n}).
\]
Recall that we have assumed for simplicity that the eigenvalues of $M$ are
non-negative. Then the top $K$ eigenvalues of $M$ in absolute value are simple
$\eigval_1, \ldots, \eigval_K$, and:
\[
    M = \sum_{k = 1}^K \eigval_k \eigvec{k} \otimes \eigvec{k}.
\]
Assume that the top $K$ eigenvalues of $A$ are the largest in magnitude -- as
argued above, this will be true with high probability as $n \to \infty$. Then
the rank $K$ approximation of $M$ is:
\[
    \hat{M} = 
    \sum_{k = 1}^K \peigval_k \peigvec{k} \otimes \peigvec{k}.
\]
Consider the $t$th eigenvalue and eigenvector for $t \in \countingset{K}$; the
following argument will hold for the remaining of the top $K$ eigenvalues since
they are of the same order.  Write $\peigval_t = \eigval_t + \epsilon_t$. We
have:
\begin{align*}
    \frobnorm{\eigval_t \eigvec{t} \otimes \eigvec{t} - \peigval_t \peigvec{t}
    \otimes \peigvec{t}}
    &=
    \frobnorm{\eigval_t \eigvec{t} \otimes \eigvec{t} - 
            (\eigval_t + \epsilon_t)
        \peigvec{t} \otimes \peigvec{t}}
            ,
            \\
    &\leq
        \underbrace{
        \eigval_t
        \frobnorm{
            \eigvec{t} \otimes \eigvec{t} - 
            \peigvec{t} \otimes \peigvec{t}
        }
        }_{A}
        +
        \underbrace{
        \abs{\epsilon_t}\cdot
        \frobnorm{
            \peigvec{t} \otimes \peigvec{t}
        }
        }_{B}
            ,
\end{align*}

Weyl's theorem gives a bound of $\abs{\epsilon_t} \leq \spectralnorm{H} =
O(\sqrt{\rho n})$. Since $\peigvec{t}$ is a unit vector, $\frobnorm{\peigvec{t}
\otimes \peigvec{t}} \leq 1$, and so $B = O(\sqrt{\rho n})$.

We now bound $A$. Let $\Delta = \peigvec{t} - \eigvec{t}$. We have:
\begin{align*}
    \frobnorm{
        \eigvec{t} \otimes \eigvec{t} 
        - 
        \peigvec{t} \otimes \peigvec{t}
    }
    &=
    \frobnorm{
        \eigvec{t} \otimes \eigvec{t} 
        - 
        (\eigvec{t} + \Delta) \otimes (\eigvec{t} + \Delta)
        }
        ,\\
    &\leq
        \frobnorm{
            \eigvec{t} \otimes \Delta
        }
        +
        \frobnorm{
            \Delta \otimes \eigvec{t}
        }
        +
        \frobnorm{
            \Delta \otimes \Delta
        }.
\end{align*}
Using the submultiplicative property of the Frobenius norm, we bound each of
these terms by $\frobnorm{\Delta} = \twonorm{\Delta} = O(\nicefrac{1}{\sqrt{\rho
n}})$. Then, since $\eigval_t = \Theta(\rho n)$, we have a bound on $A$ and
also $\frobnorm{\hat{M} - M}$ of
$O(\sqrt{\rho n})$.

Such a bound is not sufficient to cluster the columns of $\hat{M}$ in a way that
recovers the correct clustering exactly with high probability. For instance,
suppose that $i$ and $j$ belong to different clusters. Let $\hat{M}$ be the
matrix which is identical to $M$, except that column and row $i$ is made to look
exactly like column $j$.  It is easy to see that $\hat{M}$ differs from $M$ in
$O(n)$ entries, and each difference has magnitude $\rho$. Therefore,
$\frobnorm{\hat{M} - M} = O(\sqrt{\rho n})$. But by construction it is
impossible to distinguish $i$ from $j$ using $\hat{M}$. On the other hand, our
bound on $\maxnorm{\hat{M} - M}$ is sufficient, as shown in the proof of
\Cref{result:consistency} above.

\section{Eigenvalue perturbation proofs}

\subsection{Proof of \Cref{result:eigvalupper}}

\restateeigvalupper*{}

\begin{proof}
    \label{proof:eigvalupper}
    The min-max priciple says
    \begin{align*}
        \peigval_t 
        &=
            \min_{S \in \mathcal{S}_{n - t + 1}}
            \max_{\substack{x \in S\\\|x\|=1}}
                x^\transpose (M + H) x,
    \intertext{where $\mathcal{S}_{n - t + 1}$ is the set of all subspaces of
    $\reals^n$ of dimension $n - t + 1$. In particular, fix the subspace to be
    $S_{t:n} = \Vecspan(\{\eigvec{t}, \ldots, \eigvec{n}\})$ such that}
        &\leq
            \max_{x \in S_{t:n}}
                x^\transpose (M + H) x.
    \intertext{We may write any unit vector $x \in S_{t:n}$ as $\alpha u + \beta
    u_\perp$ for some unit vector $u \in S_{t:\dupper}$ and some unit vector
    $u_\perp \in S_{\dupper+1:n}$, with the constraint $\alpha^2 + \beta^2 = 1$.
    As such, the above maximization is equivalent to:}
        &=
            \max_{\substack{\alpha,\beta\\\alpha^2 + \beta^2 = 1}}\;
            \max_{u \in S_{t:\dupper}}\;
            \max_{u_\perp \in S_{\dupper+1:n}}
                (\alpha u + \beta u_\perp)^\transpose
                (M + H)
                (\alpha u + \beta u_\perp).\\
    \intertext{Expanding the quadratic form:}
        &=
            \max_{\substack{\alpha,\beta\\\alpha^2 + \beta^2 = 1}}\;
            \max_{u \in S_{t:\dupper}}\;
            \max_{u_\perp \in S_{\dupper+1:n}}
                \bigg\{\\
                    &\qquad\qquad
                    \alpha^2 u^\transpose M u
                    +
                    \alpha^2 u^\transpose H u\\
                    &\qquad\qquad
                    +
                    \xcancel{2 \alpha \beta u^\transpose M u_\perp}
                    +
                    2 \alpha \beta u^\transpose H u_\perp\\
                    &\qquad\qquad
                    +
                    \beta^2 u_\perp^\transpose M u_\perp
                    +
                    \beta^2 u_\perp^\transpose H u_\perp\\
                &\qquad \bigg\}.
    \intertext{The $u^\transpose M u_\perp$ term drops, since $M u_\perp \in
    S_{\dupper+1:n}$, and this subspace is orthogonal to $S_{t:\dupper}$, of
    which $u$ is a member. We bound the remaining terms individually. First,
    $u^\transpose M u$ is at most $\eigval_t$, since $u$ is restricted to
    $S_{t:\dupper}$. We then bound $u^\transpose H u \leq h$ using
    the assumption.  Both $u^\transpose H u_\perp$
    and $u_\perp^\transpose H u_\perp$ can be at most $\|H\|$. Lastly,
    $u_\perp^\transpose M u_\perp$ can be at most $\eigval_{\dupper+1}$, since
    $u_\perp \in S_{\dupper+1:n}$. Collecting these upper bounds, we have:}
        &\leq
            \max_{\substack{\alpha,\beta\\\alpha^2 + \beta^2 = 1}}\;
            \left\{
                \alpha^2 \eigval_t
                +
                \alpha^2 h
                +
                2 \alpha \beta \|H\|
                +
                \beta^2 \eigval_{\dupper+1}
                +
                \beta^2 \|H\|
            \right\}.
    \intertext{Now, $\alpha \beta \|H\| \leq |\beta| \|H\|$ due to the
    constraint $\alpha^2 + \beta^2 = 1$. As such, the above is bounded by:}
        &\leq
            \max_{0 \leq \beta \leq 1}
                \left\{
                    (1 - \beta^2) \eigval_t
                    +
                    (1 - \beta^2) h
                    +
                    2 \beta \|H\|
                    +
                    \beta^2 \eigval_{\dupper+1}
                    +
                    \beta^2 \|H\|
                \right\},\\
        &=
            \eigval_t
            +
            h
            +
            \max_{0 \leq \beta \leq 1}
                \bigg\{
                    \underbrace{
                        \beta^2
                        \left(
                            \eigval_{\dupper+1} 
                            - \eigval_t 
                            - h
                            + \|H\|
                        \right)
                        +
                        2 \beta \|H\|
                    }_{g(\beta)}
                \bigg\},\\
        &=
            \eigval_t
            +
            h
            +
            \max_{0 \leq \beta \leq 1}
            g(\beta).
    \end{align*}
    Thus we bound $\peigval_t$ by maximizing $g(\beta)$ subject to $\beta \in
    [0,1]$. The derivative is:
    \[
        g'(\beta) =
            2 \beta
            \left(
                \eigval_{\dupper+1}
                -
                \eigval_t
                -
                h
                +
                \|H\|
            \right)
            +
            2 \|H\|.
    \]
    Solving $g'(\beta^*) = 0$ for $\beta^*$, we have:
    \[
        \beta^* = \frac{
            \|H\|
        }{
            \eigval_t - \eigval_{\dupper+1} + h - \|H\|
        }.
    \]
    Note that $\beta^* \in [0,1]$ as a consequence of the assumption $\eigval_t
    - \eigval_{\dupper+1} > 2\|H\| - h$.  Lastly, substituting this maximizing
    value into $g(\beta)$, we obtain:
    \begin{align*}
        \peigval_t 
        &\leq
            \eigval_t
            +
            h
            +
            g(\beta^*),\\
        &\leq
            \eigval_t
            +
            h
            -
            \frac{
                \|H\|^2
            }{
                \eigval_t - \eigval_{\dupper+1} + h - \|H\|
            }
            +
            \frac{
                2 \|H\|^2
            }{
                \eigval_t - \eigval_{\dupper+1} + h - \|H\|
            },\\
        &=
            \eigval_t
            +
            h
            +
            \frac{
                \|H\|^2
            }{
                \eigval_t - \eigval_{\dupper+1} + h - \|H\|
            }.
    \end{align*}
\end{proof}

\subsection{Bounding perturbations at both ends of the spectrum}

We now give the general result which bounds the perturbation of eigenvalues at
both ends of the spectrum.

\renewcommand{\ixupper}{s^{\uparrow}}
\renewcommand{\botsmallh}{h}
\renewcommand{\topsmallh}{h}
\begin{restatable}[Eigenvalue perturbation]{theorem}{restateeigval}
    \label{result:eigval}
    Let $\ixupper, \ixlower \in \{0, \ldots, n+1\}$ be such that $\ixupper <
    \ixlower$. Let $\topsmallh$ be such that 
    $\abs{\left\langle x, H x \right\rangle} \leq \topsmallh$
    for all $x \in \topeigspan$
    and for all $x \in \boteigspan$. 
    Then for any $t \leq \ixupper$, if $\eigval_t - \eigval_{\ixupper + 1} > 2
    \spectralnorm{H} - h$:
    \[
        \eigval_t
        -
        \topsmallh
        \leq
        \peigval_t
        \leq
        \eigval_t
        +
        \topsmallh
        +
        \frac{
            \spectralnorm{H}^2
        }{
            \eigval_t
            -
            \eigval_{\ixupper + 1}
            +
            \topsmallh
            -
            \spectralnorm{H}
        },
    \]
    and for any $t \geq \ixlower$, if $\eigval_{\ixlower} - \eigval_t > 2
    \spectralnorm{H} - h$:
    \[
        \eigval_t
        -
        \botsmallh
        -
        \frac{
            \spectralnorm{H}^2
        }{
            \eigval_{\ixlower + 1}
            -
            \eigval_t
            +
            \botsmallh
            -
            \spectralnorm{H}
        }
        \leq
        \peigval_t
        \leq
        \eigval_t
        +
        \botsmallh
        .
    \]
\end{restatable}

\begin{proof}
    \label{proof:eigval}
    \newcommand{\neigvec}[1]{v^{(#1)}}
    \newcommand{\npeigvec}[1]{\tilde{v}}
    \newcommand{\neigval}{\mu}
    \newcommand{\npeigval}{\tilde{\mu}}
    The statement for $t \leq \dupper$ has already been proven in
    \Cref{result:eigvalupper,result:eigvallower}. The statement for $t \geq
    \ixlower$ follows from a symmetric argument. Let $\hat{M} = - M$ and
    $\hat{H} = -H$.  Let $\neigval_1 \geq \cdots \geq \neigval_n$ be
    the eigenvalues of $\hat{M}$. Then $\neigval_i = - \eigval_{n - i + 1}$
    for any $1 \leq i \leq n$. Similarly, $\eigval_i = - \neigval_{n - i +
    1}$.  Furthermore, define $\neigvec{i} = \eigvec{n - i + 1}$. Then
    $\neigvec{i}$ is an eigenvector of $\hat{M}$ for the eigenvalue
    $\neigval_i$. It follows that for any $x \in \Vecspan(\{\neigvec{1}, \ldots,
    \neigvec{n - \ixlower + 1}\})$, we have $|x^\transpose \hat{M} x| \leq h$.
    In addition, we have $\neigval_{n - \ixlower + 1} - \neigval_{n - \ixlower +
    2} > 2\|H\| - h$.  \renewcommand{\eigvec}{u} Therefore, applying
    \Cref{result:eigvalupper,result:eigvallower} to $\hat{M} + \hat{H}$, we
    have, for any $t \leq n - \ixlower + 1$:
    \[
            \neigval_t 
            - 
            h
        \leq
            \npeigval_t
        \leq 
            \neigval_t 
            + 
            h 
            + 
            \frac{
                \|H\|^2
            }{
            \neigval_t - \neigval_{n - \ixlower + 2} + h - \|H\|
            }.
    \]
    Now, $\npeigval_t = - \peigval_{n - t + 1}$, such that:
    \[
            -\neigval_t 
            - 
            h 
            - 
            \frac{
                \|H\|^2
            }{
            \neigval_t - \neigval_{n - \ixlower + 2} + h - \|H\|
            }
        \leq
            \peigval_{n - t + 1}
        \leq 
        -\neigval_t 
            + 
            h.
    \]
    And recall that $-\neigval_t = \eigval_{n - t + 1}$. Hence, for any $t
    \leq n - \ixlower + 1$:
    \[
            \eigval_{n - t + 1} 
            - 
            h 
            - 
            \frac{
                \|H\|^2
            }{
                \eigval_{\ixlower - 1} - \eigval_{n - t + 1} + h - \|H\|
            }
        \leq
            \peigval_{n - t + 1}
        \leq 
            \eigval_{n - t + 1}
            + 
            h.
    \]
    Finally, we make a change of index such that $t \mapsto n - t + 1$. Then for
    any $t \geq \ixlower$:
    \[
            \eigval_{t} 
            - 
            h 
            - 
            \frac{
                \|H\|^2
            }{
                \eigval_{\ixlower - 1} - \eigval_{t} + h - \|H\|
            }
        \leq
            \peigval_{t}
        \leq 
            \eigval_{t}
            + 
            h.
    \]
\end{proof}

\section{Eigenvector perturbation proofs}

\subsection{Proof of \Cref{result:neumanntrick}: the Neumann trick}

\restateneumanntrick*{}

\begin{proof}
    \label{proof:neumanntrick}
    Since $\peigvec{t}$ is an eigenvector of $M + H$ with eigenvalue
    $\peigval_t$, we have $(M + H) \peigvec{t} = \peigval_t \peigvec{t}$.
    Rearranging, we obtain $M \peigvec{t} = (\peigval_t I - H)\peigvec{t}$.  By
    the assumption that $\spectralnorm{H} < \abs{\peigval_t}$ it follows that
    $\peigval_t$ is not an eigenvalue of $H$, and and so $(\peigval_t I - H)$ is
    invertible. Therefore:
    \begin{align*}
        \peigvec{t}
        &= 
            \frac{1}{\peigval_t}
            \left(I - \frac{H}{\peigval_t}\right)^{-1} M \peigvec{t}.
    \intertext{Since $\|H\| < \peigval_t$, we may expand $(I - H/\peigval_t)$ in
    a Neumann series:}
        &= 
            \frac{1}{\peigval_t}
            \sum_{k \geq 0} 
            \left(
                \frac{H}{\peigval_t}
            \right)^k
            M \peigvec{t}.
    \intertext{The eigenvectors of $M$ form an orthonormal basis for $\reals^n$.
    We may therefore write $\peigvec{t} = \sum_{s=1}^n \langle \peigvec{t},
    \eigvec{s} \rangle \eigvec{s}$. Using this in the above, we find:}
        &= 
            \frac{1}{\peigval_t}
            \sum_{k \geq 0} 
            \left(
                \frac{H}{\peigval_t}
            \right)^k
            \sum_{s=1}^n
                \left\langle
                    \peigvec{t},
                    \eigvec{s}
                \right\rangle
                M
                \eigvec{s},\\
        &= 
            \frac{1}{\peigval_t}
            \sum_{k \geq 0} 
            \left(
                \frac{H}{\peigval_t}
            \right)^k
            \sum_{s=1}^n
                \eigval_s
                \left\langle
                    \peigvec{t},
                    \eigvec{s}
                \right\rangle
                \eigvec{s},\\
        &= 
            \sum_{s=1}^n
                \frac{\eigval_s}{\peigval_t}
                \left\langle
                    \peigvec{t},
                    \eigvec{s}
                \right\rangle
            \sum_{k \geq 0} 
            \left(
                \frac{H}{\peigval_t}
            \right)^k
                \eigvec{s}.
    \end{align*}
\end{proof}

\subsection{A general perturbation bound based on the Neumann trick}
\label{sec:neumannperturb}

The result stated in \Cref{result:dkperturb} is a corollary of a more
general perturbation result, which we state below. The theorem takes as input
bounds on the perturbation of eigenvalues and the angle of the perturbation in
eigenvectors. \Cref{result:dkperturb} uses Weyl's theorem and the Davis-Kahan to
provide these bounds, however if better bounds are available
the following result will take advantage of them.

\begin{restatable}{theorem}{restateneumannperturb}
    \label{result:neumannperturb}
    Fix $t \in \countingset{n}$.
    Define $\eigerr = \abs{\eigval_t - \peigval_t}/{\abs{\eigval_t}}$ and let
    $\theta_s$ be the angle between $\peigvec{t}$ and $\eigvec{s}$.
    Then:
    \begin{align*}
        \abs*{\eigvec{t}_\ix - \peigvec{t}_\ix}
        &\leq
            \abs*{\eigvec{t}_\ix}
            \cdot
            \left(
                \sin^2 \theta_t
            +
                \frac{\eigerr}{\abs{\eigval_t} - \eigerr}
            \right)
            +
            \left(
                \frac{\abs*{\eigval_t}}{\abs*{\eigval_t} - \eigerr}
            \right)^2
            \cdot
            \tailbound{s}_\ix
            \\
        &\qquad
            +
            \sum_{s \neq t}
                \frac{
                    \abs{\eigval_s}
                \cdot
                \abs*{\cos \theta_s}
                }
                {
                    \abs{\eigval_t} - \eigerr
                }
                \cdot
                \left(
                    \abs*{\eigvec{s}_\ix}
                    +
                    \left[
                        \frac{\abs*{\eigval_t}}{\abs*{\eigval_t} - \eigerr}
                    \right]
                    \cdot
                    \tailbound{s}_\ix
                \right).
    \end{align*}
    where $\tailbound{s}$ is the $n$-vector whose $\ix$th entry is defined to be
    \[
        \tailbound{s}_\ix = 
        \abs*{
        \left[
        \sum_{k \geq 1} \left(
            \frac{H}{\eigval_t} 
        \right)^k
        \eigvec{s}
        \right]_\ix
        }.
    \]
\end{restatable}

\begin{proof}
    \label{proof:neumannperturb}
    \newcommand{\summand}[1]{\psi^{(#1)}}%
    Define
    \[
        \summand{s}
        =
            \frac{\eigval_s}{\peigval_t}
            \left\langle
                \peigvec{t},
                \eigvec{s}
            \right\rangle
            \sum_{k \geq 0} 
            \left(
                \frac{H}{\peigval_t}
            \right)^k
                \eigvec{s}.
    \]
    Note that $\summand{s}$ is a vector, and we write
    $\summand{s}_\ix$ to denote its $\ix$th element.
    Using this notation, \Cref{result:neumanntrick} is simply restated as:
    $
        \peigvec{t}
        =
        \sum_{s = 1}^n \summand{s}.
    $
    In particular we have equality for every entry, such that:
    \[
        \peigvec{t}_\ix
        =
        \sum_{s = 1}^n \summand{s}_\ix.
    \]
    Our goal is to bound $\abs{\eigvec{t}_\ix - \peigvec{t}_\ix}$. Using the
    above expression for $\peigvec{t}_\ix$, we obtain:
    \begin{align}
        \abs*{\eigvec{t}_\ix - \peigvec{t}_\ix}
        &=\nonumber
            \abs*{\eigvec{t}_\ix - \sum_{s=1}^n \summand{s}_\ix}.
    \intertext{We extract the $s = t$ term from the sum and use the triangle
    inequality to obtain:}
        &=\nonumber
            \abs*{\eigvec{t}_\ix - \summand{t}_\ix - 
                \sum_{s \neq t} \summand{s}_\ix}
            ,\\
        &\leq\label{eqn:ntdecomp}
            \abs*{\eigvec{t}_\ix - \summand{t}_\ix}
                + 
            \sum_{s \neq t} 
            \abs*{
                \summand{s}_\ix
            }
            .
    \end{align}

    We begin by bounding the first term. We have:
    \begin{align}
        \abs*{\eigvec{t}_\ix - \summand{t}_\ix}
        &=\nonumber
            \abs*{
                \eigvec{t}_\ix
                -
                \frac{\eigval_t}{\peigval_t}
                \left\langle
                    \peigvec{t},
                    \eigvec{t}
                \right\rangle
                \left[
                \sum_{k \geq 0} 
                \left(
                    \frac{H}{\peigval_t}
                \right)^k
                    \eigvec{t}
                \right]_\ix
            }
            ,\\
        &=\nonumber
            \abs*{
                \eigvec{t}_\ix
                -
                \frac{\eigval_t}{\peigval_t}
                \cdot
                \cos \theta_t
                \cdot
                \left[
                \sum_{k \geq 0} 
                \left(
                    \frac{H}{\peigval_t}
                \right)^k
                    \eigvec{t}
                \right]_\ix
            }.
    \intertext{Here we used the assumption that the angle between $\peigvec{t}$
    and $\eigvec{t}$ is acute.  We extract the $k = 0$ term from the series
    and use the triangle inequality again:}
    &=\label{eqn:ntparts}
            \underbrace{
            \abs*{
                \eigvec{t}_\ix
                -
                \frac{\eigval_t}{\peigval_t}
                \cdot \cos \theta_t \cdot
                \eigvec{t}_\ix
            }
        }_{\text{A}}
            +
            \underbrace{
            \abs*{
                \frac{\eigval_t}{\peigval_t}
                \cdot \cos \theta_t \cdot
                \left[
                \sum_{k \geq 1} 
                \left(
                    \frac{H}{\peigval_t}
                \right)^k
                    \eigvec{t}
                \right]_\ix
            }
            }_{\text{B}}
            .
    \end{align}

    We now bound A. We have
    \begin{align}
        \abs*{
            \eigvec{t}_\ix
            -
            \frac{\eigval_t}{\peigval_t}
            \cos \theta_t
            \eigvec{t}_\ix
        }
        &=\nonumber
            \abs*{\eigvec{t}_\ix}
            \cdot
            \abs*{
                1
                -
                \frac{\eigval_t}{\peigval_t}
                \cos \theta_t
            }
            ,\\
        &=\nonumber
            \abs*{\eigvec{t}_\ix}
            \cdot
            \abs*{
                1
                -
                \frac{\peigval_t + (\eigval_t - \peigval_t)}{\peigval_t}
                \cos \theta_t
            }
            ,\\
        &=\nonumber
            \abs*{\eigvec{t}_\ix}
            \cdot
            \abs*{
                1
                -
                \left(
                    1 
                    -
                    \frac{\eigval_t - \peigval_t}{\peigval_t}
                \right)
                \cos \theta_t
            }
            ,\\
        &\leq\nonumber
            \abs*{\eigvec{t}_\ix}
            \cdot
            \left(
            \abs*{
                1
                -
                \cos \theta_t
            }
            +
            \abs*{
                \frac{\eigval_t - \peigval_t}{\peigval_t}
                \cdot
                \cos \theta_t
            }
            \right).
    \intertext{Since $\theta_t$ is an acute angle, we have $0 \leq \cos \theta_t
    \leq 1$, and so $\abs{1 - \cos \theta_t} = 1 - \cos \theta_t$. But $\cos
    \theta_t = \sqrt{1 - \sin^2 \theta_t} \leq 1 - \sin^2 \theta_t$, such that:}
        &\leq\label{eqn:ntpta}
            \abs*{\eigvec{t}_\ix}
            \cdot
            \left(
                \sin^2 \theta_t
            +
            \abs*{
                \frac{\eigval_t - \peigval_t}{\peigval_t}
            }
            \right).
    \end{align}

    Because we view $\peigval_t$ as a perturbation of $\eigval_t$, it is natural
    to assume that $\eigval_t$ is known and that we have a bound on
    $\abs{\eigval_t - \peigval_t}$, and that we do not know $\peigval_t$. It is
    therefore desirable to upper bound $1/\abs{\peigval_t}$ in terms of $\eigerr
    = \abs{\eigval_t - \peigval_t}$ and $\eigval_t$. We have:

    \begin{equation}
        \label{eqn:invpeigval}
        \frac{1}{\abs*{\peigval_t}}
        =
            \frac{1}{\abs*{\eigval_t + \peigval_t - \eigval_t}}
        \leq
            \frac{1}{\abs*{\eigval_t} - \abs*{\peigval_t - \eigval_t}}
        =
            \frac{1}{\abs*{\eigval_t} - \eigerr}.
    \end{equation}

    Therefore we may write \Cref{eqn:ntpta} as:
    \[
        \abs*{
            \eigvec{t}_\ix
            -
            \frac{\eigval_t}{\peigval_t}
            \left\langle
                \peigvec{t},
                \eigvec{t}
            \right\rangle
            \eigvec{t}_\ix
        }
        \leq
            \abs*{\eigvec{t}_\ix}
            \cdot
            \left(
                \sin^2 \theta_t
            +
                \frac{\eigerr}{\abs{\eigval_t} - \eigerr}
            \right).
    \]

    We now turn to bounding part B of \Cref{eqn:ntparts}.
    We have:
    \begin{align}
        \nonumber
            \abs*{
                \left[
                \sum_{k \geq 1} 
                \left(
                    \frac{H}{\peigval_t}
                \right)^k
                    \eigvec{t}
                \right]_\ix
            }
        &\leq\nonumber
            \sum_{k \geq 1} 
            \,
                \abs*{
                \left[
                \left(
                    \frac{H}{\peigval_t}
                \right)^k
                    \eigvec{t}
                \right]_\ix
            }
            ,\\
        &\leq\nonumber
            \sum_{k \geq 1} 
            \,
                \abs*{
                \left[
                \left(
                    \frac{\eigval_t}{\peigval_t}
                \right)^k
                \left(
                    \frac{H}{\eigval_t}
                \right)^k
                    \eigvec{t}
                \right]_\ix
            }
            ,\\
        &=\nonumber
            \sum_{k \geq 1} 
            \,
                \abs*{
                    \frac{\eigval_t}{\peigval_t}
                }^k
                \cdot
                \abs*{
                \left[
                \left(
                    \frac{H}{\eigval_t}
                \right)^k
                    \eigvec{t}
                \right]_\ix
            }
            ,\\
    \intertext{From \Cref{eqn:invpeigval}, we have:}
        &\leq\nonumber
            \sum_{k \geq 1} 
            \,
                \left(
                    \frac{\abs*{\eigval_t}}{\abs*{\eigval_t} - \eigerr}
                \right)^k
                \cdot
                \abs*{
                \left[
                \left(
                    \frac{H}{\eigval_t}
                \right)^k
                    \eigvec{t}
                \right]_\ix
            }
            ,\\
        &\nonumber\leq
            \frac{\abs*{\eigval_t}}{\abs*{\eigval_t} - \eigerr}
            \cdot
            \sum_{k \geq 1} 
            \,
                \abs*{
                \left[
                \left(
                    \frac{H}{\eigval_t}
                \right)^k
                    \eigvec{t}
                \right]_\ix
            }
            ,\\
            &=\label{eqn:ntptb}
            \frac{\abs*{\eigval_t}}{\abs*{\eigval_t} - \eigerr}
            \cdot
            \tailbound{t}_\ix
    \end{align}
    As such, part B is bounded as:
    \begin{align*}
        \abs*{
            \frac{\eigval_t}{\peigval_t}
            \cdot \cos \theta_t \cdot
            \left[
            \sum_{k \geq 1} 
            \left(
                \frac{H}{\peigval_t}
            \right)^k
                \eigvec{t}
            \right]_\ix
        }
        &\leq
            \abs*{
                \frac{\eigval_t}{\peigval_t}
            }
            \cdot \cos \theta_t \cdot
            \abs*{
            \left[
            \sum_{k \geq 1} 
            \left(
                \frac{H}{\peigval_t}
            \right)^k
                \eigvec{t}
            \right]_\ix
            }
            ,\\
        &\leq
            \left(
                \frac{\abs{\eigval_t}}{\abs{\eigval_t} - \eigerr}
            \right)^2
            \cdot 
            \tailbound{t}_\ix
            .
    \end{align*}
    Where we used the fact that $\cos \theta_t \leq 1$ in the last line.
    We have therefore bounded the first term in \Cref{eqn:ntdecomp} by:
    \begin{equation}
        \label{eqn:ntterm1}
        \abs*{\eigvec{t}_\ix - \summand{t}_\ix}
        \leq
            \abs*{\eigvec{t}_\ix}
            \cdot
            \left(
                \sin^2 \theta_t
            +
                \frac{\eigerr}{\abs{\eigval_t} - \eigerr}
            \right)
            +
            \left(
                \frac{
                    \abs{\eigval_t}
                }{
                    \abs{\eigval_t} - \eigerr
                }
            \right)^2
            \cdot
            \tailbound{t}_\ix
            .
    \end{equation}
    We now bound the second term in \Cref{eqn:ntdecomp}:
    \begin{align*}
        \abs*{\summand{s}_\ix}
        &=
            \abs*{
                \frac{\eigval_s}{\peigval_t}
                \left\langle
                    \peigvec{t},
                    \eigvec{s}
                \right\rangle
                \left[
                \sum_{k \geq 0} 
                \left(
                    \frac{H}{\peigval_t}
                \right)^k
                    \eigvec{s}
                \right]_\ix
            }
        ,\\
        \intertext{First, the magnitude of the dot product is $\abs*{\cos
        \theta_s}$ by definition, hence:}
        &=
            \abs*{
                \frac{\eigval_s}{\peigval_t}
            }
            \cdot
            \abs*{\cos \theta_s}
            \cdot
            \abs*{
                \left[
                \sum_{k \geq 0} 
                \left(
                    \frac{H}{\peigval_t}
                \right)^k
                    \eigvec{s}
                \right]_\ix
            }
            .\\
    \intertext{Extracting the $k=0$ term from the sum, we have:}
        &=
            \abs*{
                \frac{\eigval_s}{\peigval_t}
            }
            \cdot
            \abs*{\cos \theta_s}
            \cdot
            \abs*{
                \eigvec{s}_\ix
                +
                \left[
                    \sum_{k \geq 1} 
                    \left(
                        \frac{H}{\peigval_t}
                    \right)^k
                        \eigvec{s}
                \right]_\ix
            }
            ,\\
        &\leq
            \abs*{
                \frac{\eigval_s}{\peigval_t}
            }
            \cdot
            \abs*{\cos \theta_s}
            \cdot
            \left(
                \abs*{
                    \eigvec{s}_\ix
                }
                +
                \abs*{
                \left[
                    \sum_{k \geq 1} 
                    \left(
                        \frac{H}{\peigval_t}
                    \right)^k
                        \eigvec{s}
                \right]_\ix
                }
            \right).
    \intertext{We can bound the sum as we did in \Cref{eqn:ntptb}. We obtain:}
        &\leq
            \abs*{
                \frac{\eigval_s}{\peigval_t}
            }
            \cdot
            \abs*{\cos \theta_s}
            \cdot
            \left(
                \abs*{
                    \eigvec{s}_\ix
                }
                +
                \left[
                    \frac{\abs*{\eigval_t}}{\abs*{\eigval_t} - \eigerr}
                \right]
                \cdot
                \tailbound{s}_\ix
            \right).
    \intertext{Using the bound for $1/\abs{\peigval_t}$ derived in
    \Cref{eqn:invpeigval}, we have:}
        &\leq
            \frac{\abs*{\eigval_s}}{\abs*{\eigval_t} - \eigerr}
            \cdot
            \abs*{\cos \theta_s}
            \cdot
            \left(
                \abs*{
                    \eigvec{s}_\ix
                }
                +
                \left[
                    \frac{\abs*{\eigval_t}}{\abs*{\eigval_t} - \eigerr}
                \right]
                \cdot
                \tailbound{s}_\ix
            \right).
    \end{align*}

    Substituting this result and \Cref{eqn:ntterm1} into \Cref{eqn:ntdecomp}, we
    arrive at:
    \begin{align*}
        \abs*{\eigvec{t}_\ix - \peigvec{t}_\ix}
        &\leq
            \abs*{\eigvec{t}_\ix}
            \cdot
            \left(
                \sin^2 \theta_t
            +
                \frac{\eigerr}{\abs{\eigval_t} - \eigerr}
            \right)
            +
            \left(
                \frac{\abs*{\eigval_t}}{\abs*{\eigval_t} - \eigerr}
            \right)^2
            \cdot
            \tailbound{s}_\ix
            \\
        &\qquad
            +
            \sum_{s \neq t}
                \frac{
                    \abs{\eigval_s}
                \cdot
                \abs*{\cos \theta_s}
                }
                {
                    \abs{\eigval_t} - \eigerr
                }
                \cdot
                \left(
                    \abs*{\eigvec{s}_\ix}
                    +
                    \left[
                        \frac{\abs*{\eigval_t}}{\abs*{\eigval_t} - \eigerr}
                    \right]
                    \cdot
                    \tailbound{s}_\ix
                \right).
    \end{align*}

\end{proof}

\subsection{Results concerning the perturbation of subspaces}

\newcommand{\nsubspace}{\mathcal{U}}
\newcommand{\psubspace}{\tilde{\mathcal{U}}}
\newcommand{\spectralpart}{\mathcal{B}}

In this section, we state results on the perturbation of subspaces which are
used in various proofs; in particular, the proof of \Cref{result:dkperturb}.
The purpose of these results is to handle the case when an eigenspace
$\nsubspace$ of $M$ has dimensionality larger than one. In this case, the basis
of $\nsubspace$ is determined only up to an orthogonal transformation. In most
practical applications, however, 
we assume that the corresponding subspace of the perturbed matrix $M +
H$ has a fixed basis.
Therefore we wish to find a basis of $\nsubspace$ and a bijection between its
basis vectors and the basis of $\psubspace$ such that each vector is close to
its counterpart in angle.

\newcommand{\matbasis}{X}
\newcommand{\goodmatbasis}{\tilde X}
\newcommand{\basis}{x}
\newcommand{\pbasis}{y}
\newcommand{\pmatbasis}{Y}
\newcommand{\goodpmatbasis}{\tilde Y}
\newcommand{\goodbasis}{\tilde{x}}
\newcommand{\goodpbasis}{\tilde{y}}
\newcommand{\targetbasis}{\hat x}
\newcommand{\targetmatbasis}{{\hat X}}

To begin, recall the definition of the principal angles between subspaces:

\begin{defn}[Principal angles between subspaces \cite{Zhu2013-np}]
    Let $\nsubspace$ and $\psubspace$ be two $d$-dimensional subspaces of
    $\reals^n$, and let $U$ and $\tilde{U}$ be any orthogonal matrices whose
    columns form orthonormal bases for $\nsubspace$ and $\psubspace$
    respectively. Let $\sigma_1 \geq \cdots \geq \sigma_d$ be the singular
    values of $U^\transpose \tilde{U}$. The $i$th principal angle between
    $\nsubspace$ and $\psubspace$ is defined to be $\cos^{-1} \sigma_i$. We
    write
    \[
        \Theta(\nsubspace,\psubspace)
        =
        \Theta(U, \tilde{U})   
        =
        \operatorname{diag}(\cos^{-1} \sigma_1, \ldots, \cos^{-1} \sigma_d),
    \]
    for the $d \times d$ diagonal matrix of principal angles, and $\sin
    \Theta(\nsubspace, \psubspace) = \Theta(U, \tilde{U})$ for the diagonal
    matrix obtained by applying sine to every principal angle.
\end{defn}

The Davis-Kahan theorem in its full generality bounds the principal angles
between the subspaces of $M$ and the perturbation $M + H$:

\begin{theorem}[Davis-Kahan for statisticians; \cite{Yu2015-ck}]
    \label{result:daviskahan}
    Let $M$ and $H$ be $n \times n$ symmetric matrices. Let the eigenvalues
    of $M$ and $M + H$ be $\eigval_1 \geq \cdots \geq \eigval_n$ and $\peigval_1
    \cdots \peigval_n$ respectively. Fix $1 \leq r \leq s \leq n$ and 
    define $\delta = \min(\eigval_{r-1} - \eigval_r, \eigval_s - \eigval_{s +
    1})$, where we have defined $\eigval_0 = \infty$ and $\eigval_{n+1} = -
    \infty$ for convenience. Assume that $\delta > 0$. Let $d = s - r + 1$, and
    let $U = (\eigvec{r},
    \eigvec{r+1}, \ldots, \eigvec{s})$ and $\tilde{U} = (\peigvec{r},
    \peigvec{r+1}, \ldots, \peigvec{s})$ be orthonormal $n \times d$ matrices
    such that $M \eigvec{i} = \eigval_i \eigvec{i}$ and $(M + H) \peigvec{i} =
    \peigval_i \peigvec{i}$ for all $i \in \{r, \ldots, s\}$. 
    Then:
    \[
        \frobnorm{\sin \Theta(U, \tilde{U})}
        \leq
        2 \sqrt{d} \cdot \frac{\spectralnorm{H}}{\delta}.
    \]
\end{theorem}
\label{proof:daviskahan}

\renewcommand{\nsubspace}{\mathcal{X}}
\renewcommand{\psubspace}{\mathcal{Y}}

The next result shows that if the basis of $\psubspace$ is fixed and we know
that the maximum principal angle between $\psubspace$ and another subspace
$\nsubspace$ is small, then we can find a suitable orthonormal basis for
$\nsubspace$ such that the basis vectors of both subspaces are roughly aligned.

\begin{lemma}
    \label{result:pabs}
    Let $\nsubspace$ and $\psubspace$ be 
    $d$-dimensional subspaces of $\reals^n$.
    Suppose that the maximum principal angle\footnote{A principal angle
    $\theta_i$ is such that $0 \leq \theta_i \leq \pi/2$ by definition.} between
    $\nsubspace$ and $\psubspace$ is $\theta$, and define $\delta = \sin \theta$.
    Then for any orthonormal basis $\pbasis_1, \ldots, \pbasis_d$ for
    $\psubspace$, there exists an orthonormal basis $\targetbasis_1, \ldots,
    \targetbasis_d$ for $\nsubspace$ such that
    \begin{align*}
        \langle \targetbasis_i, \pbasis_i \rangle & 
            \geq 1 - \delta^2,
            \qquad \text{$\forall i$},\\
        \left| \langle \targetbasis_i, \pbasis_j \rangle \right| & 
            \leq \delta^2,
            \qquad \text{when $i \ne j$}.
    \end{align*}
\end{lemma}

\newcommand{\kronecker}{\delta}
\begin{proof}

    Let $\pmatbasis = (\pbasis_1, \ldots, \pbasis_d)$ be the $n \times d$ matrix
    of basis vectors of $\psubspace$.
    Let $\matbasis = (\basis_1, \ldots, \basis_d)$ be an $n \times d$ matrix
    whose orthonormal columns form a basis for $\nsubspace$; the choice of basis
    is arbitrary. It known that the principal angles between subspaces can be
    calculated by a singular value decomposition. In particular, let $U \Sigma
    V^\transpose$ be the SVD of $\matbasis^\transpose \pmatbasis$. Assume that
    the singular values $\sigma_i$ are placed in decreasing order along the
    diagonal of $\Sigma$.  Let $\theta_i$ be the $i$th smallest principal angle.
    Then $\sigma_i = \cos \theta_i$. Note that
    \[
        \cos \theta_i 
        = 
        \sqrt{1 - \sin^2 \theta_i} 
        \geq 
        \sqrt{1 - \sin^2 \theta}
        \geq 
        \sqrt{1 - \delta^2}
        \geq
        1 - \delta^2,
    \]
    and therefore every singular value is bounded as $1 - \delta^2 \leq \sigma_i
    \leq 1$.

    Let $\goodmatbasis = \matbasis U$ and $\goodpmatbasis = \pmatbasis V$. 
    Then
    \[
        \goodmatbasis^\transpose \goodpmatbasis
        =
        U^\transpose \matbasis^\transpose \pmatbasis V
        =
        U^\transpose U \Sigma V^\transpose V
        =
        \Sigma,
    \]
    where we used the fact that $U$ and $V$ are orthonormal $d \times d$
    matrices. Next, note that $\pmatbasis = \goodpmatbasis V^\transpose$, and
    define $\targetmatbasis = \goodmatbasis V^\transpose$. We claim that the
    columns of $\targetmatbasis$ form an orthonormal basis for $\nsubspace$. To
    see this, we first show orthonormality of the columns. We have
    \[
        \targetmatbasis^\transpose \targetmatbasis
        =
        V \goodmatbasis^\transpose \goodmatbasis V^\transpose
        =
        V (\matbasis U)^\transpose (\matbasis U) V^\transpose
        =
        V U^\transpose \matbasis^\transpose \matbasis U V^\transpose
        =
        I,
    \]
    where in the last step we use the fact that the columns of $\matbasis$ are
    orthonormal, and that $U$ and $V$ are orthonormal matrices. Next we show
    that the columns of $\targetmatbasis$ form a basis for $\nsubspace$. We do so
    by proving that the projection operator $\targetmatbasis
    \targetmatbasis^\transpose$ is in fact equal to $\matbasis
    \matbasis^\transpose$. We have
    \begin{align*}
        \targetmatbasis \targetmatbasis^\transpose
        &=
        (\goodmatbasis V^\transpose)(\goodmatbasis V^\transpose)^\transpose,\\
        &=
        \goodmatbasis V^\transpose V \goodmatbasis,\\
        &=
        \goodmatbasis \goodmatbasis^\transpose,\\
        &=
        (\matbasis U)(\matbasis U)^\transpose,\\
        &=
        \matbasis U U^\transpose \matbasis^\transpose,\\
        &=
        \matbasis \matbasis^\transpose.
    \end{align*}
    And so our claim is proven.
    
    Now we wish to show that the basis given by $\targetmatbasis$ is ``aligned''
    with the basis given by $\pmatbasis$ in the sense that the angle between
    corresponding basis elements is small. See that
    \[
        \targetmatbasis^\transpose \pmatbasis
        =
        V \goodmatbasis^\transpose \goodpmatbasis V^\transpose
        =
        V \Sigma V^\transpose.
    \]
    Defining $\targetbasis_i$ as the $i$th column of $\targetmatbasis$, we have
    that $\langle \targetbasis_i, \pbasis_j \rangle$ is the $ij$ element of $V
    \Sigma V^\transpose$. Therefore:
    \begin{align*}
        \langle \targetbasis_i, \pbasis_j \rangle
        &=
            \sum_{k=1}^d
            V_{ik}
            \sigma_k
            V_{jk}.
    \intertext{Write $\sigma_k = 1 - r_k$, where $0 \leq r_k \leq \delta^2$.
    Then:}
        &=
            \sum_{k=1}^d
            V_{ik}
            V_{jk}
            (1 - r_k),\\
        &=
            \sum_{k=1}^d
            V_{ik}
            V_{jk}
            -
            \sum_{k=1}^d
            r_k
            V_{ik}
            V_{jk}.
    \intertext{The first sum is simply the dot product between the $i$th and
    $j$th column of $V$. Since $V$ is orthogonal, this is 1 if $i = j$,
    and 0 otherwise. Using the notation $\kronecker_{i,j}$ for the Kronecker
    function, we have:}
        &=
            \kronecker_{i,j}
            -
            \sum_{k=1}^d
            r_k
            V_{ik}
            V_{jk}.
    \end{align*}
    We can easily bound the magnitude of the remaining sum:
    \begin{align*}
        \left|
        \sum_{k=1}^d
        r_k
        V_{ik}
        V_{jk}
        \right|
        &\leq
            \sum_{k=1}^d
            r_k
            \left|
            V_{ik}
            V_{jk}
            \right|
            ,\\
        &=
            r_k
            \sum_{k=1}^d
            |
            V_{ik}
            ||
            V_{jk}
            |
            ,\\
        &\leq
            \delta^2
            \sum_{k=1}^d
            |
            V_{ik}
            ||
            V_{jk}
            |
            .
    \intertext{Define the $d$-vector $\tilde{v}^{(\ell)}$ to be the entrywise
    absolute value of the $\ell$-th row of $V$; i.e.,
    $\tilde{v}^{(\ell)}_k = |V_{\ell k}|$. Then
    the above is:}
        &=
            \delta^2
            \langle
                \tilde{v}^{(i)}_k,
                \tilde{v}^{(j)}_k
            \rangle
            .
    \intertext{Applying the Cauchy-Schwarz inequality, we find:}
        &\leq
            \delta^2
            \|
                \tilde{v}^{(i)}_k
            \|
            \|
                \tilde{v}^{(j)}_k
            \|
            .
    \intertext{It is easily seen that $\|\tilde{v}_k^{(\ell)}\|$ is the norm of
    the $\ell$-th row of $V$. Since $V$ is orthonormal, this is simply one.
    Therefore:}
        &\leq
            \delta^2
            .
    \end{align*}
    As such, $\langle \targetbasis_i, \pbasis_j \rangle$ is not more than $
    \delta^2$ away from $\kronecker_{i,j}$, proving the result.
\end{proof}

The following result combines the previous lemma with the Davis-Kahan theorem.

\begin{lemma}
    \label{result:dkangles}
    Let $M$ and $H$ be $n \times n$ symmetric matrices.  
    Let the eigenvalues of $M$ be $\eigval_1, \ldots, \eigval_n$, and the
    eigenvalues of $M + H$ be $\peigval_1, \ldots, \peigval_n$. Let
    $\peigvec{1}, \ldots, \peigvec{n}$ be an orthonormal set of eigenvectors of
    $M + H$ such that $(M + H)\peigvec{t} = \peigval_t \peigvec{t}$.
    For any $s \in \countingset{n}$, let $\Lambda_s = \{ i : \eigval_i
    = \eigval_s \}$. 
    Define $d_s = \abs*{\Lambda_s}$, and let the gap be defined
    as $\gap_s = \min_{i \not \in \Lambda_s} \abs*{\eigval_s - \eigval_i}$.
    Denote by $\theta_s$ the angle between $\peigvec{t}$ and $\eigvec{s}$.
    There exists an orthonormal set of eigenvectors $\eigvec{1}, \ldots,
    \eigvec{n}$ satisfying $M \eigvec{s} = \eigval_s \eigvec{s}$ such that for
    $t \in \countingset{n}$:
    \[
        \sin \theta_t \leq 2 \sqrt{2 d_t} \cdot 
            \frac{\spectralnorm{H}}{\gap_t},
        \qquad
        \abs*{\cos \theta_s} \leq
            2 \sqrt{2} \cdot \spectralnorm{H} \cdot
            \min
            \left\{
                \frac{\sqrt{d_i}}{\gap_i}
            \right\}_{i \in \{s,t\}}.
    \]
\end{lemma}

\begin{proof}
    \label{proof:dkangles}
    We first show that there exists an orthonormal basis $\eigvec{1}, \ldots,
    \eigvec{n}$ of eigenvectors of $M$ such that $\eigvec{i}$ is close in angle
    to $\peigvec{i}$ for all $i \in \countingset{n}$, provided that the
    perturbation is too large. Choose any $s \in \countingset{n}$.  Define
    $\nsubspace_s$ to be the subspace of the range of $M$ corresponding to
    $\Lambda_s$. That is:
    \[
        \nsubspace_s
        =
        \Vecspan\left(\{
            x : M x = \eigval_s x
        \}\right).
    \]
    Similarly:
    \[
        \psubspace_s
        =
        \Vecspan(\{
            \peigvec{i} : i \in \Lambda_s
        \}).
    \]

    Let $\theta$ be the maximum principal angle between $\nsubspace_s$ and
    $\psubspace_s$. In particular, $\abs{\sin \theta} \leq
    \frobnorm{\Theta(\nsubspace_s, \psubspace_s)}$.
    Therefore, applying the Davis-Kahan theorem, we have that $\abs{\sin \theta}
    \leq 2 \sqrt{d_s} \cdot \spectralnorm{H} / \gap_s$. \Cref{result:pabs}
    states that there exists an orthonormal basis 
    $\{\eigvec{i}\}_{i \in \Lambda_s}$ for $\nsubspace_s$ such that for every
    $i \in \Lambda_s$:
    \[
        \left\langle
            \peigvec{i}, \eigvec{i}
        \right\rangle
        \geq
        1 - \sin^2 \theta
        =
        1 - 4 d_s
        \left(
            \frac{\spectralnorm{H}}{\gap_s}
        \right)^2.
    \]
    Since $\nsubspace_s$ is a $d_s$-dimensional subspace spanned by eigenvectors
    with the same eigenvalue, any vector in the subspace is an eigenvector.
    Namely, $\{\eigvec{i}\}_{i \in \Lambda_s}$ is an orthonormal set of
    eigenvectors spanning $\nsubspace_s$. We can repeat this process for each
    eigenspace of $M$, resulting in the desired orthonormal basis.

    Assume this basis, and consider $t$ as fixed. Note that for any $r \in
    \countingset{n}$ we have: 
    \begin{align*}
        \sqrt{
            1 -
            \left\langle \peigvec{r}, \eigvec{r} \right\rangle^2
        }
        &\leq
            \sqrt{
                1 -
                \left[
                    1 - 
                    4 d_r 
                    \cdot
                    \left(
                        \frac{\spectralnorm{H}}{\gap_r}
                    \right)^2
                \right]^2
            }
            ,\\
    \intertext{Expanding the square:}
        &=
            \sqrt{
                1 -
                \left[
                    1
                    -
                    8 d_r
                    \cdot
                    \left(
                        \frac{\spectralnorm{H}}{\gap_r}
                    \right)^2
                    +
                    16 d_r^2
                    \cdot
                    \left(
                        \frac{\spectralnorm{H}}{\gap_r}
                    \right)^4
                \right]
            }
            ,\\
        &=
            \sqrt{
                8 d_r
                \cdot
                \left(
                    \frac{\spectralnorm{H}}{\gap_r}
                \right)^2
                -
                16 d_r^2
                \cdot
                \left(
                    \frac{\spectralnorm{H}}{\gap_r}
                \right)^4
            }
            ,\\
        &\leq
            \sqrt{
                8 d_r
                \cdot
                \left(
                    \frac{\spectralnorm{H}}{\gap_r}
                \right)^2
            }
            ,\\
        &=
            2 \sqrt{2 d_r}
            \cdot
            \frac{\spectralnorm{H}}{\gap_r}
            .
    \end{align*}
    Define $\theta_s$ to be the angle between $\peigvec{t}$ and $\eigvec{s}$.
    Namely, we have
    \[
        \sin^2 \theta_t
        =
        1 - \cos^2 \theta_t
        =
        1 - \left\langle \peigvec{t}, \eigvec{t} \right\rangle^2
        \leq
        8 d_t \cdot \left(\frac{\spectralnorm{H}}{\gap_t}\right)^2.
    \]
    By the same token:
    \begin{align*}
        \abs*{\cos \theta_s}
        &=
            \abs*{\left\langle \peigvec{t}, \eigvec{s} \right\rangle}
        ,\\
        &\leq
            \sqrt{
                1 -
                \left\langle \peigvec{s}, \eigvec{s} \right\rangle^2
            }
        ,\\
        &\leq
            2 \sqrt{2 d_s}
            \cdot
            \frac{\spectralnorm{H}}{\gap_s}
            .
    \end{align*}

    But we also have
    \begin{align*}
        \abs*{\cos \theta_s}
        &=
            \abs*{\left\langle \peigvec{t}, \eigvec{t} \right\rangle}
        ,\\
        &\leq
            \sqrt{
                1 -
                \left\langle \peigvec{s}, \eigvec{t} \right\rangle^2
            }
        ,\\
        &\leq
            2 \sqrt{2 d_t}
            \cdot
            \frac{\spectralnorm{H}}{\gap_t}
            .
    \end{align*}
    Therefore:
    \[
        \abs*{\left\langle \peigvec{t}, \eigvec{s} \right\rangle}
        \leq
        2\sqrt{2}\cdot \spectralnorm{H}
        \cdot
        \min
        \left\{
            \frac{\sqrt{d_i}}{\gap_i}
        \right\}_{i \in \{s,t\}}.
    \]

\end{proof}

\subsection{Proof of \Cref{result:dkperturb}}

We will prove the following theorem which was originally stated in
\Cref{sec:eigenvectors}.

\restatedkperturb*{}

\begin{proof}
    \label{proof:dkperturb}
    The proof is an immediate corollary of combining
    \Cref{result:neumannperturb} (given in \Cref{sec:neumannperturb}) with
    \Cref{result:dkangles} (given in \Cref{proof:dkangles}), and using Weyl's
    bound of $\spectralnorm{H}$ for the perturbation of eigenvalues.
\end{proof}

\section{Results concerning random perturbations}

In the following, the term \emph{symmetric random matrix} will have a technical
meaning.

\begin{defn}
    \label{defn:srm}
    A \term{symmetric random matrix} $H$ is an $n \times n$ matrix whose entries
    are random variables satisfying $\expectation H_{ij} = 0$. Furthermore, we
    assume that the entries along the diagonal and in the upper-triangle $(j
    \geq i)$ are statistically independent, while the entries in the
    lower-triangle $(j < i)$ are constrained to be equal to their transposes:
    $H_{ij} = H_{ji}$.
\end{defn}

\subsection{The spectral norm of random matrices}

Throughout this paper we have used the following standard result from random
matrix theory:

\begin{theorem}[Spectral norm of random matrices, \cite{Van_H2007-ca}]
    \label{result:spectralnorm}
    There are constants $C$ and $C'$ such that the following holds.
    Let $H$ be an $n \times n$ symmetric random matrix whose entries satisfy,
    \[
        \expectation H_{ij} = 0,
        \qquad
        \expectation \left(H_{ij}\right)^2 \leq \sigma^2,
        \qquad
        \abs*{H_{ij}} \leq B.
    \]
    where $\sigma \geq C' n^{-\nicefrac12} B \log^2 n$. Then, almost surely:
    \[
        \spectralnorm{H}
        \leq
        2 \sigma \sqrt{n} 
        +
        C \sqrt{B \sigma}
        \cdot
        n^{1/4}
        \log n.
    \]
\end{theorem}
\label{proof:spectralnorm}

It can be shown that a similar lower bound holds in many cases. For instance,
when the entries of $H$ have the Gaussian distribution with unit variance, the
spectral norm of $H$ is not only $O(\sqrt{n})$, but $\Theta(\sqrt{n})$ with high
probability. Since we typically use $\spectralnorm{H}$ to obtain an upper-bound
on the size of the perturbation, we will not need this result.

\subsection{Proof of \Cref{result:hoeffding}}

\restatehoeffding*{}

\begin{proof}
    \label{proof:hoeffding}
    We have
    \[
        \langle u, H v \rangle
        =
            \sum_{i=1}^n
            \sum_{j=1}^n
                u_i
                H_{ij}
                v_j
        =
            \sum_{i=1}^n
                u_i v_i H_{ii}
            +
            \sum_{j > i}
                (u_i v_j + u_j v_i) H_{ij}.
    \]
    The right hand side is a sum of independent random variables. We therefore
    apply the Hoeffding inequality in its general form for sub-Gaussian random
    variables to obtain an upper bound (see Proposition 5.10 in
    \cite{Vershynin2010-we}). We find:
    \begin{align}
        \prob(
            \abs{\langle u, H v \rangle}
            \geq \gamma
        )
        &\leq\nonumber
            2 \exp
            \left\{
                - \frac{
                    \frac{1}{2}\gamma^2
                }{
                    \sum_{i=1}^n 
                    (u_i v_i \sigma_{ii})^2
                    +
                    \sum_{j > i} 
                    \left[(u_i v_j + u_j v_i) \sigma_{ij} \right]^2
                }
            \right\}
            ,\\
        &\leq\label{eqn:hoeffding}
            2 \exp
            \left\{
                - \frac{
                    \frac{1}{2}\gamma^2
                }{
                \sigma^2
                \left[
                    \sum_{i=1}^n 
                    (u_i v_i)^2
                    +
                    \sum_{j > i} 
                    (u_i v_j + u_j v_i)^2
                \right]
                }
            \right\}
            .
    \end{align}

    We have 
    \begin{equation}
        \label{eqn:norm1}
        \sum_{i = 1}^n (u_i v_i)^2
        \leq
        \sum_{i = 1}^n
        \sum_{j = 1}^n
        (u_i v_j)^2
        =
        \sum_{i = 1}^n
            u_i^2
        \sum_{j = 1}^n
            v_j^2
        =
        \|u\|_2^2 \cdot \|v\|_2^2
        =
        1.
    \end{equation}
    Similarly,
    \begin{align*}
        \sum_{j > i} (u_i v_j + u_j v_i)^2
        &\leq
            \sum_{j > i}
            \left[
                (u_i v_j)^2 + (u_j v_i)^2 + 2 \abs{u_i u_j v_i v_j}
            \right]
            ,\\
        &=
            \sum_{j > i}
                (u_i v_j)^2
            +
            \sum_{j > i}
                (u_j v_i)^2 
            +
            \sum_{j > i}
                + 2 \abs{u_i u_j v_i v_j}
            ,\\
        &\leq
            \sum_{i=1}^n
            \sum_{j=1}^n
                (u_i v_j)^2
            +
            \sum_{i=1}^n
            \sum_{j=1}^n
                (u_j v_i)^2 
            +
            \sum_{j > i}
                + 2 \abs{u_i u_j v_i v_j}
            .
    \intertext{The first two sums are each bounded by 1, as before:}
        &\leq
            2
            +
            \sum_{j > i}
                + 2 \abs{u_i u_j v_i v_j}
            ,\\
        &\leq
            2
            +
            \sum_{i=1}^n
            \sum_{j=1}^n
                \abs{u_i u_j v_i v_j}
            ,\\
        &=
            2
            +
            \sum_{i=1}^n
                \abs{u_i v_i}
            \sum_{j=1}^n
                \abs{u_j v_j}
            .
    \end{align*}
    Each sum is bounded by 1 by an application of Cauchy-Schwarz.
    Therefore we find that the total sum is bounded by 3. Substituting this and
    \Cref{eqn:norm1} into \Cref{eqn:hoeffding} we see that
    \[
        \prob(
            \abs{\langle u, H v \rangle}
            \geq \gamma
        )
        \leq
        2 \exp\left\{
            -\frac{\gamma^2}{8 \sigma^2}
        \right\}.
    \]

\end{proof}

\subsection{Proof of \Cref{result:spanh}}

\begin{restatable}{lemma}{restatespanh}
    \label{result:spanh}
    Let $\{\eigvec{1}, \ldots, \eigvec{d}\}$ be an orthonormal set of $d$
    vectors, and suppose that $\abs{\langle \eigvec{i}, H \eigvec{j} \rangle}
    \leq h$ for all $i ,j \in \countingset{d}$. Then 
    $
        \abs{
            \langle x, H x \rangle
        }
        \leq
        dh
    $
    for any unit vector $x \in
    \Vecspan({\eigvec{1}, \ldots, \eigvec{d}})$.
\end{restatable}

\begin{proof}
    \label{proof:spanh}
    Since $\eigvec{1}, \ldots, \eigvec{d}$ form an orthonormal basis for the
    space in which $x$ lies, we can expand $x$ as
    \[
        x = \sum_{i=1}^d \alpha_i \eigvec{i},
    \]
    where $\alpha_i = \langle x, \eigvec{i} \rangle$. Therefore:
    \begin{align*}
        \left\langle
            x, H x
        \right\rangle
        &=
            \left\langle
                \sum_{i=1}^d \alpha_i \eigvec{i},
                \;
                H
                \sum_{j=1}^d \alpha_j \eigvec{j}
            \right\rangle
            ,\\
        &=
            \sum_{i=1}^d
            \sum_{j=1}^d
                \alpha_i
                \alpha_j
                \left\langle
                    \eigvec{i},
                    H
                    \eigvec{j}
                \right\rangle
                ,\\
        &\leq
            h
            \sum_{i=1}^d
            \sum_{j=1}^d
            \,
                \abs{
                \alpha_i
                \alpha_j
                }
            ,\\
        &=
            h
            \left(
            \sum_{i=1}^d
                \abs{
                    \alpha_i
                }
            \right)
            \left(
            \sum_{j=1}^d
                \abs{
                \alpha_j
                }
            \right)
                .
    \intertext{Let $\alpha$ be the vector $(\alpha_1, \ldots,
    \alpha_d)^\transpose$. Then:}
        &=
            h
            \|\alpha\|_1^2
                .
    \end{align*}
    We know that $\twonorm{\alpha} = 1$ since $x$ is a unit vector. The 1-norm
    is bounded by $\sqrt{d}$ times the 2-norm. Hence
    $\|\alpha\|_1 \leq \sqrt{d} \cdot \twonorm{\alpha} = \sqrt{d}$. Hence 
    $\abs{\langle x, H x \rangle} \leq h d$.

\end{proof}

\section{Powers of random matrices and their interaction with delocalized
vectors}

We have seen that using the Neumann trick to bound the perturbation in
eigenvectors requires bounding series expansions of the form
\[
    \tailboundraw(u; H,\eigval) = 
    \sum_{p \geq 0} \left(\frac{H}{\eigval}\right)^p u,
\]
where $H$ is a random matrix. We have given one result in 
\Cref{result:boundzeta} which shows that the $\infty$-norm of this series is
small when $u$ has small $\infty$-norm. We now give two related results which
give finer \emph{entrywise} bounds on $\tailboundraw$.
We have not needed to use these results in the main paper, but we present them
here for completeness. We believe that they may be useful, for example, in the
analysis of stochastic blockmodels in which the community sizes scale at
different asymptotic rates.

The following \namecref{result:boundzetamag} is useful when the \emph{magnitude}
of $H_{ij}$ decreases like $\nicefrac{1}{n}$ -- this is as opposed to the
central moments decreasing like $\nicefrac{1}{n}$ as assumed in
\Cref{result:boundzeta}. Clearly this is a stronger condition, as there are
cases in which the variance decays with $n$ but the magnitude does not; the
random graph noise in the blockmodel setting is one such example. However, by
making the stronger assumption it is possible to localize the effect of $H$ to
the indices of $\tailboundraw$ which correspond to nonzero entries of $u$. That
is, if $u_i = 0$ for all $i$ in a set $F$, then $\tailboundraw_i$ is smaller for
$i \in F$ than for $i \not \in F$.

In this and what follows, \term{symmetric random matrix} has the precise meaning
as given in \Cref{defn:srm} above.

\begin{restatable}{theorem}{restateboundzetamag}
    \label{result:boundzetamag}
    Let $H$ be an $n \times n$ symmetric random matrix satisfying $\expectation
    H_{ij} = 0$. Suppose $\gamma$ is such that
    $
        \abs{H_{ij}/\gamma} \leq \nicefrac{1}{\sqrt{n}}
    $.
    Choose $\xi > 1$ and $\kappa \in (0,1)$. 
    Let $\eigval \in \reals$ and suppose that $\gamma < \eigval (\log n)^\xi$
    and $\eigval > \spectralnorm{H}$.
    Fix $u \in \reals^n$ and let $F = \{i : u_i \neq 0\}$.
    Define
    \[
        \beta_\ix =
        \begin{cases}
            1,& \ix \in F,\\
            \sqrt{\frac{\abs{F}}{n}},& \ix \not \in F.
        \end{cases}
    \]
    Then for all $\ix \in \countingset{n}$ simultaneously,
    \[
        \tailboundraw_\ix(H, \eigval, u)
        \leq
            \frac{\beta_\ix \, \gamma \, 
                (\log n)^\xi}{\eigval - \gamma \, (\log n)^\xi}
            \cdot \inftynorm{u}
            +
            \frac{
                \spectralnorm{H/\eigval}^{\lfloor 
                    \frac{\kappa}{8}(\log n)^\xi + 1
                \rfloor}
            }{
                1 - \spectralnorm{H/\eigval}
            }
            \cdot
            \twonorm{u}
            .
    \]
    with probability
    $
        1 
        - 
            n^{-\frac{1}{4} (\log_b n)^{\xi-1} (\log_b
            e)^{-\xi} + 1},
    $
    where 
    $b = \left(\frac{\kappa + 1}{2}\right)^{-1}$.
\end{restatable}

\begin{proof}
    Located in \Cref{proof:boundzetamag} on page \pageref{proof:boundzetamag}.
\end{proof}

A corollary of the above \namecref{result:boundzetamag} is the following which
applies specifically to vectors with block structure. We say that $u$ is an
$(n,K)$-block vector if it has $n$ elements which can be partitioned into $K$
groups such that the value of $u$ is homogeneous across the group. That is,
there exists a partition $F_1, \ldots, F_K$ of $\countingset{n}$ such that for
any $F_k$, if $i,j \in F_k$ then $u_i = u_j$.

\begin{restatable}{theorem}{restateboundzetablock}
    \label{result:boundzetablock}
    Let $H$ be an $n \times n$ symmetric random matrix satisfying $\expectation
    H_{ij} = 0$. Suppose $\gamma$ is such that
    $
        \abs{H_{ij}/\gamma} \leq \nicefrac{1}{\sqrt{n}}
    $.
    Choose $\xi > 1$ and $\kappa \in (0,1)$. 
    Let $\eigval \in \reals$ and suppose that $\gamma < \eigval (\log n)^\xi$
    and $\eigval > \spectralnorm{H}$.
    Let $F_1, \ldots, F_K$ be the $K$ blocks of a partition of
    $\countingset{n}$. If $u$ is an $(n, K)$-block vector with blocks $F_1,
    \ldots, F_K$, write $c_k(u)$ to denote the value that $u$ takes on block
    $F_k$.
    Define for all $\ix \in \countingset{n}$ and $k \in \countingset{K}$:
    \[
        \beta_{\ix,k} =
        \begin{cases}
            1,& \ix \in F_k,\\
            \sqrt{\frac{\abs{F_k}}{n}},& \ix \not \in F_k.
        \end{cases}
    \]
    Then for all $(n,K)$-block vectors $u$ and all $\ix \in \countingset{n}$
    simultaneously:
    \[
        \tailboundraw_\ix(H, \eigval, u)
        \leq
            \sum_{k=1}^K
                c_k(u)
                \cdot
                \left(
                    \frac{\beta_{\ix,k} \, \gamma \, 
                        (\log n)^\xi}{\eigval - \gamma \, (\log n)^\xi}
                    +
                    \sqrt{\abs{F_k}} \cdot
                    \frac{
                        \spectralnorm{H/\eigval}^{\lfloor 
                            \frac{\kappa}{8}(\log n)^\xi + 1
                        \rfloor}
                    }{
                        1 - \spectralnorm{H/\eigval}
                    }
                \right)
    \]
    with probability
    $
        1 
        - 
            K n^{-\frac{1}{4} (\log_b n)^{\xi-1} (\log_b
            e)^{-\xi} + 1},
    $
    where 
    $b = \left(\frac{\kappa + 1}{2}\right)^{-1}$.
\end{restatable}

\begin{proof}
    Located in \Cref{proof:boundzetablock} on page
    \pageref{proof:boundzetablock}.
\end{proof}

\newcommand{\otherr}{\tilde{r}}
\newcommand{\otherell}{\tilde{\ell}}
\newcommand{\thirdr}{\hat{r}}
\newcommand{\thirdell}{\hat{\ell}}
\newcommand{\specialr}{r^*}
\newcommand{\specialell}{\ell^*}
\newcommand{\singletonr}{r^*}
\newcommand{\singletonell}{\ell^*}
\newcommand{\singletonpair}{(\singletonr, \singletonell)}
\newcommand{\pfirst}{\operatorname{\mathsf{First}}}
\newcommand{\plast}{\operatorname{\mathsf{Last}}}
\newcommand{\pnext}{\operatorname{\mathsf{Next}}}
\newcommand{\pprev}{\operatorname{\mathsf{Prev}}}
\newcommand{\zeroblock}{\tilde{\block}}
\newcommand{\eqblock}[1][\partition]{\stackrel{#1}{\sim}}
\newcommand{\rlendset}{Q}
\newcommand{\elements}[1]{[#1]}
\newcommand{\indexings}{\mathcal{Z}_{p,k,\ix}}
\newcommand{\validindexings}{\mathcal{Z}^{+(F)}_{p,k,\ix}\{\partition\}}
\newcommand{\posindexings}{\mathcal{Z}_{p,k,\ix}^{\neq 0}}
\newcommand{\partitions}{{\mathcal{P}}_{p,k}}
\newcommand{\twinpartitions}{\partitions^{+(F,\alpha)}}
\newcommand{\partition}{\Gamma}
\newcommand{\block}{\gamma}

\subsection{Proof of the main interaction}

The proofs of
\Cref{result:boundzeta,result:boundzetamag,result:boundzetablock} depend heavily
on the following \Cref{result:interaction}. Part of the proof of 
\Cref{result:interaction} is due to \cite{Erdos2011-ol}. We have amended this
proof to provide precise bounds on the probability of the event. Moreover, the
proof of the second part of the following result (when the magnitude of $H_{ij}$
is small) is novel and possibly of independent interest.

\begin{restatable}{theorem}{thminteraction}
    \label{result:interaction}
    Let $\rv{X}$ be a symmetric and centered random matrix of size $n \times n$. 
    Let $u$ be an $n$-vector with $\inftynorm{u} = 1$.
    Choose $\xi > 1$ and $0 < \kappa < 1$.
    Define $\mu = \left(\frac{\kappa + 1}{2}\right)^{-1}$. 
    Then with probability
    $
            1 - 
            n^{-\frac{1}{4} (\log_\mu n)^{\xi-1} (\log_\mu
            e)^{-\xi}}
    $,
    for any
    $k \leq \frac{\kappa}{8} (\log n)^\xi$:
    \begin{enumerate}
        \item If $\expectation \abs{\rv{X}_{ij}}^p \leq \frac{1}{n}$ for all $p
            \geq 2$, we have
            \[
                \abs*{\left(\rv{X}^k u \right)_\ix}
                <
                \left(\log n\right)^{k \xi}.
            \]
        \item 
            Let $F = \{ i : u_i \neq 0 \}$.
            If $\abs{\rv{X}_{ij}} \leq \nicefrac{1}{\sqrt{n}}$, we have
            \[
                \abs*{\left(\rv{X}^k u \right)_\ix}
                <
                \left(\log n\right)^{k \xi} \cdot
                \begin{cases}
                    1,
                    & \alpha \in F,\\
                    \sqrt{\frac{|F|}{n}},
                    & \alpha \not \in F.
                \end{cases}
            \]
    \end{enumerate}
\end{restatable}

\newcommand{\elem}{\left(\rv{X}^k u\right)_\ix}

\begin{proof}
    \label{proof:maininteraction}
    We will bound $\abs*{\elem}$ with a high-moment Markov inequality. Let $p$
    be a positive even integer. Then
    \begin{equation}
        \label{eqn:markov}
        \prob\left(
            \abs*{\elem}
        \right) \leq 
        \frac{\expectation 
            \left[
                \elem^p
            \right]} {t^p}.
    \end{equation}
    Bounding the expectation is non-trivial. We will utilize the following
    lemmas whose extensive proofs are to be found in the next subsection.

    \begin{restatable}{lemma}{lemmamexpone}
        \label{lemma:mexp1}
        If $\expectation \left[\abs*{\rv{X}_{ij}}^s\right] \leq \nicefrac{1}{n}$
        for all $s
        \geq 2$, then
        \[
            \expectation 
            \left[
                \elem^p
            \right]
            \leq 
            (2 pk)^{pk}.
        \]
    \end{restatable}

    \begin{restatable}{lemma}{lemmamexptwo}
        \label{lemma:mexp2}
        If $\abs*{\rv{X}_{ij}} \leq \nicefrac{1}{\sqrt{n}}$
        and $F = \{i : u_i \neq 0\}$, then
        \[
            \expectation 
            \left[
                \elem^p
            \right]
            \leq 
            (2 pk)^{pk}
            \cdot
            \begin{cases}
                1,& \ix \in F,\\
                \left(\frac{|F|}{n}\right)^{\nicefrac{p}{2}},& \ix \not \in F.
            \end{cases}
        \]
    \end{restatable}

    The assumptions of the first lemma are weaker than the second, but we can
    consider both cases simultaneously by defining
    \[
        B_\ix = 
        \begin{cases}
            \sqrt{\frac{|F|}{n}}
                ,& \abs*{\rv{X}_{ij}} \leq \nicefrac{1}{\sqrt{n}}
                \text{ and } \ix \in F,\\
            1,& \text{ otherwise }.
        \end{cases}
    \]
    Then in both cases: 
    \[
            \expectation 
            \left[
                \elem^p
            \right]
            \leq
            B_\ix^p (2pk)^{pk}.
    \]
    Returning to the Markov inequality in \Cref{eqn:markov},
    we will choose $t = B_\ix (\log n)^{k\xi}$, giving:
    \begin{align*}
        \prob\left(\abs*{\elem} \geq 
                B_\ix (\log n)^{k \xi}
            \right) 
            &\leq 
            \frac{\expectation \left[\elem^p \right]}
            {\left[B_\ix (\log n)^{k \xi}\right]^p},\\
            &=
                \frac{
                    B_\ix^p (2pk)^{pk}
                }{
                    B_\ix^p (\log n)^{p k \xi}
                },\\
            &=
            \left[
                \frac{2pk}{(\log n)^\xi}
            \right]^{pk}.
    \end{align*}
    \newcommand{\realp}{\tilde{p}}%
    \newcommand{\ourp}{\hat{p}}%
    The bound above holds for any positive even
    integer $p$. We will choose $p = \ourp$, where $\ourp$ is the smallest even
    integer greater than or equal to $\realp = \frac{1}{4k} (\log n)^\xi$. Since
    $k < \frac{1}{8}(\log n)^\xi$, we have $\realp \geq 2$, and so $\ourp \geq
    2$.  Furthermore, we have $\ourp = \realp + \delta$, where $0 \leq \delta <
    2$.  Hence:
    \begin{align*}
        \left[
            \frac{2\ourp k}{(\log n)^\xi}
        \right]^{\ourp k}
        &=
            \left[
                \frac{2(\realp + \delta) k}{(\log n)^\xi}
            \right]^{(\realp + \delta) k},\\
        &=
            \left[
                \frac{2(\realp + \delta) k}{(\log n)^\xi}
            \right]^{\realp k}
            \cdot
            \left[
                \frac{2(\realp + \delta) k}{(\log n)^\xi}
            \right]^{\delta k}.
    \intertext{We see that $2 \realp k / (\log n)^\xi = \nicefrac12$,
    hence:}
        &=
            \left[
                \frac12 +
                \frac{2 \delta k}{(\log n)^\xi}
            \right]^{\realp k}
            \cdot
            \left[
                \frac12 +
                \frac{2 \delta k}{(\log n)^\xi}
            \right]^{\delta k}.
    \intertext{Because $0 \leq \delta < 2$, we have $\frac{1}{2} < \frac{1}{2} +
    \frac{2 \delta k}{(\log n)^\xi} < 1$. And since $\delta k > 0$,
    the second term in the above is at most $1$. Therefore:}
        &\leq
            \left[
                \frac12 +
                \frac{2 \delta k}{(\log n)^\xi}
            \right]^{\realp k}.
    \intertext{Using $\delta < 2$ and substituting the definitions of $\realp$
    and $k$, we arrive at:}
        &\leq
            \left[
                \frac12 +
                \frac{4 k}{(\log n)^\xi}
            \right]^{\frac{1}{4} (\log n)^\xi},\\
        &<
            \left[
                \frac{\kappa + 1}{2}
            \right]^{\frac{1}{4} (\log n)^\xi}.
    \intertext{We recognize the base of the exponent as $\mu^{-1}$, therefore:}
        &=
            \mu^{-\frac{1}{4} (\log n)^\xi},\\
        &=
            \mu^{-\frac{1}{4} (\log_\mu n)^\xi (\log_\mu e)^{-\xi}},\\
        &=
            \mu^{-\frac{1}{4} (\log_\mu n) (\log_\mu n)^{\xi - 1} (\log_\mu
            e)^{-\xi}},\\
        &=
            n^{-\frac{1}{4} (\log_\mu n)^{\xi - 1} (\log_\mu
            e)^{-\xi}}.
    \end{align*}
    Therefore:
    \begin{align*}
        \prob\left(\abs*{\elem} \geq 
                B_\ix (\log n)^{k \xi}
            \right) 
        &\leq
        \left[
            \frac{2\ourp k}{(\log n)^\xi}
        \right]^{\ourp k},\\
        &\leq
            n^{-\frac{1}{4} (\log_\mu n)^{\xi-1} (\log_\mu
            e)^{-\xi}}.
    \end{align*}
\end{proof}

\subsection{Proofs of moment bounds: \Cref{lemma:mexp1,lemma:mexp2}}

In this subsection we derive bounds on $\expectation \left[\elem^p \right]$
under different assumptions on the entries of $\rv{X}$.  In particular, we will
prove \Cref{lemma:mexp1,lemma:mexp2} which are critical components of
\Cref{result:interaction}. 

\subsubsection{Some useful results}

    First we derive a formalism for working with moments of
    random matrix products.
    It follows from the definition of matrix multiplication that
    the $\ix$th element of the vector $\rv{X}^k u$ has the expansion:
    \[
        \left(\rv{X}^k u\right)_\ix =
            \sum_{i_1, \ldots, i_k}
            \rv{X}_{\ix i_1} \rv{X}_{i_1 i_2} \cdots \rv{X}_{i_{k-1}i_k} u_{i_k}.
    \]
    As a result, we have:
    \begin{align*}
        \expectation \left[(\rv{X}^k u)_\ix^p\right]
        &=
            \expectation 
            \left[
            \left(
                \sum_{i_1, \ldots, i_k}
                \rv{X}_{\ix i_1} \rv{X}_{i_1 i_2} \cdots 
                    \rv{X}_{i_{k-1}i_k} u_{i_k}
            \right)^p
            \,
            \right]
            ,\\
        &=
            \expectation 
            \left[
            \sum_{i_1^{(1)}, \ldots, i_{k}^{(1)}}
            \cdots
            \sum_{i_1^{(p)}, \ldots, i_{k}^{(p)}}
            \,
            \prod_{r=1}^p
                \rv{X}_{\ix i_1^{(r)}} 
                \rv{X}_{i_1^{(r)} i_2^{(r)}} 
                \cdots 
                \rv{X}_{i_{k-1}^{(r)} i_{k}^{(r)}}
                u_{i_{k}^{(r)}}
            \right]
                .\\
    \end{align*}
    Here there are $p$ summations, each over an independently-varying
    set of $k$ variables $i_1^{(r)}, \ldots, i_k^{(r)}$ which range from
    $1$ to $n$.
    We replace the variables of summation with indexing
    functions, defined as follows. 

    \begin{defn}
        For positive integers $p$ and $k$ and an index $\ix \in
        \countingset{n}$, a \term{$(p,k,\ix)$-indexing function} is a discrete
        map $\tau : \countingset{p} \times \{0,\ldots,k\} \to \countingset{n}$
        satisfying $\tau(r,0) = \alpha$ for all $r \in \countingset{p}$.
    \end{defn}

    An indexing function $\tau$ corresponds to a single configuration of the
    variables of summation in the expectation above. That is, we may interpret
    $\tau(r,\ell)$ as the value of the variable $i_\ell^{(r)}$ in a particular
    configuration. As such, we will use the shorthand notation $\tau_\ell^{(r)}
    = \tau(r, \ell)$ so that $\rv{X}_{i_{\ell - 1}^{(r)} i_{\ell}^{(r)}}$ is
    replaced by $\rv{X}_{\tau_{\ell - 1}^{(r)} \tau_{\ell}^{(r)}}$.

    Let $\indexings$ be the set of all $(p, k, \alpha)$-index functions.
    The above expectation can be written as:
    \begin{align*}
        \expectation \left[(\rv{X}^k u)_\ix^p\right]
        &=
            \expectation
            \left[
            \sum_{\tau \in \indexings}
            \,
            \prod_{r=1}^p
            \,
                \rv{X}_{\tau_{0}^{(r)} \tau_{1}^{(r)}}
                \rv{X}_{\tau_{1}^{(r)} \tau_{2}^{(r)}}
                \cdots
                \rv{X}_{\tau_{k-1}^{(r)} \tau_{k}^{(r)}}
                u_{\tau_{k}^{(r)}}
            \right]
                ,\\
        &=
            \sum_{\tau \in \indexings}
            \,
            \expectation
            \left[
            \,
            \prod_{r=1}^p
            \,
            \rv{X}_{\tau_{0}^{(r)} \tau_{1}^{(r)}}
            \rv{X}_{\tau_{1}^{(r)} \tau_{2}^{(r)}}
                \cdots
                \rv{X}_{\tau_{k-1}^{(r)} \tau_{k}^{(r)}}
                u_{\tau_{k}^{(r)}}
            \right]
            ,\\
        &\leq
            \sum_{\tau \in \indexings}
            \,
            \abs*{
                \,
            \expectation
            \left[
            \,
            \prod_{r=1}^p
            \,
            \rv{X}_{\tau_{0}^{(r)} \tau_{1}^{(r)}}
            \rv{X}_{\tau_{1}^{(r)} \tau_{2}^{(r)}}
                \cdots
                \rv{X}_{\tau_{k-1}^{(r)} \tau_{k}^{(r)}}
                u_{\tau_{k}^{(r)}}
            \right]
            }
            ,\\
        &=
            \sum_{\tau \in \indexings}
            \,
            \underbrace{
            \abs*{
            \left(
                \,
            \prod_{r=1}^p
                u_{\tau_k^{(r)}}
            \right)
            }
            }_{\omega_u(\tau)}
            \cdot
            \underbrace{
            \abs*{
                \,
            \expectation
            \left[
            \,
            \prod_{r=1}^p
            \,
            \rv{X}_{\tau_{0}^{(r)} \tau_{1}^{(r)}}
            \rv{X}_{\tau_{1}^{(r)} \tau_{2}^{(r)}}
                \cdots
                \rv{X}_{\tau_{k-1}^{(r)} \tau_{k}^{(r)}}
            \right]
            }
            }_{\varphi(\tau)},\\
        &=
            \sum_{\tau \in \indexings}
            \omega_u(\tau) \cdot \varphi(\tau)
            \stepcounter{equation}\tag{\theequation}\label{eqn:summand}.
    \end{align*}
    Here we write $\omega_u$ to show that $\omega_u$ is parametrized by the
    vector $u$. On the other hand, $\varphi$ does not depend on $u$.
    In the following two parts, we derive bounds on this quantity under
    assumptions on the magnitude or variance of $\rv{X}_{ij}$. In each case the
    core approach is the same: we bound the size of $\varphi(\tau)$ for any
    $\tau$ by using the assumptions on $\rv{X}$, and then bound the number of
    $\tau$ for which $\varphi$ and $\omega_u$ are non-zero.

    The entries in the upper-triangle of the random matrix $\rv{X}$ are
    independent, but not necessarily identically distributed. Rather, the
    assumptions that we will place on the entries of $\rv{X}$ will not depend on
    the indices. As a result, it is not important to use the precise knowledge
    of which entries of $\rv{X}$ are selected by an indexing function $\tau$ in
    order to bound $\varphi(\tau)$. We will therefore partition the set of
    indexing functions into equivalence classes which characterize the important
    structure of the indexing, and then derive a bound for each
    equivalence class independently.

    First, some notation: For a set of sets $A$, we write
    $\elements{A}$ to denote the union of all elements of $A$; i.e.,
    $\elements{A} = \bigcup_{\block \in A} \block$. We introduce the following
    notion:

    \begin{defn}
        \label{defn:index_partition}
        A \term{$(p,k)$-index partition} $\partition$ is a partition of a subset
        of $\{1,\ldots,p\} \times \{0,\ldots,k\}$ with the property that there
        exists a block $\zeroblock \in \partition$ such that every pair of the
        form $(r,0)$ is in $\zeroblock$; that is:
        \[
            \exists \zeroblock \in \partition
            \text{ s.t. }
            \zeroblock \supset
            \left\{(r,0) : r \in \{1,\ldots,p\}\right\}.
        \]
        We call $\zeroblock$ the \term{root block} of $\partition$. 
    \end{defn}
    Note that a $(p,k)$-index partition is a partition of a \emph{subset} of
    $\countingset{p} \times \{0,\ldots,k\}$; i.e., it is not necessarily the
    case that $\elements{\partition}$ is the full set $\countingset{p} \times
    \{0,\ldots,k\}$. For example, any $(p-1,k-1)$-index partition is also a
    $(p,k)$-index partition by definition. We will later find it useful to make
    use of such ``subpartitions'', but for the time being we will only consider
    index partitions which in fact partition the full set.
    Let $\partitions$ be the set of all ``full'' $(p,k)$-index partitions
    $\partition$ such that $\elements{\partition} = \countingset{p} \times
    \{0,\ldots,k\}$.

    Next, note that an index partition $\partition \in \partitions$ defines an
    equivalence relation on $\elements{\partition}$. We use the following
    notation to denote this relation:

    \begin{notation}
    For pairs $(r,\ell), (\otherr, \otherell) \in \elements{\partition}$ we
    write $(r, \ell) \eqblock (\otherr, \otherell)$ if and only if there
    exists a block $\block \in \partition$ such that $\block$ contains both
    $(r,\ell)$ and $(\otherr, \otherell)$. 
    \end{notation}
    
    We relate indexing functions and index partitions in the following way:
    \begin{defn}
    We say that an indexing function
    $\tau$ \emph{respects} the partition $\Gamma \in \partitions$ when
    $
        \tau_\ell^{(r)} = \tau_{\ell'}^{(r')}
    $
    if and only if
    $
        (r,\ell) \eqblock (r', \ell').
    $
    \end{defn}

    It is clear that for any indexing function $\tau$, there is exactly one
    partition $\Gamma \in \partitions$ such that $\tau$ respects $\Gamma$.
    As such, we have implicitly established an equivalence relation between
    indexing functions: $\tau$ and $\tau'$ are equivalent if and only if they
    respect the same index partition.
    For an index partition $\partition \in \partitions$, write
    $\indexings\{\partition\}$ to denote the set of all indexing functions which
    respect $\partition$. Then \Cref{eqn:summand} can be re-written as:
    \begin{equation}
        \label{eqn:summand2}
        \expectation\left[
            \elem^p 
        \right]
        \leq
            \sum_{\partition \in \partitions}
            \,
            \sum_{\tau \in \indexings\{\partition\}}
            \omega_u(\tau)
            \cdot
            \varphi(\tau).
    \end{equation}

    \begin{defn}[Twin property]
        Let $\partition \in \partitions$. Let $(r,\ell) \in
        \elements{\partition}$ and $(\otherr, \otherell) \in
        \elements{\partition}$ be distinct and such that $\ell,\otherell > 0$.
        We say that $(r,\ell)$ and $(\otherr,\otherell)$ are \term{twins in
        $\partition$} if either:
        \begin{enumerate}
            \item $(r, \ell) \eqblock (\otherr, \otherell)$ 
                and 
                $(r, \ell - 1)
                \eqblock 
                (\otherr,
                \otherell-1))$; or
            \item $(r, \ell) \eqblock (\otherr, \otherell - 1)$
                and 
                $(r, \ell - 1)
                \eqblock 
                (\otherr, \otherell)$.
        \end{enumerate}
        We say that a $(p,k)$-index partition $\partition$ satisfies the
        \term{twin property} if for any pair $(r,\ell) \in
        \elements{\partition}$ with $\ell > 0$ there exists a distinct
        $(\otherr, \otherell) \in \elements{\partition}$ with $\otherell > 0$
        such that $(r,\ell)$ and $(\otherr, \otherell)$ are twins in
        $\partition$.
    \end{defn}

    \begin{lemma}
        \label{lemma:twinsiffrveq}
        Let $\tau$ be an indexing function respecting the partition
        $\partition$. 
        Then $(r, \ell)$ and $(\otherr, \otherell)$ are twins in $\partition$ if
        and only if 
        $
        \rv{X}_{\tau_{\otherell-1}^{(\otherr)} \tau_{\otherell}^{(\otherr)}} 
        =
        \rv{X}_{\tau_{\ell-1}^{(r)} \tau_{\ell}^{(r)}} 
        $.
    \end{lemma}

    \begin{proof}
        Due to the symmetry of $\rv{X}$ and the independence of its entries
        along the upper triangle, we have that
        for any indices $i, j, i', j'$,  $\rv{X}_{ij} = \rv{X}_{i'j'}$ if
        and only if either 1) $(i,j) = (i',j')$ or 2) $(i,j) = (j',i')$.
        This is the case if and only if
        $\tau_{\otherell-1}^{(\otherr)} =  \tau_{\ell-1}^{(r)}$
        and
        $\tau_{\otherell}^{(\otherr)} =  \tau_{\ell}^{(r)}$
        or 2)
        $\tau_{\otherell-1}^{(\otherr)} =  \tau_{\ell}^{(r)}$
        and
        $\tau_{\otherell}^{(\otherr)} =  \tau_{\ell-1}^{(r)}$
        .
        The fact that that this holds if and only if $(r,\ell)$ and $(\otherr,
        \otherell)$ are twins follows from the definition of twins and the
        notion of $\tau$ respecting the partition $\partition$.
    \end{proof}

    \newcommand{\twinclasses}{T\{\partition\}}
    \newcommand{\twinblock}{\rho}

    \begin{defn}
        For any index partition $\partition \in \partitions$, denote by
        $T\{\partition\}$ the set of equivalence classes of the \emph{twin
        relation}, defined on $\countingset{p} \times \countingset{k}$ by $(r,
        \ell) \sim (\otherr, \otherell)$ if and only if $(r,\ell)$ and
        $(\otherr, \otherell)$ are twins.
    \end{defn}

    \begin{notation}
        If $\tau$ is an indexing function which respects $\partition$ and
        $\twinblock \in \twinclasses$, we write $\rv{X}_\twinblock$ to denote
        the random variable $\rv{X}_{ij}$ such that $\rv{X}_{ij} =
        \rv{X}_{\tau_{\ell - 1}^{(r)} \tau_\ell^{(r)}}$ for every $(r,\ell) \in
        \twinblock$; this is well-defined as a result of
        \Cref{lemma:twinsiffrveq}.
    \end{notation}

    \begin{lemma}
        \label{result:phifactor}
        Let $\partition \in \partitions$ and suppose $\tau$ is an indexing
        function which respects $\partition$. Then:
        \[
            \varphi(\tau) = 
            \prod_{\twinblock \in \twinclasses}
            \abs*{
            \expectation
            \left[
                \rv{X}_{\twinblock}^{\abs{\twinblock}}
            \right]
            }
            .
        \]
    \end{lemma}

    \begin{proof}
        \Cref{lemma:twinsiffrveq} implies that the equivalence classes of the
        twin relation partition the $pk$ terms of the product in $\varphi$ into
        sets of random variables which are equal. Since the entries of $\rv{X}$
        are independent random variables, the expectation factors.
    \end{proof}

    Since $\expectation \rv{X} = 0$, we have the following corollary:

    \begin{corollary}
        \label{lemma:twinpartitions}
        Suppose that $\partition \in \partitions$ does not satisfy the twin
        property; i.e., there exists a pair $(\thirdr, \thirdell) \in
        \elements{\partition}$ that does not have a twin in $\partition$. Then
        $\varphi(\tau) = 0$ for every $\tau$ respecting $\partition$.
    \end{corollary}

    \Cref{lemma:twinpartitions} implies that only partitions satisfying the
    twin property contribute to the sum in \Cref{eqn:summand2}.

    \begin{lemma}
        Let $F = \{i : u_i \neq 0 \}$.
        Fix $\ix \in \countingset{n}$.
        Suppose that $\partition \in \partitions$ is such that the root block
        $\zeroblock$ contains an element of the form $(r,k)$ for some $r \in
        \countingset{p}$. Then if $\ix \not \in F$ we have $\omega_u(\tau) =
        0$ for every $\tau$ which respects $\partition$.
    \end{lemma}

    \begin{proof}
        By the definition of an indexing function, $\tau_0^{(r)} = \ix$ for
        every $r \in \countingset{p}$. 
        Let $\specialr$ be such that $(\specialr,k) \in \zeroblock$.
        If $\tau$ respects $\partition$, then it is necessarily the case that
        $\tau_{k}^{(r^*)} = \tau_{0}^{(r^*)} = \ix$. 
        Then $u_{\tau_{k}^{(\specialr)}} = u_\ix$. If $\alpha \not \in F$, then
        $u_\ix = 0$ and hence $\omega_u(\tau) = 0$.
    \end{proof}

    \begin{defn}
        Fix a set $F \subset \countingset{n}$ and an index $\ix \in
        \countingset{n}$. We write
        $\twinpartitions$ to denote the set of all $\partition \in \partitions$
        such that
        \begin{enumerate}
            \item $\partition$ satisfies the twin property; and 
            \item if $\alpha \not \in F$, the root block $\zeroblock \in
                \partition$ contains no elements of the form $(r,k)$.
        \end{enumerate}
    \end{defn}
    
    The partitions in $\twinpartitions$ do not contribute to 
    \Cref{eqn:summand2}. Hence:
    \newcommand{\partitionbound}{B}%
    \begin{align*}
        \expectation\left[
            \elem^p 
        \right]
        &\leq
            \sum_{\partition \in \twinpartitions}
            \,
            \sum_{\tau \in \indexings\{\partition\}}
            \omega_u(\tau)
            \cdot
            \varphi(\tau).
    \end{align*}
    It is necessary for a partition $\partition$ to be an element of
    $\twinpartitions$ in
    order for a $\tau$ respecting it to be such that $\varphi(\tau) \neq 0$,
    however this is not a sufficient condition. Suppose that $\tau_{k}^{(r)}
    \not \in F$ for some $r$. Then $u_{\tau_{k}^{(r)}} = 0$ and hence
    $\omega_u(\tau) = 0$. Therefore, we can restrict ourselves to considering
    $\tau$ which map $(r,k)$ to $F$.
    Define:
    \[
        \validindexings
        = 
        \{
            \tau \in \indexings\{\partition\}
            :
            \tau_{k}^{(r)} \in F
            \quad
            \forall
            r \in \countingset{p}
        \}.
    \]
    Then:
    \begin{align*}
        \expectation\left[
            \elem^p 
        \right]
        &\leq
            \sum_{\partition \in \twinpartitions}
            \,
            \sum_{\tau \in \validindexings}
            \omega_u(\tau)
            \cdot
            \varphi(\tau).
    \end{align*}
    Fix $\partition \in \partitions$. Suppose that $\varphi(\tau) \leq
    \Phi_\partition$ for any $\tau \in
    \validindexings$. Furthermore, suppose that
    $
        \abs{\validindexings}
        \leq
        Z_\partition.
    $
    Note that $\omega_u(\tau) \in [0, 1]$, since it is the product of magnitudes
    of entries of $u$ and $\inftynorm{u} = 1$.
    Therefore:
    \begin{align*}
        \expectation\left[
            \elem^p 
        \right]
        &\leq
            \sum_{\partition \in \twinpartitions}
            Z_\partition
            \cdot
            \Phi_\partition.
    \intertext{If $Z_\partition \cdot \Phi_\partition \leq \partitionbound$ for
    all $\partition \in \twinpartitions$, then:}
        &\leq
            \sum_{\partition \in \twinpartitions}
            \partitionbound,\\
        &=
            \abs*{\twinpartitions}
            \cdot
            \partitionbound.
    \end{align*}
    We can bound the number of partitions loosely using the following lemma:

    \begin{lemma}
        \label{lemma:countpartitions}
        $|\partitions| \leq (2pk)^{pk}$.
    \end{lemma}

    \begin{proof}
        \newcommand{\restrictedpartitions}{\partitions'}

        Let $\restrictedpartitions$ be the set of all partitions of $\{1,
        \ldots, p\} \times \{1, \ldots, k\}$. The number of such partitions is
        the $pk$-th Bell number; a well-known bound gives
        $|\restrictedpartitions| \leq (pk)^{pk}$. We generate $\partitions$ from
        $\restrictedpartitions$ in the following way: For every $\partition \in
        \restrictedpartitions$, we
        \begin{enumerate}
            \item Create a new block $\zeroblock = \{(r,0) : r \in
                \{1,\ldots,p\}\}$.
            \item For every element $(r,\ell)$ in $\{1,\ldots,p\} \times
                \{1,\ldots,k\}$, make an independent decision about
                whether to move $(r,\ell)$ from the block of $\partition$
                containing it to the new block $\zeroblock$. 
                There are $2^{pk}$ possible ways of deciding which elements to
                move, and so there are $2^{pk}$ partitions of $\{1,\ldots,p\}
                \times \{0,\ldots,k\}$ generated from $\partition$.
        \end{enumerate}

        For each partition $\partition \in \restrictedpartitions$ we generate
        $2^{pk}$ partitions; in total, we generate $2^{pk} \cdot
        |\restrictedpartitions| = (2pk)^{pk}$. It is clear that $\partitions$ is
        a subset of the generated partitions.
        Since some of the partitions
        generated from $\partition$ and a distinct partition $\partition'$
        will be identical, $(2pk)^{pk}$ is only an upper-bound on
        $|\partitions|$.
    \end{proof}

    Since $\twinpartitions \subset \partitions$, we have $\abs*{\twinpartitions}
    \leq (2pk)^{pk}$. We have therefore derived the following result:
    \begin{lemma}
        \label{lemma:starting}
        Fix a vector $u$ and let $F = \{i : u_i \neq 0\}$.
        Fix an index $\ix \in
        \countingset{n}$.
        For an index partition $\partition \in
        \twinpartitions$, 
        suppose that $\varphi(\tau) \leq
        \Phi_\partition$ for any $\tau \in
        \validindexings$, and that $\abs{\validindexings}
        \leq Z_\partition$. If $Z_\partition \cdot \Phi_\partition \leq
        \partitionbound$ for
        all $\partition \in \twinpartitions$, then:
        \[
        \expectation\left[
            \elem^p 
        \right]
        \leq
        (2pk)^{pk} \cdot \partitionbound.
        \]
    \end{lemma}
    We will use this result as a starting point for proving
    \Cref{lemma:mexp1,lemma:mexp2}.
    In the next two parts, we will derive $\partitionbound$ under different
    assumptions on the entries of $\rv{X}$.

    \begin{lemma}
        \label{result:numberoftau}
        Fix a vector $u$ and let $F = \{i : u_i \neq 0\}$. 
        Let $\partition \in \twinpartitions$.
        Then
        \[
            \abs*{\validindexings} \leq n^{\abs{\partition} - 1}.
        \]
        Moreover, let $Q \subset \partition$ be the set of blocks in
        $\partition$ which contain an element of the form $(r,k)$ for some $r
        \in \countingset{p}$. 
        Suppose that $\ix \not \in F$.
        Then:
        \[
            \abs*{\validindexings} 
            \leq 
            n^{\abs{\partition} - \abs{Q} - 1}
            \cdot
            \abs{F}^{\abs{Q}}
            .
        \]
    \end{lemma}

    \begin{proof}
        By definition, $\tau_\ell^{(r)} = \tau_{\otherell}^{(\otherr)}$ if and
        only if $(r, \ell) \eqblock (\otherr, \otherell)$. Hence an indexing
        function $\tau$ respecting $\partition$ takes a distinct value on each
        $\block \in \partition$. Exactly one block of the partition 
        contains the pairs of the form $(r,0)$, and on this block $\tau$ must
        take the value $\alpha$. On the remaining $\abs{\partition - 1}$ blocks
        $\tau$ takes a value in $\countingset{n}$. Ignoring the constraint that
        these values be distinct between blocks to obtain an upper bound, there
        are $n^{\abs{\partition} - 1}$ possible choices for the values of $\tau$
        on these blocks; this gives the desired upper bound.

        For the second part, recognize that since $\tau \in \validindexings$ we
        have
        $\tau_k^{(r)} \in F$ by assumption.
        Hence the number of possible values which $\tau$ may take on a
        block in $Q$ is bounded above by $\abs{F}$. Furthermore, it is true that
        $Q$ does not contain the root block of the partition -- this follows
        from the definition of $\twinpartitions$ and the assumption that $\ix
        \not \in F$. The result then follows immediately.
    \end{proof}

\subsubsection{Proof of \Cref{lemma:mexp1}}

In this part, we will bound $\expectation\left[ \elem^p \right]$ under the
assumption that $\expectation \left[ \abs*{\rv{X}_{ij}}^s\right] \leq
\nicefrac{1}{n}$ for all $s \geq 2$. As per \Cref{lemma:starting}, it is
sufficient to bound $Z_\partition \cdot \Phi_\partition$ for all partitions
$\partition$ satisying the twin property.
In the following two lemmas, let $\partition \in \twinpartitions$ and suppose
that $\expectation \left[ \abs*{\rv{X}_{ij}}^s\right] \leq \nicefrac{1}{n}$ for
all $s \geq 2$. 

\begin{lemma}
    \label{result:mexp1bound1}
    For any $\tau \in \validindexings$
    we have $\varphi(\tau) \leq \Phi_\partition$, where
    $
        \Phi_\partition
        =
        n^{-\abs{\twinclasses}}.
    $
\end{lemma}
\begin{proof}
    As a result of \Cref{result:phifactor}:
    \begin{align*}
        \varphi(\tau) 
        & = \prod_{\twinblock \in \twinclasses}
            \abs*{
            \expectation\left[
                \rv{X}_\twinblock^{\abs{\twinblock}}
            \right]
            }
            .
    \intertext{We upper bound this by:}
        &\leq 
            \prod_{\twinblock \in \twinclasses}
            \expectation\left[
                \abs{\rv{X}_\twinblock}^{\abs{\twinblock}}
            \right].
    \intertext{Since $\partition$ satisfies the twin property we have
    $\abs{\twinblock} \geq 2$. Then $\expectation \left[
    \abs{\rv{X}_\twinblock}^{\abs{\twinblock}} \right] \leq \nicefrac{1}{n}$ by
    assumption, and so:}
        &\leq 
            \prod_{\twinblock \in \twinclasses}
            n^{-1},\\
        &=
            n^{-\abs{\twinclasses}}.
    \end{align*}
\end{proof}

\begin{lemma}
    \label{result:mexp1bound2}
    We have $\abs{\validindexings} \leq Z_\partition$, where
    $
        Z_\partition = n^{\abs{\twinclasses}}.
    $
\end{lemma}

\begin{proof}
    From \Cref{result:numberoftau} we have
    $
        \abs{\validindexings}
        \leq 
        n^{\abs{\partition} - 1}.
    $
    We now show that $\abs{\partition} - 1 \leq \abs{\twinclasses}$.
    It is sufficient to find an injection from the set $V \subset \partition$ of
    non-root blocks of $\partition$ to $\twinclasses$; The existence of an
    injection proves that $\abs{V} \leq \abs{\twinclasses}$, and since
    $\partition$ has exactly one root block it follows that $\abs{\partition} -
    1 \leq \abs{\twinclasses}$. 
    We construct an injection $g : V \to
    \twinclasses$ as follows. For any block $\block \in \partition$, let $\min
    \block$ be the pair $(\specialr,\specialell) \in \block$ which is the
    minimum element with respect to the natural lexicographical order. That is,
    $(\specialr, \specialell) \in \block$ is the pair such that for any other
    $(r, \ell) \in \block$, either $r > \specialr$ or it is the case that both
    $r = \specialr$ and $\ell > \specialell$. The injection $g$ is defined by:
    \[
        g : \block  \mapsto \text{the equivalence class $\twinblock \in
        \twinclasses$ containing $\min \block$}.
    \]
    
    First note that this is a function since $\twinclasses$ partitions the set
    $\countingset{p} \times \countingset{k}$ such that $g(\gamma)$ is uniquely
    defined. Next we show that it is indeed an injection. Suppose for a
    contradiction that $\block$ and $\block'$ are distinct members of
    $\partition$ and that $g(\block) = g(\block')$. Let $(r,\ell) = \min \block$
    and $(r',\ell') = \min \block'$, and assume (without loss of generality)
    that $(r,\ell) < (r', \ell')$ with respect to the lexicographical order on
    pairs.

    The fact that $g(\block) = g(\block')$ implies that $(r,\ell)$ and $(r',
    \ell')$ are twins. Therefore one of two cases hold: In the first case,
    $(r,\ell) \eqblock (r',\ell')$ and $(r,\ell-1) \eqblock (r',\ell')$. This
    results in a contradiction, because then $\block = \block'$; i.e., they are
    not distinct. In the second case, $(r, \ell) \eqblock (r', \ell'-1)$ and
    $(r, \ell -1) \eqblock (r', \ell')$. In particular, $(r, \ell -1)$ and
    $(r',\ell')$ are both in the same block $\block'$ of $\partition$.
    Note that $(r',\ell') = \min \block'$.
    But $(r,\ell -1) < (r,\ell) < (r', \ell')$. This is a contradiction.
    Since both cases lead to contradictions, the assumption cannot hold.
    Therefore $g(\block) \neq g(\block')$ when $\block \neq \block'$, and $g$ is
    an injection.
\end{proof}

With these results it is easy to prove \Cref{lemma:mexp1}, restated below:

\lemmamexpone*

\begin{proof}
    Let $Z_\partition$ and $\Phi_\partition$ be as defined in
    \Cref{lemma:starting}. Using the bounds derived in
    \Cref{result:mexp1bound1,result:mexp1bound2}, we have for any $\partition
    \in \twinpartitions$:
    \[
        Z_\partition \cdot \Phi_\partition
        \leq
        n^{\abs{\twinclasses}}
        \cdot
        n^{-\abs{\twinclasses}}
        = 1.
    \]
    The result then follows immediately from \Cref{lemma:starting}.
\end{proof}

\subsubsection{Proof of \Cref{lemma:mexp2}}

In this part, we will bound $\expectation\left[ \elem^p \right]$ under the
assumption that $\abs*{\rv{X}_{ij}} \leq \nicefrac{1}{\sqrt{n}}$ almost surely.
Again, as per \Cref{lemma:starting}, it is sufficient to bound $Z_\partition
\cdot \Phi_\partition$ for all partitions $\partition$ satisfying the twin
property. In the following part, assume that $\partition \in
\twinpartitions$ and $\abs*{\rv{X}_{ij}} \leq \nicefrac{1}{\sqrt{n}}$ unless
otherwise stated.

We begin by obtaining a bound on the size of $\varphi$:
\begin{lemma}
    \label{result:mexp2bound1}
    For any $\tau \in \validindexings$ we have $\varphi(\tau) \leq
    \Phi_\partition$, where $\Phi_\partition = n^{-pk/2}$.
\end{lemma}

\begin{proof}
    We have
    \begin{align*}
        \varphi(\tau)
        &=
            \abs*{\expectation
            \prod_{r=1}^{p}
            \prod_{\ell=1}^{k}
            \rv{X}_{\tau_{\ell - 1}^{(r)} \tau_{\ell}^{(r)}}
            }
            ,\\
        &\leq
            \expectation
            \prod_{r=1}^{p}
            \prod_{\ell=1}^{k}
            \,
            \abs*{\rv{X}_{\tau_{\ell - 1}^{(r)} \tau_{\ell}^{(r)}}}
            .\\
    \intertext{By assumption, the magnitude of each entry of $\rv{X}$ is
    bounded by $\nicefrac{1}{\sqrt{n}}$. Therefore:}
        &\leq
            \prod_{r=1}^{p}
            \prod_{\ell=1}^{k}
            \,
            n^{\nicefrac{-1}{2}}
            ,\\
        &=
            n^{\nicefrac{-pk}{2}}.
    \end{align*}
\end{proof}

We next bound the number of indexing functions $\tau \in
\validindexings$. To this end, we will use the second part of
\Cref{result:numberoftau} in combination with the following non-trivial result
whose proof will constitute much of this section.

\begin{restatable}{lemma}{resultnumberofblocks}
    \label{result:numberofblocks}
    Let $\partition \in \twinpartitions$.
    Let $Q \subset \partition$ be the
    set of blocks in $\partition$ which contain an element of the form $(r,k)$
    for some $r \in \countingset{p}$. Then
    \[
        \abs*{\partition} \leq 
        \begin{cases}
            1 + \nicefrac{pk}{2},& \ix \in F,\\
            1 + \nicefrac{pk}{2} + \abs{Q} -
            \max\{\abs{Q}, \nicefrac{p}{2}\},&
                \ix \not \in F.
        \end{cases}
    \]
\end{restatable}

\begin{lemma}
    \label{result:mexp2bound2}
    Fix a set $F \subset \countingset{n}$
    and an $\ix \in \countingset{n}$.
    We have $\abs{\validindexings} \leq Z_\partition$, where
    \[
        Z_\partition = 
            n^{\nicefrac{pk}{2}}
            \cdot
            \begin{cases}
                1,& \ix \in F,\\
                \left(\frac{|F|}{n}\right)^{\nicefrac{p}{2}},& \ix \not \in F.
            \end{cases}
    \]
\end{lemma}

\begin{proof}
    Suppose first that $\ix \in F$. 
    From \Cref{result:numberoftau} we have:
    \begin{align*}
        \abs{\validindexings}
        &\leq
            n^{\abs{\partition} - \abs{Q} - 1}
            \cdot
            \abs{F}^{\abs{Q}}.
    \intertext{Applying the bound of $\abs{\partition} \leq
    1 + \nicefrac{pk}{2}$ from \Cref{result:numberofblocks}, we have}
        &\leq
            n^{\nicefrac{pk}{2} - \abs{Q}}
            \cdot
            \abs{F}^{\abs{Q}}, 
            \qquad(\ix \in F).
    \intertext{Since $F \subset \countingset{n}$, we have $\abs{F} \leq n$, and
    so:}
        &\leq
            n^{\nicefrac{pk}{2} - \abs{Q}}
            \cdot
            n^{\abs{Q}},\\
        &=
            n^{\nicefrac{pk}{2}}.
    \end{align*}
    This gives us the first part of our result: when $\ix \in F$.
    Now assume that $\ix \not \in F$.
    Again, from \Cref{result:numberoftau} we have
    \begin{align*}
        \abs{\validindexings}
        &\leq
            n^{\abs{\partition} - \abs{Q} - 1}
            \cdot
            \abs{F}^{\abs{Q}}.
    \intertext{Applying the bound on $\abs{\partition}$ from
    \Cref{result:numberofblocks}, we find:}
        &\leq
            n^{\nicefrac{pk}{2} - \max\{\abs{Q}, \nicefrac{p}{2}\}}
            \cdot
            \abs{F}^{\abs{Q}},
            \qquad (\ix \not \in F).
    \end{align*}

    There are two cases: $|Q| \geq \nicefrac{p}{2}$ and $|Q| <
    \nicefrac{p}{2}$. In the first case we find
    \begin{align*}
        |\validindexings|
        &\leq 
            |F|^{|Q|} 
            \cdot
            n^{\nicefrac{pk}{2} - 
            \max\{ |Q|, \nicefrac{p}{2} \}}
            ,\\
        &=
            |F|^{|Q|} 
            \cdot
            n^{\nicefrac{pk}{2} - |Q|}
            ,\\
        &=
            \left(\frac{|F|}{n}\right)^{|Q|}
            n^{\nicefrac{pk}{2}}
            .\\
    \intertext{Since $|F| \leq n$, we have:}
        &\leq
            \left(\frac{|F|}{n}\right)^{\nicefrac{p}{2}}
            n^{\nicefrac{pk}{2}}
            .\\
    \end{align*}
    In the other case where $|Q| < p/2$, we have:
    \begin{align*}
        |\validindexings|
        &\leq 
            |F|^{|Q|} 
            \cdot
            n^{\nicefrac{pk}{2} - 
            \max\{ |Q|, \nicefrac{p}{2} \}}
            ,\\
        &=
            |F|^{|Q|} 
            \cdot
            n^{\nicefrac{pk}{2} - \nicefrac{p}{2}}
            ,\\
        &\leq
            |F|^{\nicefrac{p}{2}} 
            \cdot
            n^{\nicefrac{pk}{2} - \nicefrac{p}{2}}
            ,\\
        &=
            \left(\frac{|F|}{n}\right)^{\nicefrac{p}{2}}
            n^{\nicefrac{pk}{2}}
            .\\
    \end{align*}
    Therefore, in both cases we find
    \[
        |\validindexings|
        \leq
            \left(\frac{|F|}{n}\right)^{\nicefrac{p}{2}}
            n^{\nicefrac{pk}{2}}
    \]
    when $\ix \not \in F$. This proves the result.
\end{proof}

We may now easily prove \Cref{lemma:mexp2}, restated below:

\lemmamexptwo*

\begin{proof}
    Let $Z_\partition$ and $\Phi_\partition$ be as defined in
    \Cref{lemma:starting}. Using the bounds derived in
    \Cref{result:mexp2bound1,result:mexp2bound2}, we have for any $\partition
    \in \twinpartitions$:
    \begin{align*}
        Z_\partition \cdot \Phi_\partition
        &\leq
            n^{\nicefrac{-pk}{2}}
            \cdot
            n^{\nicefrac{pk}{2}}
            \cdot
            \begin{cases}
                1,& \ix \in F,\\
                \left(\frac{|F|}{n}\right)^{\nicefrac{p}{2}},& \ix \not \in F.
            \end{cases},\\
        &=
            \begin{cases}
                1,& \ix \in F,\\
                \left(\frac{|F|}{n}\right)^{\nicefrac{p}{2}},& \ix \not \in F.
            \end{cases}
            .\\
    \end{align*}
    The result then follows immediately from \Cref{lemma:starting}.
\end{proof}

\paragraph{A careful count of index partitions.} In the remainder of this part,
we prove \Cref{result:numberofblocks} which was an important part of the proof
of \Cref{lemma:mexp2}. First we establish new notation and some intermediate
results.

Recall from \Cref{defn:index_partition} that a $(p,k)$-index partition is a
partition of a \emph{subset} of $\countingset{p} \times \{0,\ldots,k\}$. We have
so far only made use of ``full'' index partitions $\partition \in \partitions$
which partition the full set $\countingset{p} \times \{0,\ldots,k\}$. In this
section we will use index partitions in their full generality. In particular, we
will consider index subpartitions, defined as follows:

\begin{defn}
    Let $\partition$ be a $(p,k)$-index partition.
    $\partition'$ is an \term{$(p,k)$-index subpartition} of $\partition$ if 
    $\partition'$ is a $(p,k)$-index partition, 
    $\elements{\partition'} \subset \elements{\partition}$,
    and for any $(r,\ell),
    (\otherr, \otherell) \in \elements{\partition'}$, $(r, \ell)
    \eqblock[\partition'] (\otherr, \otherell)$ if and only if 
    $(r, \ell) \eqblock (\otherr, \otherell)$.
\end{defn}

We now define notation for referencing meaningful elements of
$\elements{\partition}$:

\begin{defn}
    Let $\partition$ be a $(p,k)$-index partition. For any $r \in
    \{1,\ldots, p\}$, define
    \begin{align*}
        \plast_\partition(r) 
            &= \max \{\ell : (r, \ell) \in \elements{\partition}\}.
    \intertext{This is always well-defined, since by the definition of an
    $(p,k)$-index partition, $(r,0) \in \elements{\partition}$. Similarly,
    we may define:}
        \pfirst_\partition(r) 
            &= \min \{\ell : (r, \ell) \in \elements{\partition}\}.
    \intertext{However, since $(r,0) \in \elements{\partition}$ by
    definition, it is always the case that $\pfirst_\partition(r) = 0$.
    For $(r, \ell) \in \elements{\partition}$ such that $\ell > 0$, define:}
        \pprev_\partition(r,\ell)
        &=
            \max \{
                \ell' : \ell' < \ell \text{ and } (r, \ell') \in
                \elements{\partition}
            \}
            ,
    \intertext{and for $(r,\ell) \in \elements{\partition}$ such
    that $\ell < \plast_\partition(r)$, define:}
        \pnext_\partition(r,\ell)
        &=
            \min \{
                \ell' : \ell' > \ell \text{ and } (r, \ell') \in
                \elements{\partition}
            \}
            .
    \end{align*}
\end{defn}

With this notation we may define a generalized notion of the twin property which
applies to any $(p,k)$-index partition, not just full partitions.

\begin{defn}[Twin property, generalized]
    Let $\partition$ be a $(p,k)$-index partition. Let $(r,\ell) \in
    \elements{\partition}$ and $(\otherr, \otherell) \in
    \elements{\partition}$ be distinct and such that $\ell,\otherell > 0$.
    We say that $(r,\ell)$ and $(\otherr,\otherell)$ are \term{twins in
    $\partition$} if either:
    \begin{enumerate}
        \item $(r, \ell) \eqblock (\otherr, \otherell)$ 
            and 
            $(r, \pprev_\partition(r, \ell)) 
            \eqblock 
            (\otherr,
            \pprev_\partition(\otherr, \otherell))$; or
        \item $(r, \ell) \eqblock (\otherr, \pprev_\partition(\otherr, \otherell))$ 
            and 
            $(r, \pprev_\partition(r, \ell)) 
            \eqblock 
            (\otherr, \otherell)$.
    \end{enumerate}
    We say that a $(p,k)$-index partition $\partition$ satisfies the
    \term{twin property} if for any pair $(r,\ell) \in
    \elements{\partition}$ with $\ell > 0$ there exists a distinct
    $(\otherr, \otherell) \in \elements{\partition}$ with $\otherell > 0$
    such that $(r,\ell)$ and $(\otherr, \otherell)$ are twins in
    $\partition$.
\end{defn}

\begin{claim}
    \label{claim:singletons}
    Let $\partition$ be a $(p,k)$-index partition satisfying the twin
    property with $p\geq 2$, and suppose that $\{\singletonpair\}$ is a singleton
    block in $\partition$. Then:
    \begin{enumerate}
        \item $0 < \singletonell < \plast_\partition(\singletonr)$, hence both
            $\pprev_\partition\singletonpair$ and
            $\pnext_\partition\singletonpair$
            are well-defined;
        \item $(\singletonr,\pnext_\partition\singletonpair)$ is the unique twin of
            $\singletonpair$ in $\partition$; and
        \item $(\singletonr, \pprev_\partition\singletonpair) \eqblock
            (\singletonr,
            \pnext_\partition\singletonpair)$.
    \end{enumerate}
\end{claim}

\begin{proof}
    First, suppose $\singletonell = 0$. Then $\singletonpair$ is in a block containing
    $\{(r,0) : r \in \{1,\ldots,p\}\}$ due to the definition of an
    $(p,k)$-index partition. Since $p \geq 2$, this block cannot be a
    singleton, and so it must be that $\singletonell > 0$.

    Because $\singletonell > 0$, we may invoke the twin property of $\partition$ to
    find a distinct pair $(\otherr, \otherell) \in \elements{\partition}$
    such that $(\otherr, \otherell)$ and $\singletonpair$ are twins and $\otherell
    > 0$. From the definition of the twin property, it must be that either
    $\singletonpair \eqblock (\otherr, \otherell)$ or $\singletonpair \eqblock (\otherr,
    \pprev_\partition(\otherr, \otherell))$. It cannot be that
    $\singletonpair \eqblock (\otherr,
    \otherell)$, as then the block containing $\singletonpair$ also contains the distinct
    pair $(\otherr, \otherell)$ by definition of $\eqblock$, contradicting the fact
    that the block is a singleton. Hence it must be the case that
    $\singletonpair
    \eqblock (\otherr, \pprev_\partition(\otherr, \otherell))$. Since the block
    contains only one element, this implies that $\otherr = \singletonr$ and
    $\pprev_\partition(\otherr, \otherell) = \pprev_\partition(\singletonr, \otherell)
    = \singletonell$. This implies that $\otherell =
    \pnext_\partition(\singletonell)$. Since
    $\pnext_\partition\singletonpair$ is unique, this implies that
    $(\singletonr,\pnext_\partition\singletonpair)$ is the unique twin of
    $\singletonpair$ in
    $\partition$; it's existence implies that $\singletonell <
    \plast_\partition(\singletonr)$.

    We now prove the third part of the claim.
    Since it cannot be that $\singletonpair \eqblock (\otherr, \otherell)$, the
    twin property implies that $(\singletonr, \pprev_\partition(\singletonell)) \eqblock
    (\otherr, \otherell)$. But, as shown above, $(\otherr, \otherell) =
    (\singletonr,
    \pnext_\partition\singletonpair)$. Therefore $(\singletonr,
    \pprev_\partition\singletonpair)
    \eqblock (\singletonr, \pnext_\partition\singletonpair)$.
\end{proof}

\begin{lemma}[Removing singletons preserves twin property]
    \label{lemma:remove}
    Let $\partition$ be a $(p,k)$-index partition satisfying the twin
    property with $p \geq 2$, and suppose that $\{\singletonpair\}$ is a singleton
    block in $\partition$. Let $\partition'$ be obtained by removing the
    block $\{\singletonpair\}$ from $\partition$ and deleting $(\singletonr,
    \pnext_\partition\singletonpair)$ from the block which contains it
    (note that $\pnext_\partition\singletonpair$ is well-defined according to
    \Cref{claim:singletons}). Then 
    \begin{enumerate}
        \item $\partition'$ is a $(p,k)$-index subpartition of $\partition$
            which satisfies the twin property;
        \item $|\partition'| = |\partition| - 1$ and
            $\left|\elements{\partition'}\right| =
            \left|\elements{\partition'}\right| - 2$;
        \item for every $\otherr \in \{1,\ldots,p\}$, $(\otherr,
        \plast_\partition(\otherr)) \eqblock (\otherr,
        \plast_{\partition'}(\otherr))$.
    \end{enumerate}
\end{lemma}

\begin{proof}
    First, $\partition'$ is a partition of
    $\elements{\partition} \setminus \{\singletonpair,
        (\singletonr,\pnext_\partition\singletonpair\}$; we need only verify
        that removing $(\singletonr,
    \pnext_\partition\singletonpair)$ from the block containing it cannot create
    an empty block. This is true, as $(\singletonr, \pprev_\partition\singletonpair)$ was in
    the same block of $\partition$ as $(\singletonr, \pnext_\partition\singletonpair)$
    according to \Cref{claim:singletons}, and $(\singletonr,
    \pprev_\partition(\singletonr,
    \singletonell))$ remains in $\elements{\partition'}$. Hence $|\partition'| =
    |\partition| - 1$, and $\left|\elements{\partition'}\right| =
    \left|\elements{\partition}\right| - 2$. Moreover, $(\singletonr,0)$ was not
    removed from the block containing it, since $\singletonell >
    0$ by \Cref{claim:singletons}, and it is clear that
    $\pnext_\partition\singletonpair$ cannot be $(\singletonr,0)$ by the definition of
    $\pnext$. Therefore, there exists a block $\block \in \partition'$ which
    contains all of $\{(r',0) : r \in \{1,\ldots,p\}\}$. This is sufficient
    to show that $\partition'$ is a $(p,k)$-index partition. Furthermore, it
    is clear that any two elements in $\elements{\partition}$ are in the
    same block of $\partition'$ if and only if they are in the same block of
    $\partition$. Hence $\partition'$ is a $(p,k)$-index subpartition of
    $\partition$

    We now show that $\partition'$ satisfies the twin property. First we
    prove a useful intermediate result: for any $(\otherr, \otherell) \in
    \elements{\partition'}$ with $\otherell > 0$,
    $(\otherr, \pprev_{\partition'}(\otherr, \otherell)) \eqblock
    (\otherr, \pprev_\partition(\otherr,\otherell))$. 
    If $\pnext_\partition(\singletonr, \singletonell) = \plast(\singletonr)$ then the above is trivially
    true, since in fact $(\otherr, \pprev_{\partition'}(\otherr, \otherell)) =
    (\otherr, \pprev_\partition(\otherr,\otherell))$ for all $(\otherr,
\otherell) \in \elements{\partition'}$. If $\pnext_\partition\singletonpair \neq
\plast(\singletonr)$, then $(r', \ell') :=
\pnext_\partition(\pnext_\partition\singletonpair)$ is
well-defined; moreover, it 
is the only element in
$\elements{\partition'}$ 
for which $\pprev_{\partition'} \neq
\pprev_{\partition}$.
It follows from the definitions of $\pnext$ and $\pprev$ that
$\pprev_{\partition'}(r',\ell') =
\pprev_{\partition}\singletonpair$
and $\pprev_\partition(r',\ell') = \pnext_\partition\singletonpair$. Furthermore,
$(\singletonr,\pprev_\partition\singletonpair) \in \elements{\partition}$ by
\Cref{claim:singletons}, and it was not removed with the singleton, so it is
also in $\elements{\partition'}$. Also by \Cref{claim:singletons},
$\pprev_\partition\singletonpair \eqblock \pnext_\partition\singletonpair$. In
summary:
\[
    \pprev_{\partition'}(r',\ell')
    =
    \pprev_\partition\singletonpair
    \eqblock
    \pnext_\partition\singletonpair
    =
    \pprev_\partition(r',\ell').
\]
Therefore $\pprev_{\partition'}(r',\ell') \eqblock
\pprev_{\partition}(r',\ell')$, and the intermediate result is proven.

Now let $(\thirdr, \thirdell) \in \elements{\partition'}$ with $\thirdell >
0$. We will show that it necessarily has a twin in $\partition'$. Let
$(\otherr, \otherell)$ be its twin in $\partition$ -- the existence of such
a twin is guaranteed by the fact that $\partition$ satisfies the twin
property. 
It cannot be that $(\otherr, \otherell)$ is missing from $\partition'$; if
it were missing, it would mean that $(\otherr, \otherell)$ is one of
$(\singletonr,
\singletonell)$ or $(\singletonr,\pnext_\partition\singletonpair)$, and since
$(\singletonr,
\singletonell)$ and $(\singletonr,
\pnext_\partition\singletonpair)$ are unique twins in $\partition$, this would imply
that $(\thirdr, \thirdell)$ is either $\singletonpair$ or $(\singletonr,
\pnext_\partition\singletonpair)$, and would therefore have been removed.  Hence
$(\otherr, \otherell) \in \elements{\partition'}$.

By the definition of twins, it must be that either 
\begin{enumerate}
    \item $
        (\otherr, \otherell) 
        \eqblock 
        (\thirdr, \thirdell)
        $ and
        $
        (\otherr, \pprev_\partition(\otherr, \otherell)) 
        \eqblock 
        (\thirdr, \pprev_\partition(\thirdr, \thirdell))
        $; or
    \item $
        (\otherr, \otherell) 
        \eqblock 
        (\thirdr, \pprev_\partition(\thirdr, \thirdell))
        $ and
        $
        (\otherr, \pprev_\partition(\otherr, \otherell)) 
        \eqblock 
        (\thirdr, \thirdell).
        $
\end{enumerate}

Assume the first case. Then since $(\thirdr, \thirdell) \eqblock (\otherr,
\otherell)$ and both pairs exist in $\elements{\partition'}$, we have $(\thirdr,
\thirdell) \eqblock[\partition'] (\otherr, \otherell)$. From the above
intermediate result, we see that 
$(\otherr, \pprev_{\partition'}(\otherr,
\otherell)) \eqblock (\otherr, \pprev_{\partition}(\otherr,\otherell))$
and
$(\thirdr, \pprev_{\partition'}(\thirdr,
\thirdell)) \eqblock (\thirdr, \pprev_{\partition}(\thirdr,\thirdell))$.
Since in this case
        $
        (\otherr, \pprev_\partition(\otherr, \otherell)) 
        \eqblock 
        (\thirdr, \pprev_\partition(\thirdr, \thirdell))
        $, this implies 
$(\otherr, \pprev_{\partition'}(\otherr,
\otherell)) \eqblock[\partition'] (\thirdr,
\pprev_{\partition}(\thirdr,\thirdell))$. Hence $(\otherr, \otherell)$ and
$(\thirdr, \thirdell)$ are twins in $\partition'$.

Now assume the second case. From the intermediate result, we see that:
$(\thirdr, \pprev_{\partition'}(\thirdr,
\thirdell)) \eqblock (\thirdr, \pprev_{\partition}(\thirdr,\thirdell))$ and
hence $(\otherr, \otherell) \eqblock
(\thirdr, \pprev_{\partition'}(\thirdr,\thirdell))$. Therefore
$(\otherr, \otherell) \eqblock[\partition']
(\thirdr, \pprev_{\partition'}(\thirdr,\thirdell))$.
The symmetric argument shows
$(\thirdr, \thirdell) \eqblock[\partition']
(\otherr, \pprev_{\partition'}(\otherr,\otherell))$. As a result, $(\otherr,
\otherell)$ and $(\thirdr, \thirdell)$ are twins in $\partition'$.

In both cases, $(\otherr, \otherell)$ and $(\thirdr, \thirdell)$ are twins in
$\partition'$. Since $(\thirdr, \thirdell)$ was arbitrary, it follows that
$\partition'$ satisfies the twin property.

Finally, we show that for every $\otherr \in \elements{\partition}$, 
$\plast_{\partition'}(\otherr) \eqblock \plast_{\partition}(\otherr)$. There
are two cases. If $\plast_\partition(\otherr) \in \elements{\partition'}$,
we are done. If $\plast_\partition(\otherr) \not\in \elements{\partition'}$,
then it must have been removed by the above procedure, and hence either
$(\otherr, \plast_\partition(\otherr))$ is $\singletonpair$ or $(\singletonr,
\pnext_\partition\singletonpair)$. It cannot be $\singletonpair$ as per
\Cref{claim:singletons}. Therefore it must be that $(\otherr,
\plast_\partition(\otherr)) = (r, \pnext_\partition\singletonpair)$. Then
$\plast_{\partition'}(\otherr) = \plast_{\partition'}(\singletonr) =
\pprev_{\partition}\singletonpair$. Since $(\singletonr, \pprev_{\partition}\singletonpair)
\eqblock (\singletonr, \pnext_{\partition}\singletonpair)$, it follows that
$(\otherr, \plast_{\partition'}(\otherr)) \eqblock
(\otherr, \plast_{\partition}(\otherr))$, proving the claim.
\end{proof}

\begin{lemma}
    \label{lemma:countindexings}
    Let $\partition$ be a $(p,k)$-index partition satisfying the twin
    property, with $p \geq 2$. Then
    \[
        |\partition| \leq 
        1 +
        \frac{1}{2} 
        \left(
            |\elements{\partition}| - p
        \right).
    \]
    Moreover, let $\zeroblock$ be the block of $\partition$ which contains
    $\{(r,0) : r \in \{1,\ldots,p\}\}$ as a subset, and define
    \[
        \rlendset = \{ \block \in \partition
            :
            (r,\plast_\partition(r)) \in \block
            \text{ for some } r \in \{1,\ldots,p\}
        \}.
    \]
    If $\zeroblock \not \in \rlendset$, then
    \[
        |\partition|
        \leq
            1
            +
            |\rlendset|
            +
            \frac{1}{2}\left(|\elements{\partition}| - p -
                \max\{ 2|\rlendset|, p\}
            \right).
    \]
\end{lemma}

\begin{proof}
    Define $\partition_0 = \partition$, and let $\partition_0, \partition_1,
    \ldots, \partition_T$ be a sequence of $(p,k)$-index partitions in which
    $\partition_{t+1}$ is obtained by removing an arbitrary singleton
    $(\singletonr_{t}, \singletonell_{t})$ from $\partition_{t}$ along with
    $(\singletonr_t,
    \pnext_{\partition_{t}}(\singletonr_t, \singletonell_t))$ in the manner
    of \Cref{lemma:remove}, such that $\partition_T$ has no singletons.

    Because
    $\partition_T$ is a $(p,k)$-index partition, it contains a block which
    is a superset of $\{(r,0) : r \in \{1,\ldots,p\}\}$. There are at least
    $p$ elements in this block, implying that there are at most
    $|\elements{\partition_T}| - p$ pairs \emph{outside} of this block,
    distributed over $|\partition_T| - 1$ blocks. Since each of these blocks
    is not a singleton by construction of $\partition_T$, each block has at
    least 2 elements. Therefore:
    \[
        |\partition_T| - 1 \leq \frac{1}{2}
        \left(
            |\elements{\partition_T}| - p
        \right).
    \]
    Since $|\partition_{t+1}| = |\partition_t| - 1$, it follows that
    $|\partition_T| = |\partition| - T$. Similarly, since
    $|\elements{\partition_{t+1}}| = |\elements{\partition_t}| - 2$, we have
    $|\elements{\partition_T}| = |\elements{\partition}| - 2T$. Substituting
    these into the above inequality, we find:
    \begin{align*}
        |\partition| 
        &\leq
            1 + T
            + \frac{1}{2}
            \left(
                |\elements{\partition}| - 2T - p
            \right)\\
        &= 1 + \frac{1}{2} \left(
                |\elements{\partition}| - p
            \right).
    \end{align*}

    This proves the general claim. Now suppose that $\zeroblock \not \in
    \rlendset$.
    Let $\zeroblock_T$ be the block of $\partition_T$ which contains
    $\{(r,0) : r \in \{1,\ldots,p\}\}$ as a subset, and let
    \[
        \rlendset_T = \{ \block \in \partition_T
            :
            (r,\plast_{\partition_T}(r)) \in \block
            \text{ for some } r \in \{1,\ldots,p\}
        \}.
    \]
    We claim that $\zeroblock_T \not \in \rlendset_T$, and $|\rlendset_T| =
    |\rlendset|$. First,
    it is easy to see that $\partition_T$ is a $(p,k)$-index subpartition of
    $\partition_0 = \partition$. Second, it follows from the third part of
    \Cref{lemma:remove} and induction that $(r, \plast_\partition(r)) \eqblock
    (r,\plast_{\partition_T}(r))$ for any $r \in \{1,\ldots,p\}$; in other
    words, for each block $\block \in \rlendset$ there is a block $\block_T
    \in \rlendset_T$
    such that $\block_T \subset \block$. Recall
    that
    $(r,\ell) \eqblock[\partition_T] (\otherr, \otherell)$ if and only if
    $(r, \ell) \eqblock (\otherr, \otherell)$ by the definition of
    $(p,k)$-index subpartition. Therefore, the blocks of $\rlendset_T$ are in
    bijection with the blocks of $\rlendset$, such that $\block_T \in \rlendset_T$ maps to
    $\block \in \rlendset$ if and only if $\block \supset \block_T$. Hence
    $|\rlendset_T| =
    |\rlendset|$. Moreover, since $\zeroblock_T \subset \zeroblock$, it follows that
    $\zeroblock_T \not \in \rlendset_T$.

    We use this to improve the bound on $|\partition_T|$. As argued above,
    there are at least $p$ elements in $\zeroblock_T$. Now, however, there
    is a set of blocks $\rlendset_T$ which (importantly) does not contain
    $\zeroblock_T$. The total number of elements in all blocks of $\rlendset_T$ is
    \[
        \sum_{\block \in \rlendset_T} |\block|.
    \]
    Hence there are
    \[
        |\elements{\partition_T}| - p -
        \sum_{\block \in \rlendset_T} |\block|
    \]
    pairs distributed among the blocks of $\partition_T$ which are not in
    $\rlendset_T$ and which are not $\zeroblock_T$. Since none of these blocks are
    singleton by construction of $\partition_T$, these elements can form at
    most
    \[
        \frac{1}{2}\left(|\elements{\partition_T}| - p -
        \sum_{\block \in \rlendset_T} |\block|\right)
    \]
    blocks. Hence:
    \[
        |\partition_T| \leq 
        1
        +
        |\rlendset_T|
        +
        \frac{1}{2}\left(|\elements{\partition_T}| - p -
        \sum_{\block \in \rlendset_T} |\block|\right).
    \]
    Now, each $\block \in \rlendset_T$ has at least two elements, since
    $\partition_T$ contains no singletons. Therefore:
    \[
        \sum_{\block \in \partition} |\block|
        \geq
        2 |\rlendset_T|.
    \]
    Simultaneously, we know that
    \[
        \{(r, \plast_{\partition'}(r)) : r \in \{1,\ldots,p\} \}
        \subset
        \bigcup_{\block \in \rlendset_T} \block.
    \]
    The set on the LHS contains exactly $p$ elements. Hence
    \[
        \sum_{\block \in \partition} |\block|
        \geq
        p.
    \]
    Therefore,
    \[
        \sum_{\block \in \partition} |\block|
        \geq
        \max\{
            2|\rlendset_T|, p
        \}.
    \]
    Substituting this into the bound for $|\partition_T|$, we find:
    \[
        |\partition_T| \leq 
        1
        +
        |\rlendset_T|
        +
        \frac{1}{2}\left(|\elements{\partition_T}| - p -
            \max\{ 2|\rlendset_T|, p\}
        \right).
    \]
    As before, we have $|\partition| = |\partition_T| + T$ and
    $|\elements{\partition_T}| = |\elements{\partition}| - 2T$, and
    therefore
    \begin{align*}
        |\partition| 
        &\leq 
            1
            +
            |\rlendset_T|
            +
            \frac{1}{2}\left(|\elements{\partition}| - 2T - p -
                \max\{ 2|\rlendset_T|, p\}
            \right)
            +
            T,\\
        &= 
            1
            +
            |\rlendset_T|
            +
            \frac{1}{2}\left(|\elements{\partition}| - p -
                \max\{ 2|\rlendset_T|, p\}
            \right)
            .
    \intertext{Since $|\rlendset_T| = |\rlendset|$, we arrive at:}
        &= 
            1
            +
            |\rlendset|
            +
            \frac{1}{2}\left(|\elements{\partition}| - p -
                \max\{ 2|\rlendset|, p\}
            \right)
            .
    \end{align*}
\end{proof}

We can now prove \Cref{result:numberofblocks}, restated below:

\resultnumberofblocks*

\begin{proof}
    $\partition$ is a full $(p,k)$-index partition and it satisfies the twin
    property by assumption. Moreover, if $\ix \not \in F$, then the root block
    is not in 
    $
        \rlendset = \{ \block \in \partition
            :
            (r,\plast_\partition(r)) \in \block
            \text{ for some } r \in \{1,\ldots,p\}
        \}.
    $
    We have that $|\elements{\partition}| = p(k+1)$, and so the result
    follows immediately from
    \Cref{lemma:countindexings}.
\end{proof}

\label{sec:interactionproofs}

\begin{lemma}
    \label{result:interactionseries}
    Suppose that $\abs*{\left(X^k u\right)_\ix} \leq \beta Q^k$ for all $k \leq
    K$.  Let $\eta$ be a positive number, and suppose $\eta < \min \{Q^{-1},
    \spectralnorm{X}^{-1}\}$. Then:
    \[
        \sum_{k \geq 1}
        \,
        \abs*{\left[(\eta X)^k u\right]_\ix}
        \leq
        \frac{\beta \eta Q}{1 - \eta Q}
        +
        \frac{
            \twonorm{u}
            \cdot
            \spectralnorm{\eta X}^{K+1}
        }{
            1 - \spectralnorm{\eta X}
        }.
    \]
\end{lemma}

\begin{proof}
    We have:
    \[
        \sum_{k \geq 1}
        \,
        \abs*{\left[(\eta X)^k u\right]_\ix}
        =
        \underbrace{
            \sum_{k = 1}^K
            \abs*{\left[(\eta X)^k u\right]_\ix}
        }_{\#1}
        +
        \underbrace{
            \sum_{k > K}
            \abs*{\left[(\eta X)^k u\right]_\ix}.
        }_{\#2}
    \]
    We begin by bounding \#1. For each $1 \leq k \leq K$, we have 
    \[
        \abs*{\left[(\eta X)^k u\right]_\ix}
        =
        \eta^k \abs*{\left(X^k u\right)_\ix}
        \leq
        \beta (\eta Q)^k.
    \]
    The last step follows from the assumption that $\eta Q < 1$. As a result:
    \begin{align*}
        \sum_{k = 1}^K
        \,
        \abs*{\left[(\eta X)^k u\right]_\ix}
        &\leq
            \beta
            \sum_{k = 1}^K
            (\eta Q)^k
            ,\\
        &\leq
            \beta
            \sum_{k = 1}^\infty
            (\eta Q)^k
            ,\\
        &=
            \beta \eta Q
            \sum_{k = 0}^\infty
            (\eta Q)^k
            ,\\
        &=
            \frac{
                \beta \eta Q
            }{
                1 - \eta Q
            }
            .
    \end{align*}
    
    We next bound \#2. Here we will use the assumption that
    $\spectralnorm{\eta X} < 1$ combined with the fact that the
    $\infty$-norm of a vector is bounded above by the 2-norm. We have:
    \begin{align*}
        \sum_{k > K}
        \abs*{\left[(\eta X)^k u\right]_\ix}
        &\leq
            \sum_{k > K}
            \inftynorm*{(\eta X)^k u}
            ,\\
        &\leq
            \sum_{k > K}
            \twonorm*{(\eta X)^k u}
            ,\\
        &\leq
            \sum_{k > K}
            \spectralnorm*{(\eta X)^k} \cdot \twonorm{u}
            ,\\
        &=
            \sum_{k > K}
            \spectralnorm*{\eta X}^k \cdot \twonorm{u}
            ,\\
        &=
            \twonorm{u}
            \cdot
            \spectralnorm{\eta X}^{K+1}
            \sum_{k \geq 0}
            \spectralnorm{\eta X}^k
            ,\\
        &=
            \frac{
                \twonorm{u}
                \cdot
                \spectralnorm{\eta X}^{K+1}
            }{
                1 - \spectralnorm{\eta X}
            }
            .
    \end{align*}
\end{proof}

\subsection{Proofs of
\Cref{result:boundzeta,result:boundzetamag,result:boundzetablock}}

We are now able to prove the main results of this section, restated below:

\restateboundzeta*{}

\begin{proof}
    \label{proof:boundzeta}
    We have
    \begin{align}
        \tailboundraw_\ix
        &=\nonumber
            \sum_{p \geq 1}
            \;
                \abs*{
                    \left[
                    \left(
                        \frac{H}{\eigval}
                    \right)^p
                    u
                    \right]_\ix
                }
            ,\\
        &=\nonumber
            \inftynorm{u}
            \,
            \sum_{p \geq 1}
            \;
                \abs*{
                    \left[
                    \left(
                        \frac{H}{\eigval}
                    \right)^p
                    \cdot
                    \frac{u}{\inftynorm{u}}
                    \right]_\ix
                }
            ,\\
        &=\nonumber
            \inftynorm{u}
            \,
            \sum_{p \geq 1}
            \;
                \abs*{
                    \left[
                    \left(
                        \frac{\gamma}{\eigval}
                        \cdot
                        \frac{H}{\gamma}
                    \right)^p
                    \cdot
                    \frac{u}{\inftynorm{u}}
                    \right]_\ix
                }
            ,\\
    \intertext{Define $X = H / \gamma$, $\eta = \frac{\gamma}{\eigval}$, and $v
    = u / \inftynorm{u}$. Then:}
        &=
            \label{eqn:seriesintermed}
            \inftynorm{u}
            \sum_{p \geq 1}
            \;
                \abs*{
                    \left[
                    \left(
                        \eta X
                        \right)^p
                        v
                    \right]_\ix
                }
                .
    \end{align}
    Note that
    $
        \expectation \abs{X_{ij}}^p
        =
        \expectation \abs{H_{ij} / \gamma}^p.
    $
    Thus for all $p \geq 2$ we have $\expectation \abs{X_{ij}}^p \leq
    \nicefrac{1}{n}$. We may therefore invoke the first result in
    \Cref{result:interaction} to derive, for all $p \leq \frac{\kappa}{8} (\log
    n)^{\xi}$,
    \begin{equation}
        \prob\left(
            \abs*{\left(X^p v \right)_\ix}^k
            \,\geq\,
            (\log n)^{k \xi}
        \right)
        \leq
        1 - 
            n^{-\frac{1}{4} (\log_\mu n)^{\xi-1} (\log_\mu
            e)^{-\xi}}.
    \end{equation}
    We now bound
    $
        \sum_{p \geq 1}
        \,
            \abs*{
                \left[
                \left(
                    \eta X
                    \right)^p
                    v
                \right]_\ix
            }
    $
    by applying \Cref{result:interactionseries} with $X = H/\gamma$, $\beta =
    1$, $\eta = \gamma / \eigval$, $Q = (\log n)^\xi$ and $K =
    \left\lfloor\frac{\kappa}{8} (\log n)^\xi\right\rfloor$.
    One of the requirements of \Cref{result:interactionseries} is that $\eta =
    \gamma / \eigval$ must satisfy:
    \begin{align*}
        \label{eqn:Dbound}
        \frac{\gamma}{\eigval}
        &< 
            \min\left\{
                Q^{-1},\;
                \spectralnorm{X}^{-1}
            \right\}
        =
            \min\left\{
                (\log n)^{-\xi}
                    ,\;
                \gamma
                \spectralnorm{H}^{-1}
            \right\}
            .
    \end{align*}
    Hence we must have $\gamma < \eigval (\log n)^{-\xi}$ and $\eigval >
    \spectralnorm{H}$, as assumed. Then, applying the result of
    \Cref{result:interactionseries}, we have:
    \begin{align*}
        \tailboundraw_\ix(H, \eigval, u)
        &=
            \inftynorm{u}
            \sum_{p \geq 1}
            \;
                \abs*{
                    \left[
                    \left(
                        \eta X
                        \right)^p
                        v
                    \right]_\ix
                }
                ,\\
        &\leq
            \inftynorm{u}
            \left(
                \frac{\gamma (\log n)^\xi}{\eigval - \gamma (\log n)^\xi}
                +
                \frac{
                    \spectralnorm{H/\eigval}^{\lfloor 
                        \frac{\kappa}{8}(\log n)^\xi + 1
                    \rfloor}
                }{
                    1 - \spectralnorm{H/\eigval}
                }
                \cdot
                \frac{\twonorm{u}}{\inftynorm{u}} 
            \right)
            ,\\
        &=
            \frac{\gamma (\log n)^\xi}{\eigval - \gamma (\log n)^\xi}
            \cdot \inftynorm{u}
            +
            \frac{
                \spectralnorm{H/\eigval}^{\lfloor 
                    \frac{\kappa}{8}(\log n)^\xi + 1
                \rfloor}
            }{
                1 - \spectralnorm{H/\eigval}
            }
            \cdot
            \twonorm{u}
            .
    \end{align*}
\end{proof}

\restateboundzetamag*{}

\begin{proof}
    \label{proof:boundzetamag}
    The proof follows that of \Cref{result:boundzeta} almost identically.
    Picking up from  \Cref{eqn:seriesintermed} in that proof,
    we invoke the second result in
    \Cref{result:interaction} to derive, for all $p \leq \frac{\kappa}{8} (\log
    n)^{\xi}$,
    \begin{equation}
        \prob\left(
            \abs*{\left(X^p v \right)_\ix}^k
            \,\geq\,
            \beta_\ix\, (\log n)^{k \xi}
        \right)
        \leq
        1 - 
            n^{-\frac{1}{4} (\log_\mu n)^{\xi-1} (\log_\mu
            e)^{-\xi}},
    \end{equation}
    where $\beta_\ix$ is defined as:
    \[
        \beta_\ix =
        \begin{cases}
            1,& \ix \in F,\\
            \sqrt{\frac{\abs{F}}{n}},& \ix \not \in F.
        \end{cases}
    \]
    We now bound
    $
        \sum_{p \geq 1}
        \,
            \abs*{
                \left[
                \left(
                    \eta X
                    \right)^p
                    v
                \right]_\ix
            }
    $
    by applying \Cref{result:interactionseries} with $X = H/\gamma$, $\beta =
    \beta_\ix$, $\eta = \gamma / \eigval$, $Q = (\log n)^\xi$ and $K =
    \left\lfloor\frac{\kappa}{8} (\log n)^\xi\right\rfloor$.
    We see that the result will be the same as that of \Cref{result:boundzeta}
    except for an extra factor of $\beta_\ix$ in the first term. 
\end{proof}

\restateboundzetablock*{}

\begin{proof}
    \label{proof:boundzetablock}
    For any $(n,K)$-block vector $u$ with blocks $F_1, \ldots, F_K$, we have
    \[
        u = 
        \sum_{k=1}^K c_k(u) \indicator{F_k}.
    \]
    Hence:
    \begin{align*}
        \tailboundraw_\ix(H, \eigval, u)
        &=\nonumber
            \sum_{p \geq 1}
            \;
                \abs*{
                    \left[
                    \left(
                        \frac{H}{\eigval}
                    \right)^p
                    u
                    \right]_\ix
                }
            ,\\
        &=\nonumber
            \sum_{p \geq 1}
            \;
                \abs*{
                    \left[
                    \left(
                        \frac{H}{\eigval}
                    \right)^p
                    \sum_{k=1}^K c_k(u) \indicator{F_k}
                    \right]_\ix
                }
            ,\\
        &=\nonumber
            \sum_{k=1}^K
                c_k(u)
                \cdot
            \sum_{p \geq 1}
            \;
                \abs*{
                    \left[
                    \left(
                        \frac{H}{\eigval}
                    \right)^p
                        \indicator{F_k}
                    \right]_\ix
                }
            ,\\
        &=\nonumber
            \sum_{k=1}^K
                c_k(u)
                \cdot
                \tailboundraw_\ix(H, \eigval, \indicator{F_k})
            .
    \intertext{All of the assumptions of \Cref{result:boundzetamag}
    hold, and we use it to bound $\tailboundraw_\ix(H, \eigval,
    \indicator{F_k})$ for each $k \in \countingset{K}$. We find:}
        &=
            \sum_{k=1}^K
                c_k(u)
                \cdot
                \left(
                    \frac{\beta_{\ix,k}\, \gamma \, 
                        (\log n)^\xi}{\eigval - \gamma \, (\log n)^\xi}
                    +
                    \sqrt{\abs{F_k}} \cdot
                    \frac{
                        \spectralnorm{H/\eigval}^{\lfloor 
                            \frac{\kappa}{8}(\log n)^\xi + 1
                        \rfloor}
                    }{
                        1 - \spectralnorm{H/\eigval}
                    }
                \right),
    \end{align*}
    where we have defined
    \[
        \beta_{\ix,k} =
        \begin{cases}
            1,& \ix \in F_k,\\
            \sqrt{\frac{\abs{F_k}}{n}},& \ix \not \in F_k.
        \end{cases}
    \]
    In the last line we used the fact that $\inftynorm{\indicator{F_k}} = 1$ and
    $\twonorm{\indicator{F_k}} = \sqrt{\abs{F_k}}$.  Since we invoked
    \Cref{result:boundzetamag} for each of the $K$ indicator vectors, a union
    bound gives that this result holds with probability at least:
    $
        1 
        - 
            K n^{-\frac{1}{4} (\log_b n)^{\xi-1} (\log_b
            e)^{-\xi} + 1},
    $
    where 
    $b = \left(\frac{\kappa + 1}{2}\right)^{-1}$.

\end{proof}

\end{document}